%% file: arxiv_v1.tex
\documentclass{article}

\usepackage{arxiv}

\usepackage[utf8]{inputenc} 
\usepackage[T1]{fontenc}    
\usepackage{hyperref}       
\usepackage{url}            
\usepackage{booktabs}       
\usepackage{amsfonts}       
\usepackage{nicefrac}       
\usepackage{microtype}      
\usepackage{lipsum}
\usepackage{graphicx}
\input{package.tex}

\input{definitions.tex}

\usepackage{changes}
\usepackage{lipsum}
\usepackage{lineno,hyperref}
\modulolinenumbers[5]
\title{A Parallel Technique for Multi-objective Bayesian Global Optimization: Using a Batch Selection of Probability of Improvement}

\author{
 Kaifeng Yang \\
  HEAL, University of Applied Sciences Upper Austria,\\  Softwarepark 11, 4232, Hagenberg, \\
  Austria \\
  \texttt{kaifeng.yang@fh-hagenberg.at} \\
   \And
 Michael Affenzeller \\
  HEAL, University of Applied Sciences Upper Austria,\\  Softwarepark 11, 4232, Hagenberg, \\
  Austria \\
  \texttt{michael.affenzeller@fh-hagenberg.at} \\
  \And
Guozhi Dong \\
School of Mathematics and Statistics, \\
Central South University, 
\\ Lushan South-road No. 932, 410083 \\
Changsha, China
}
\begin{document}
\maketitle
\begin{abstract}
Bayesian global optimization (BGO) is an efficient surrogate-assisted technique for problems involving expensive evaluations. A parallel technique can be used to parallelly evaluate the true-expensive objective functions in one iteration to boost the execution time. An effective and straightforward approach is to design an acquisition function that can evaluate the performance of a bath of multiple solutions, instead of a single point/solution, in one iteration. 
This paper proposes five alternatives of \emph{Probability of Improvement} (PoI) with multiple points in a batch (q-PoI) for multi-objective Bayesian global optimization (MOBGO), taking the covariance among multiple points into account. Both exact computational formulas and the Monte Carlo approximation algorithms for all proposed q-PoIs are provided. Based on the distribution of the multiple points relevant to the Pareto-front, the position-dependent behavior of the five q-PoIs is investigated. Moreover, the five q-PoIs are compared with the other nine state-of-the-art and recently proposed batch MOBGO algorithms on twenty bio-objective benchmarks. The empirical experiments on different variety of benchmarks are conducted to demonstrate the effectiveness of two greedy q-PoIs ($\kpoi_{\mbox{best}}$ and $\kpoi_{\mbox{all}}$) on low-dimensional problems and the effectiveness of two explorative q-PoIs ($\kpoi_{\mbox{one}}$ and $\kpoi_{\mbox{worst}}$) on high-dimensional problems with difficult-to-approximate Pareto front boundaries.
\end{abstract}


\section{Introduction}

The surrogate model is regarded as a promising approach to incorporating powerful computational techniques into simulation-based optimization, as they replace exact but expensive simulation outputs with approximations learned from past outputs. Compared to the exact evaluations, the more expensive the evaluation of the simulation is, the greater the leverage of such approaches is in terms of the runtime boost. Two typical examples of time-consuming simulation models are finite element simulations~\cite{Fleck2019} and computational fluid dynamics (CDF) simulations~\cite{LI2021100203}, and in both cases, a single run can consume many hours. 
Therefore, most evolutionary optimization algorithms can not be directly applied to such expensive simulation problems. The most common remedy for computationally expensive simulation is to replace exact objective function evaluations with  predictions from surrogate models, for instance, Gaussian processes or Kriging models~\cite{LI2021100203, HAN2022100988}, random forest~\cite{FeurerKES0H19}, supported vector machine~\cite{DAI2020115332}, symbolic regression~\cite{Fleck2019}, to name a few.

\emph{Bayesian Global Optimization} (BGO) was proposed by Jonas Mockus and Antanas {\v{Z}}ilinskas \cite{vzilinskas1972one,Mockus1978}, and it is popularized by Jones et al. \cite{jones1998efficient} (known as EGO). Underpinned by surrogate-assisted modeling techniques, it sequentially selects promising solutions by using the prediction and uncertainty of the surrogate model. The basic idea of BGO is to build a Gaussian process model to reflect the relationship between a decision vector $\mathbf{x} = (x_1, \cdots, x_d)^\top$ and its corresponding objective value $y = f(\mathbf{x})$. 
A BGO algorithm searches for an optimal solution $\mathbf{x}^*$ by using the predictions of the Gaussian process model instead of evaluating the really expensive objective function. During this step, an infill criterion, also known as an \emph{acquisition function}, quantitatively measures how good the optimal solution $\mathbf{x}^*$ is. Then, the optimal solution $\mathbf{x}^*$ and its corresponding objective value, evaluated by the real objective function $f$, will be used to update the Gaussian process model. However, only a single solution can be optimized within one iteration, which prevents to take advantage of parallel computation facilities for expensive simulations. 

A batch acquisition function for single-objective problems, \emph{multiple-point expected improvement} (q-EI), was firstly pointed out, though not developed, by Schonlau \cite{schonlau1997computer} to measure the \emph{expected improvement} (EI) for a batch with $q$ points $(\mathbf{x}^{(1)}, \cdots, \mathbf{x}^{(q)})$ by their corresponding predicted distributions $\hat{Y} = (\hat{y}^{(1)}, \cdots, \hat{y}^{(q)})$.
The predicted distributions $\hat{Y}$
follows a multivariate normal distribution with a mean vector $M=(\mu^{(1)},\cdots, \mu^{(q)})$ and a covariance matrix $\Sigma$, where both $M$ and $\Sigma$ can be estimated by a Gaussian process predictor.
Later, q-EI was well developed by Ginsbourger et al. in \cite{ginsbourger2009metamodeles,Ginsbourger2010,Ginsbourger2012fastqei}, where the explicit formula of q-EI was provided. By maximizing the q-EI, an optimal batch consisting of $q$ points can be searched. This batch criterion is of particular interest in real-world applications, as it allows multiple CPUs to evaluate the $q$ points simultaneously and thus shortens the execution time of Bayesian global optimization. Moreover, the correlation among predictions, which can be reflected by the covariance ($\mathop{\E}[\hat{y}^{(i)}, \hat{y}^{(j)}] - \mathop{\E}[\hat{y}^{(i)}]\mathop{\E}[\hat{y}^{(j)}]|_{i,j=\{ 1,\cdots, q\}, i\neq j}$), are involved in the computation of q-EI~\cite{Ginsbourger2010,Ginsbourger2012fastqei}. The correlation in the computation of a batch criterion can measure the difference affected by the relationships between any two predictions in $\hat{Y}$. In this sense, correlations can theoretically promote the performance of a batch-criterion-based Bayesian global optimization algorithm, as shown in~\cite{Ginsbourger2010}. 

Generalizing the BGO into multi-objective cases, a multi-objective Bayesian global optimization (MOBGO) algorithm builds up independent models for each objective. Various approaches have been studied and can be utilized for parallelization in MOBGO. 
For instance, MOEA/D-EGO proposed in ~\cite{zhang2009expensive} generalizes ParEGO by setting different weights and then performs MOEA/D parallelly to search for the multiple points. 
Ginsbourger et al. \cite{Ginsbourger2010} proposed to approximate $q$-EI by using the techniques of Kriging believer (KB) and Constant liar (CL). Both KB and CL can be directly and easily extended for MOBGO algorithms. 
Bradford et al. proposed to utilize Thompson Sampling on the GP posterior as an acquisition function and to select the batch by maximizing the hypervolume (HV) from the population optimized by NSGA-II. 
Yang et al. proposed to divide an objective space into several sub-spaces, and to search/evaluate the optimal solutions in each sub-space by maximizing the \emph{truncated expected hypervolume improvement} parallelly \cite{yangk2019gecco}.
Gaudrie et al. \cite{Gaudrie2020} proposed to search for multiple optimal solutions in several different preferred regions simultaneously by maximizing the \emph{expected hypervolume improvement} (EHVI) with different reference point settings.
DGEMO proposed in~\cite{dgemo} utilizes the mean function
of GP posterior as the acquisition function, then optimizes the acquisition function by a so-called 'discovery algorithm' in~\cite{discovery}, and then selects the batch based on the diversity information in both decision and objective spaces. MOEA/D-ASS proposed in~\cite{moea/d-ass} utilizes the \emph{adaptive lower confidence bound} as the acquisition function and introduces an adaptive subproblem selection (ASS) to identify the most promising subproblems for further modeling. Recently, the $\epsilon$-greedy strategy, which combines greedy search and a random selection, also shows the efficacy on both single- and multi-objective problems~\cite{de2021asynchronous,gibson2021multi} especially on high-dimensional problems~\cite{greedyisgood,gibson2021multi}. 

However, distinguished from q-EI in the single-objective BGO, none of the parallel techniques mentioned above consider correlations among multiple predictions within a batch in MOBGO. Recently, Daulton et al.~\cite{daulton2020differentiable} proposed the multiple-point EHVI (q-EHVI) that incorporates the correlations among multiple predictions in a single coordinate. However, the correlation is utilized by using the Monte Carlo (MC) method instead of an exact calculation method. Consequently, the approximation error of q-EHVI by the MC method may render MOBGO's performance in optimization processes. Because for an indicator-based optimization algorithm, the optimization results of using an approximation method are not competitive compared with that of exact computation methods~\cite{coello2020evolutionary}. 

This paper focuses on another common acquisition function, namely, \emph{probability of improvement} (PoI)~\cite{Zilinskas92,Jones01}, which is widely used within the frameworks of BGO and MOBGO~\cite{Emmerich2020}. The main contribution of the paper is that we propose five different types of \emph{multiple-point probability of improvement} (q-PoI), and also provide explicit formulas for their exact computations. The remaining parts of this paper are structured as follows: Section 2 introduces the related definitions and the background of multi-objective Bayesian global optimization. Section 3 describes the assumptions and proposes the five different q-PoI, provides both the MC method and explicit formulas to approximate/compute q-PoIs, and analyzes the position-dependent behaviors of q-PoIs w.r.t. standard deviation and coefficient. Section 4 discusses the optimization studies using the five q-PoIs within the MOBGO framework on twenty bio-objective optimization problems.  

\section{Multi-objective Bayesian Global Optimization}
\subsection{Multi-objective Optimization Problems}
A multi-objective optimization (MOO) problem involves multiple objective functions to be minimized simultaneously. A MOO problem can be formulated as:
$$
\operatorname{minimize} \quad  \mathbf{f}(\mathbf{x}):= [{f}_1(\mathbf{x}),{f}_2(\mathbf{x}),\cdots,{f}_m(\mathbf{x}) ]^\top \qquad \text{for} \qquad  \mathbf{x} \in \mathcal{X} \subseteq \mathbb{R}^d 
$$
where $m$ is the number of objective functions, ${f}_i$ stands for the $i$-th objective functions
$f_i: \mathcal{X} \rightarrow \mathbb{R}$, $i = 1, \dots, m$, $\mathcal{X}$ is a decision vector subset. For simplicity, we restrict $\mathcal{X}$ to be a subset of a continuous space, that is $\mathcal{X} \subseteq \mathbb{R}^d$ in this paper. Theoretically, $\mathcal{X}$ can also be a subset of a discrete alphabet $\mathcal{X} \subseteq \{0, 1\}^d$ or even a mixed space, e.g., $\mathcal{X} \subseteq \mathbb{R}^{d_1}\times\mathbb{N}^{d_2}\times \{0, 1\}^{d_3}$, which can be achieved by a heterogeneous metric for the computation of distance function \cite{YangMixedMOBGO2018} or the one-hot strategy encoding \cite{GARRIDOMERCHAN202020} in Gaussian process. In this paper, $d$ denotes the dimension of the search space $\mathcal{X}$. 

\subsection{Gaussian Process}
\label{subsubsec:GPR}
Gaussian process regression is used as the surrogate model in Bayesian Global optimization to approximate the unknown objective function and quantify the uncertainty of a prediction. In this technique, the uncertainty of the objective function $f$ is modeled as a probability distribution of function, which is achieved by posing a prior Gaussian process on it.
We consider $\mathrm{X}=\{\mathbf{x}^{(1)}, \mathbf{x}^{(2)}, \cdots, \mathbf{x}^{(n)}\}$, $\mathbf{x}^{(i)} \in \X$ a set of decision vectors, which are usually obtained by some sampling methods (e.g., the Latin hypercube
sampling~\cite{mckaya1979comparison}) and associated objective function values
\[\BF{\psi} = f(\mathrm{X}) =\big[{f}(\mathbf{x}^{(1)}), {f}(\mathbf{x}^{(2)}), \cdots, {f}(\mathbf{x}^{(n)})\big]^{\top}.\]
Then the objective function can be modeled as a centered Gaussian process (GP) prior to an unknown constant trend term $\mu$ (to be estimated):
\begin{equation}
    f \sim \GP(0, k(\cdot, \cdot))
\end{equation}
where $k\colon \X \times \X \rightarrow \mathbb{R}$ is a positive definite function (a.k.a.~kernel) that computes the autocovariance of the process, namely $k(\BF{x}, \BF{x}^{\prime}) = \Cov (f(\BF{x}),f(\BF{x}^{\prime}))$. The most well-known kernel is the so-called Gaussian kernel (a.k.a~radial basis function. (RBF))~\cite{books/daglib/0039031}:
\begin{equation}
    k(\BF{x},\BF{x}') = \sigma^2\exp\left(-\sum_{i=1}^{d}\frac{(x_i - x_i')^2}{2\theta_i^2} \right),
    \label{term:guassianKernel}
\end{equation}
where $\sigma^2$ models the variance of function values at each point and $\theta_i$ are kernel parameters representing variables' importance, they are typically estimated from data using the maximum likelihood principle. 
The optimal $\boldsymbol{\theta} = (\theta_1^{opt}, \cdots, \theta_d^{opt})$ of GP models can be optimized by any continuous optimization algorithm.

For an unknown point $\BF{x}$, a Bayesian inference yields the posterior distribution of $f$, i.e., $\conProb{f(\BF{x})}{\BF{\psi}} \propto p(\BF{\psi}\mid f(\BF{x}))p(f(\BF{x}))$.
Note that this posterior probability is also a conditional probability 
due to the fact that $f(\BF{x})$ and $\BF{\psi}$ are jointly Gaussian, namely,
$$\begin{bmatrix}
    f(\BF{x}) \\
    \BF{\psi}
\end{bmatrix} \sim \bm{\mathcal{N}}\left(\mathbf{0}, \begin{bmatrix}
    \sigma^2 & \BF{k}^\top \\
    \BF{k} & \BF{K}
\end{bmatrix}\right),$$
where $\BF{K}_{ij} = k(\BF{x}^{(i)}, \BF{x}^{(j)})$ and $\BF{k}(\BF{x}) = (k(\BF{x}, \BF{x}^{(1)}),k(\BF{x}, \BF{x}^{(2)}),\ldots,k(\BF{x}, \BF{x}^{(n)}))^\top$.
Conditioning on $\BF{\psi}$, we obtain the posterior of $f$:
\begin{equation}
    f(\BF{x}) \;|\; \BF{\psi} \sim \bm{\mathcal{N}}\left(\BF{k}^\top\BF{K}^{-1}\BF{\psi}, \sigma^2 - \BF{k}^\top\BF{K}^{-1}\BF{k}\right). \label{eq:SK-conditioning}
\end{equation}

Given this posterior, it is obvious that the best-unbiased predictor of $Y$ is the posterior mean, i.e., $\mu= \BF{k}^\top\BF{K}^{-1}\BF{\psi}$, which is also the Maximum a Posterior Probability (MAP) estimation. The MSE of $\mu$ is $s^2 = \E\{\mu-f\}^2 =\sigma^2 - \BF{k}^\top\BF{K}^{-1}\BF{k}$, which is also the posterior variance.

Moreover, to see the covariance structure of the posterior process, we can consider $q$ unknown points $\BF{x}^{\prime(1)},\BF{x}^{\prime(2)}, \cdots, \BF{x}^{\prime(q)}\in\X$:
$$
\begin{bmatrix}
    f(\BF{x}^{\prime(1)}) \\
    f(\BF{x}^{\prime(2)}) \\
    \vdots \\
    f(\BF{x}^{\prime(q)}) \\
    \BF{\psi}
\end{bmatrix} \sim \bm{\mathcal{N}}\left(\BF{0}, \begin{bmatrix}
    \sigma^2 & k(\BF{x}^{\prime(1)}, \BF{x}^{\prime(2)}) & \cdots & k(\BF{x}^{\prime(1)}, \BF{x}^{\prime(q)})  & \BF{k}_1^\top \\
     k(\BF{x}^{\prime(2)}, \BF{x}^{\prime(1)}) & \sigma^2 & \cdots & k(\BF{x}^{\prime(2)}, \BF{x}^{\prime(q)}) & \BF{k}_2^\top \\
     \vdots & \vdots &\ddots &\vdots &\vdots \\
        k(\BF{x}^{\prime(q)}, \BF{x}^{\prime(1)}) 
    &   k(\BF{x}^{\prime(q)}, \BF{x}^{\prime(2)})
    &   \cdots 
    &   \sigma^2 
    &   \BF{k}_q^\top \\ 
    \BF{k}_1 & \BF{k}_2 & \dots & \BF{k}_q  & \BF{K}
\end{bmatrix}\right),
$$
in which $\BF{k}_i = \BF{k}(\BF{x}^{(i)}), i\in\{1, 2, \cdots, q\}$.  After conditioning on $\BF{\psi}$, we obtained the following distribution:
\begin{align*}
\big[f(\BF{x}^{\prime(1)}, f(\BF{x}^{\prime(2)}), \cdots, f(\BF{x}^{\prime(q)})\big]^\top
\vert \;\BF{\psi} \sim \bm{\mathcal{N}}(\BF{\mu},
\Sigma). \numberthis
\label{gp:posteriorMatrix}
\end{align*}
, where $\BF{\mu} =
[   \BF{k}_1^\top\BF{K}^{-1}\BF{\psi}, 
    \BF{k}_2^\top\BF{K}^{-1}\BF{\psi}, \cdots,  
    \BF{k}_q^\top\BF{K}^{-1}\BF{\psi} ]^\top$
and 
$$\Sigma = \begin{bmatrix}
    \sigma^2 - \BF{k}_1^\top\BF{K}^{-1}\BF{k}_1 
    & k(\BF{x}^{\prime(1)}, \BF{x}^{\prime(2)}) - \BF{k}_1^\top\BF{K}^{-1}\BF{k}_2 &\cdots 
    & k(\BF{x}^{\prime(1)}, \BF{x}^{\prime(q)}) - \BF{k}_1^\top\BF{K}^{-1}\BF{k}_q\\
    k(\BF{x}^{\prime(2)}, \BF{x}^{\prime(1)})- \BF{k}_2^\top\BF{K}^{-1}\BF{k}_1 
    & \sigma^2 - \BF{k}_2^\top\BF{K}^{-1}\BF{k}_2 & \cdots 
    & k(\BF{x}^{\prime(2)}, \BF{x}^{\prime(q)}) - \BF{k}_2^\top\BF{K}^{-1}\BF{k}_q\\
    \vdots & \vdots & \ddots & \vdots \\
    k(\BF{x}^{\prime(q)}, \BF{x}^{\prime(1)})- \BF{k}_q^\top\BF{K}^{-1}\BF{k}_1 
    & k(\BF{x}^{\prime(q)}, \BF{x}^{\prime(2)})- \BF{k}_q^\top\BF{K}^{-1}\BF{k}_2  & \cdots 
    &  \sigma^2 - \BF{k}_q^\top\BF{K}^{-1}\BF{k}_q\\
\end{bmatrix}.$$

In this posterior formulation, it is clear to see that the covariance at two different arbitrary locations ($\BF{x}'^{(1)}$ and $\BF{x}'^{(2)}$) is expressed in the cross-term of the posterior covariance matrix as follows:
\begin{align}
    k'(\BF{x}'^{(1)},\BF{x}'^{(2)}) &\coloneqq \Cov \{f(\BF{x}'^{(1)}), f(\BF{x}'^{(2)}) \mid \BF{\psi}\} = k(\BF{x}'^{(1)}, \BF{x}'^{(2)})- \BF{k}(\BF{x}'^{(1)})^\top\BF{K}^{-1}\BF{k}(\BF{x}'^{(2)}), \label{eq:SK-post-cov}
\end{align}
where $k'$ is additional information and is essential to design a multiple-point infill criterion (a.k.a. acquisition function) for parallel techniques in BGO algorithms, e.g., the multi-point expected improvement (q-EI)~\cite{ChevalierG13}. However, $k'$ is not utilized to compute a multi-objective acquisition function in terms of an exact computational approach\footnote{A recent work in \cite{daulton2020differentiable} utilizes $\Sigma$ 
using MC integration with samples from the joint posterior
in terms of sampling by an MC method to approximate multiple-point \emph{Expected Hypervolume Improvement}, instead of an exact computational method.}. 

\subsection{Structure of MOBGO}
Similar to single-objective Bayesian global optimization, MOBGO starts with sampling an initial design of experiment (DoE) with a size of $\eta$ (line 2 in Alg. \ref{alg:mobgo}), $\mathrm{X}=\{\mathbf{x}^{(1)},\mathbf{x}^{(2)},$ $\ldots, \mathbf{x}^{(\eta)}\} \subseteq \X$. DoE is usually generated by simple random sampling or Latin Hypercube Sampling~\cite{daglib/0022270}. By using the initial DoE, $\mathrm{X}$ and its corresponding objective values, $\mathbf{Y}=\{\mathbf{f}(\mathbf{x}^{(1)}),\mathbf{f}(\mathbf{x}^{(2)}),\cdots,\mathbf{f}(\mathbf{x}^{(\eta)})\}\subseteq \mathbb{R}^{m\times\eta}$ (line 3 in Alg. \ref{alg:mobgo}), surrogate models $\M_i$ can be constructed to describe the probability distribution of the objective function $f_i$ conditioned on the initial evidence $\mathbf{Y}$, namely $\Pr ( f_i \; | \; \mathrm{X}, \mathbf{Y}_i)$ (line 4 in Alg. \ref{alg:mobgo}), where $\mathbf{Y}_i$ represents $f_i(\mathrm{X})$. Between two surrogate models, it is widely assumed that model $\mathcal{M}_i$ is independent of $\mathcal{M}_j$\footnote{In~\cite{BoyleF04}, a so-called dependent Gaussian process was proposed to learn the correlation between different processes. The research in this paper only considers independent Gaussian Processes.}, $\forall i \neq j, i,j \in \{ 1, 2, \cdots, m \} $. 

\begin{algorithm}[!htbp]
    \normalsize
	\caption{ Multi-objective Bayesian Global Optimization}\label{alg:mobgo}
		{\bf MOBGO}{$(\mathbf{f},\af,\X,\mathcal{M},\gamma, \eta, T_c)$} \\
		\tcc{$\BF{f}$: objective functions, $\af$: acquisition function, $\X$: search space, $\gamma$: parameters of $\af$, $\mathcal{M}$: a surrogate model to train, $T_c$: maximum number of function evaluations}{
		{Generate the initial DoE: $\mathrm{X} = \{\BF{x}^{(1)},\BF{x}^{(2)},\ldots,\BF{x}^{(\eta)}\} \subset \X$}\;
		{Evaluate $\mathbf{Y} \leftarrow \left\{\BF{f}(\BF{x}^{(1)}),\BF{f}(\BF{x}^{(2)}),\ldots,\BF{f}(\BF{x}^{(\eta)})\right\}$}\;
		{Train surrogate models $\mathcal{M}_i$ on $(\mathrm{X},\mathbf{Y}_i)$, where $i=1, \cdots, m$}\;
		$g \leftarrow \eta$\;
		\While{$g<T_c$}{
	 		{$\mathrm{X}^* \leftarrow \argmax\af(\mathrm{X}';\mathbfcal{M},\gamma)$, where $\mathbfcal{M}=\{\M_1, \cdots, \M_m \}$ and $\mathrm{X}' = \{ \mathbf{x}^{\prime(1)}, \mathbf{x}^{\prime(2)}, \cdots, \mathbf{x}^{\prime(q)} \} \in \X$}\;
	 		{$\mathbf{Y}^* \leftarrow \mathbf{f}(\mathrm{X}^*)$}\;
	 		{$\mathrm{X} \leftarrow \mathrm{X} \cup \{\mathrm{X}^*\}$}\;
	 		{$\mathbf{Y} \leftarrow \mathbf{Y} \cup \{\mathbf{Y}^*\}$}\;
	 		{Re-train the surrogate models $\mathcal{M}_i$ on $(\mathrm{X},\mathbf{Y}_i)$, where $i=1, \cdots, m$}\;
	 		{$g \leftarrow g+q$}
		}
		}
\end{algorithm}

Once the surrogate models $\mathbfcal{M}$ are constructed, MOBGO enters the main loop until a stopping criterion is fulfilled\footnote{In this paper, we restrict the stopping criterion as the number of iterations.}, as shown in Alg. \ref{alg:mobgo} from line 6 to line 12. The main-loop starts with searching for a decision vector set $\mathrm{X}'$ in the search space $\X$ by maximizing the acquisition function $\af$ with parameters of $\gamma$ and surrogate models $\mathbfcal{M}$ (line 7 in Alg. \ref{alg:mobgo}). $q$ represents the batch size or the number of possible decision vectors in $\mathrm{X}'$. The value of $q$ is determined by the acquisition function's theoretical definition and properties $\af$. For the acquisition functions of $\poi$ and $\ehvi$ in multi-objective cases, $q$ can only be set as $1$ by using an exact computation of the $\af$. Other approaches, like KB, CL\footnote{The details of KB and CL can be found in Algorithm 1, and Algorithm 2 in \cite{Ginsbourger2010}.}, and other methods in~\cite{yang2019gecco,Gaudrie2020}, theoretically compute exact $\af$ by using $q=1$. Therefore, these methods can not utilize the covariance among multiple predictions $k'$ in Eq. (\ref{eq:SK-post-cov}) to guide the optimization processes at line 7 in Alg. \ref{alg:mobgo}. A single-objective optimization algorithm searches for the optimal decision vector set $\X^*$. Theoretically, any single-objective optimization algorithm can be utilized, e.g., genetic algorithm (GA), particle swarm optimization (PSO), ant colony optimization algorithm (ACO), covariance matrix adaptation evolution strategy (CMA-ES), and even gradient-ascent algorithms~\cite{YANG2019945}. The optimal decision vector set $\mathrm{X}^*$ will then be evaluated by the 'true' objective functions $\BF{f}$. When $q>1$, parallelization techniques can be utilized to evaluate multiple solutions in $\mathrm{X}^*$. The surrogate models $\mathbfcal{M}$ will be retrained by the updated $\mathrm{X}$ and $\mathbf{Y}$. 

\section{Multiple-point Probability of Improvement}
\subsection{Related Definitions}
\emph{Pareto dominance}, or briefly \emph{dominance}, is an ordering relationship on a set of potential solutions. \emph{Dominance} is defined as follows:
\begin{definition}[Dominance -- $\prec$ \textnormal{\cite{CoelloCoello2011}}]%
Given two decision vectors $\mathbf{x}^{(1)},\mathbf{x}^{(2)} \in \mathbb{R}^d $ and their corresponding objective values $\mathbf{y}^{(1)}=\mathbf{f}(\mathbf{x}^{(1)})$, $\mathbf{y}^{(2)}=\mathbf{f}(\mathbf{x}^{(2)})$ in a minimization problem, it is said that $\mathbf{y}^{(1)}$ dominates $\mathbf{y}^{(2)}$, being represented by $\mathbf{y}^{(1)} \prec \mathbf{y}^{(2)}$, iff \enskip $\forall i \in \{ 1, 2, \cdots, m \}: {f}_i(\mathbf{x}^{(1)}) \leq {f}_i(\mathbf{x}^{(2)})$ and $\exists j \in \{ 1, 2, \cdots, m \}: {f}_{j}(\mathbf{x}^{(1)}) < {f}_{j}(\mathbf{x}^{(2)})$.
\label{def:dominance}
\end{definition}

\begin{definition}[Non-Dominated Space of a Set\textnormal{~\cite{yang2019efficient}}]
Let $\pfa$ be a subset of $\mathbb{R}^m$ and let a reference point $\mathbf{r} \in \mathbb{R}^m$ be such that $\forall \mathbf{p} \in \pfa: \mathbf{p} \prec \mathbf{r}$. The non-dominated space of $\pfa$ with respect to $\mathbf{r}$, denoted as $\mbox{ndom}(\pfa)$, is then defined as:
\begin{align*}
\mbox{ndom} (\pfa, \mathbf{r}):= \{ \mathbf{y} \in \mathbb{R}^m\, |\, \mathbf{y} \prec \mathbf{r} \mbox{ and } \not \exists \mathbf{p} \in \pfa \mbox{ such that } \mathbf{p} \prec \mathbf{y} \, \} \numberthis
\end{align*} 
\label{def:non_dominated_space}
\end{definition}
Note that a reference point $\mathbf{r}$ shall be chosen so that every possible solution dominates $\mathbf{r}$. A reference point $\mathbf{r}$ should avoid being an infinity vector in \emph{hypervolume-based} indicators. In this paper, $\mathbf{r}$ is an infinity vector for PoI and its variants, as the maximum value is bounded by $1$.

\begin{definition}[Probability of Improvement\textnormal{~\cite{Zilinskas92,Emmerich2020}}]
Given the predictions $\hat{\vy} = (\hat{y_1}, \hat{y_2}, \\ \cdots, \hat{y_m})^\top$\footnote{Here, $\hat{y}_i \sim \mathcal{N}(\mu_i, s_i), i \in {1, 2, ... ,m}$.} with the parameters of the multivariate predictive distribution $\boldsymbol\mu = (\mu_1,\mu_2, \\ \cdots, \mu_m)^\top$, $\boldsymbol{s} = (s_1, s_2, \cdots, s_m)^\top$ and the Pareto-front approximation set $\pfa$, the \emph{Probability of Improvement} (PoI) is defined as:
\begin{align*}
\poi(\boldsymbol\mu, \boldsymbol{s}, \pfa) :&= \int_{\mathbb{R}^m} \mathrm{I}(\mathbf{\hat{y}} \mbox{ impr } \pfa )  \boldsymbol{\pdf}_{\boldsymbol\mu,\boldsymbol{s}}(\mathbf{\hat{y}}) d\mathbf{\hat{y}} \text{ and }
\mathrm{I}(v) =
\left\{ 
\begin{array}{ll} 
1 ,&  v= \mathrm{true}\\
0, &  v= \mathrm{false}
\end{array} 
\right.
\numberthis
\label{def:poi}
\end{align*}
where $\boldsymbol{\pdf}_{\boldsymbol{\mu}, \boldsymbol{s}}$ is the multivariate independent normal distribution with the mean values $\boldsymbol{\mu} \in \mathbb{R}^m$ and the standard deviations $\boldsymbol{s} \in \mathbb{R}^m_+$. Here $\mathbf{\hat{y}} \mbox{ impr } \pfa$ represents $\hat{\mathbf{y}}\in ndom(\pfa)$.
\end{definition}

\subsection{Assumptions in q-PoI}
Definition \ref{def:poi} can be generalized to multiple-point \poi (q-PoI), which is a cumulative probability of a batch $\hat{\mathbf{Y}} = \{ \mathbf{\hat{y}}^{(1)}, \mathbf{\hat{y}}^{(2)}, \cdots, \mathbf{\hat{y}}^{(q)} \}$ in the whole $ndom(\pfa)$, where $\hat{\mathbf{y}}^{(i)} = (\hat{y}_1^{(i)}, \cdots, \hat{y}_m^{(i)} )^\top|_{i=\{ 1,\cdots, q \}}$. In this section, five q-PoIs are defined accordingly by generalizing the single point \poi in Eq. \eqref{def:poi}.

Like the assumption in single-point \poi in Eq. (\ref{def:poi}), each objective is also assumed to be independent of the other in q-PoI. However, the correlation of multiple points in a particular objective, which is available in the posterior distributions of Gaussian processes, will be considered in q-PoI. That is to say:
Suppose a prediction $y_i^{(j)} \sim \mathcal{N}(\mu_i^{(j)},s_i^{(j)})$, where $i \in \mathbb{M}:=\{1,2,\cdots,m\}$ and $j\in \mathbb{Q}:=\{1,2,\cdots,q\}$, we consider that: 
\begin{enumerate}
    \item $y_i^{(j)} \notindependent y_i^{(jj)}:= \Cov(y_i^{(j)}, y_i^{(jj)}) \neq 0,  \qquad \forall i \in  \mathbb{M}$ and $\forall j, jj \in  \mathbb{Q} |_{j \neq jj}$ 
    \item $y_i^{(j)} \independent y_{ii}^{(jj)} := \Cov(y_i^{(j)},y_{ii}^{(jj)})=0,  \qquad  \forall i, ii \in  \mathbb{M}|_{i \neq ii}$ and $\forall j, jj \in  \mathbb{Q}|_{j\neq jj}$ 
\end{enumerate}

For simplicity, we use $M$ and $\boldsymbol{\Sigma}$ to denote the predicted $\mu$ matrix and covariance matrices of a batch $\hat{\mathbf{Y}}= \{ \mathbf{\hat{y}}^{(1)} 
,\mathbf{\hat{y}}^{(2)}, \cdots
\mathbf{\hat{y}}^{(q)}
\}
$, respectively. They are defined as: 
\begin{align*}
    & M_{m \times q} = 
    \begin{bmatrix}
    \mu_1^{(1)} & \cdots    & \mu_1^{(q)} \\
    \vdots      & \ddots    & \vdots      \\
    \mu_m^{(1)} & \cdots    & \mu_m^{(q)}
    \end{bmatrix}
    = [\boldsymbol{\mu}_1, \cdots, \boldsymbol{\mu}_m]^\top
    = [\boldsymbol{\mu}^{(1)}, \cdots, \boldsymbol{\mu}^{(q)}] ,\\
    & \text{ where } 
    \boldsymbol{\mu}_i = (\mu_i^{(1)}, \cdots, \mu_i^{(q)})^\top
    \text{ and } 
    \boldsymbol{\mu}^{(j)}= (\mu_1^{(j)}, \cdots, \mu_m^{(j)})^\top,
    \numberthis\\
    & \boldsymbol{\Sigma} = \{\Sigma_1, \cdots, \Sigma_m\} \text{ where }
    \Sigma_{i} = 
    \begin{bmatrix}
    \Cov(s_i^{(1)},s_i^{(1)})   & \cdots
    & \Cov(s_i^{(1)},s_i^{(q)}) \\
    \vdots  & \ddots   & \vdots \\
    \Cov(s_i^{(1)},s_i^{(q)})   & \cdots
    & \Cov(s_i^{(q)},s_i^{(q)}) 
    \end{bmatrix} \numberthis
    \label{eq:Simga_Matrix}
\end{align*}
In Eq. \eqref{eq:Simga_Matrix}, $\Cov(s_i^{(j)},s_i^{(j)}) = (s_i^{(j)})^2$, $i=1, 2, \cdots, m$ and $j=1, 2, \cdots, q$. Moreover, to simplify, $\boldsymbol{\Sigma}$ represents the set of covariance matrices of $q$ predictions over $m$ objectives/coordinates ($\Sigma_i|_{i\in\mathbb{M}}$) with a size of $m \times q \times q$. If only the diagonal elements in $\Sigma_i|_{i\in \mathbb{M}}$ are considered. In the other words, the correlation is not considered. $\boldsymbol{\Sigma}$ can be reduced from a $m\times q \times q$ matrix into a $m \times q$ matrix, and noted as: 
\begin{align*}
    & \Lambda =
    \begin{bmatrix}
    s_1^{(1)} & \cdots    & s_1^{(q)} \\
    \vdots      & \ddots    & \vdots      \\
    s_m^{(1)} & \cdots    & s_m^{(q)}
    \end{bmatrix}
    = [\boldsymbol{s}^{(1)}, \cdots, \boldsymbol{s}^{(q)}] \enskip
    \text{where} 
    \left\{
    \begin{array}{ll}
                  \boldsymbol{s}^{(i)}=[{s}_1^{(i)},  \cdots, {s}_m^{(i)}]^\top \\
                  i \in \{1,\cdots, q\}
    \end{array}
    \right. 
\end{align*}
\emph{\textbf{Remark:}} Notice that each element in $\Lambda$ is a standard deviation of a normal distribution. It differs from $\boldsymbol{\Sigma}$ in which each element is a (co-)variance matrix.

\begin{example}
In Fig. \ref{fig:multi-pdf}, a Pareto-front approximation is composed of $\mathbf{y}^{(1)}=(3,1),$ $\mathbf{y}^{(2)}=(2,1.5)$ and $\mathbf{y}^{(3)}=(1,2.5)$ and a batch $\hat{\mathbf{Y}}=\{ \hat{\mathbf{y}}^{(1)}, \hat{\mathbf{y}}^{(2)} \}$, of which parameters $\boldsymbol{\mu}$ and $\BF{s}$ are: $\mu_1^{(1)}=1.5, \mu_2^{(1)}=0.5, \mu_1^{(2)}=2.5, \mu_2^{(2)}=0$ and $s_1^{(i)}=0.6, s_2^{(i)}=0.7, i\in\{1,2\}$. The orange spheres are the multivariate normal distributions of $\hat{\mathbf{y}}^{(1)}$ and $\hat{\mathbf{y}}^{(2)}$.
\begin{figure}
    \centering
    \includegraphics[width=0.8\textwidth]{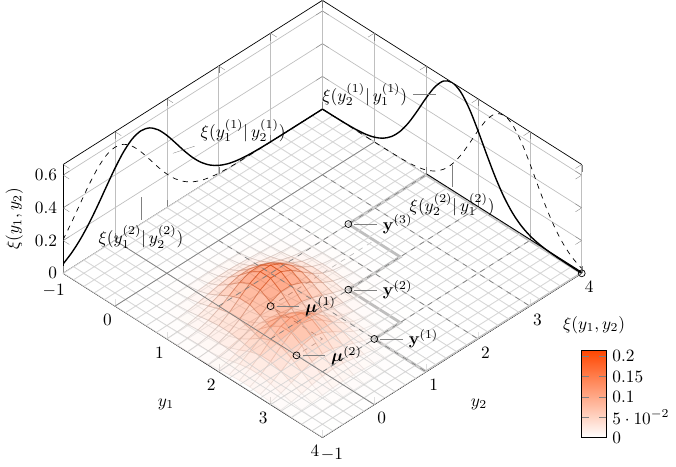}
    \caption{Landscape of multi-joint normal distributions.}
    \label{fig:multi-pdf}
\end{figure}
\end{example}

\subsection{Definitions of Five Proposed q-PoIs in Different Cases}
\label{subsec:kpoi_definitions}
The motivation for introducing the following q-PoI variants is based on different perspectives on designing the improvement function $\mathrm{I}(.)$. The first straightforward idea is to guarantee that \emph{all} points in the batch can improve $\pfa$ simultaneously. This idea formulates the concept of $\kpoi_{\mbox{all}}$. Similarly, the idea behind $\kpoi_{\mbox{one}}$ is to make sure \emph{at least one} point in the batch size can improve $\pfa$. The idea of $\kpoi_{\mbox{best}}$ and $\kpoi_{\mbox{worst}}$ is to ensure that the \emph{best} or \emph{worst} projected values (from all points onto each dimension) can improve $\pfa$. In $\kpoi_{\mbox{mean}}$, the average PoI value of each point in a batch is computed.  
In the following, we present five different q-PoIs under concern.
\begin{definition}
\label{def:kpois}
Given $m\times q$ multivariate normal distributions  $\hat{\boldsymbol{Y}} = \{ \hat{\vy}^{(1)},
\hat{\vy}^{(2)},\cdots, \hat{\vy}^{(q)}\}$ with a mean matrix $M = [\boldsymbol{\mu}^{(1)}, \cdots, \boldsymbol{\mu}^{(q)}]^\top$ and covariance matrices $\boldsymbol{\Sigma} =\{ \Sigma_1, \cdots, \\ \Sigma_{m} \}$, where a root of diagonal elements in ${\Sigma_i}|_{i \in \mathbb{M}}$ composes a standard deviation matrix $\Lambda= [\boldsymbol{s}^{(1)}, \cdots,  \boldsymbol{s}^{(q)}]^\top$. For a Pareto-front approximation set $\pfa$, we define the following PoIs for the batch $\hat{\boldsymbol{Y}}$: 
\begin{align*}
&\kpoi_{\mbox{all}}(M, \boldsymbol{\Sigma}, \pfa) 
:= 
\int_{\mathbb{R}^{m \times q}} \mathrm{I}(\hat{\boldsymbol{y}}^{(1)} 
\mbox{ impr } \pfa ) \cap 
\cdots \cap \mathrm{I}(\hat{\boldsymbol{y}}^{(q)} \mbox{ impr } \pfa ) 
\boldsymbol{\pdf}_{M, \boldsymbol{\Sigma}}(\hat{\mathbf{Y}}) d\hat{\mathbf{Y}} \qquad
\numberthis
\label{def:kpoi_for_all}\\
&\kpoi_{\mbox{one}}(M, \boldsymbol{\Sigma}, \pfa) 
:= 
\int_{\mathbb{R}^{m \times q}} \mathrm{I}(\hat{\boldsymbol{y}}^{(1)} \mbox{ impr } \pfa ) \cup 
\cdots \cup \mathrm{I}(\hat{\boldsymbol{y}}^{(q)} \mbox{ impr } \pfa ) 
\boldsymbol{\pdf}_{M, \boldsymbol{\Sigma}}(\hat{\mathbf{Y}}) d\hat{\mathbf{Y}} \qquad
\numberthis
\label{def:kpoi_for_one}\\
&\kpoi_{\mbox{best}}(M, \boldsymbol{\Sigma}, \pfa) 
:= 
\int_{\mathbb{R}^{m \times q}} \mathrm{I}(\mbox{DisC}(\hat{\boldsymbol{Y}}) \mbox{ impr } \pfa ) 
\boldsymbol{\pdf}_{M, \boldsymbol{\Sigma}}(\hat{\mathbf{Y}}) d\hat{\mathbf{Y}} 
\numberthis
\label{def:kpoi_for_best} \\
&\kpoi_{\mbox{worst}}(M, \boldsymbol{\Sigma}, \pfa) 
:= 
\int_{\mathbb{R}^{m \times q}} \mathrm{I}(\mbox{ConJ}(\hat{\boldsymbol{Y}}) \mbox{ impr } \pfa )
\boldsymbol{\pdf}_{M, \boldsymbol{\Sigma}}(\hat{\mathbf{Y}}) d\hat{\mathbf{Y}} 
\numberthis
\label{def:kpoi_for_worst} \\
&\kpoi_{\mbox{mean}}(M, \boldsymbol{\Sigma}, \pfa) 
:= \frac{1}{q}\sum_{i=1}^{q}
\int_{\mathbb{R}^m} \mathrm{I}(\hat{\mathbf{y}}^{(i)} \mbox{ impr } \pfa ) 
\boldsymbol{\pdf}_{\boldsymbol{\mu}^{(i)}, \boldsymbol{s}^{(i)}}(\hat{\mathbf{y}}^{(i)}) d\hat{\mathbf{y}}^{(i)} \qquad
\numberthis
\label{def:kpoi_for_mean}
\end{align*}
where $\boldsymbol{\pdf}_{{M}, \boldsymbol{\Sigma}}$ is the multivariate normal distribution, $\mbox{DisC}(\hat{\boldsymbol{Y}}) = (\hat{\boldsymbol{Y}}_1 \lor \hat{\boldsymbol{Y}}_2 \lor \cdots \lor \hat{\boldsymbol{Y}}_m) = \mbox{max}(\hat{\boldsymbol{Y}}_1,\hat{\boldsymbol{Y}}_2, \cdots, \hat{\boldsymbol{Y}}_m)^\top$,  $\mbox{ConJ}(\hat{\boldsymbol{Y}}) = (\hat{\boldsymbol{Y}}_1 \land \hat{\boldsymbol{Y}}_2 \land  \cdots \land  \hat{\boldsymbol{Y}}_m) =
\mbox{min}(\hat{\boldsymbol{Y}}_1,\hat{\boldsymbol{Y}}_2, \\ \cdots, \hat{\boldsymbol{Y}}_m)^\top$, $\hat{\mathbf{Y}}_i = (\hat{{y}}_i^{(1)}, \cdots, \hat{{y}}_i^{(q)})|_{ i\in \mathbb{M}}$, and $\mathrm{I}(\cdot)$ is again the indicator function defined in Eq. \eqref{def:poi}.
\end{definition}
Note that correlations in Eq. \eqref{def:kpoi_for_mean} are exclusive of marginal probability density functions, and only the diagonal elements $\Lambda$ in $\Sigma_{i}|_{i\in \mathbb{M}}$ are used here. 

The required condition in $\kpoi_{\mbox{best}}$ is sufficient for the condition in $\kpoi_{\mbox{all}}$, but not necessary. Given a $\hat{\boldsymbol{Y}}$ of parameters $M$, $\boldsymbol{\Sigma}$ and a $\pfa$, if $\kpoi_{\mbox{best}}(M, \boldsymbol{\Sigma}, \pfa)=1$, then $\kpoi_{\mbox{all}}(M, \boldsymbol{\Sigma}, \pfa)=1$. However, $\kpoi_{\mbox{all}}(M, \boldsymbol{\Sigma}, \pfa)=1$ can not guarantee that $\kpoi_{\mbox{best}}(M, \boldsymbol{\Sigma}, \pfa)=1$, due to the stricter condition in $\kpoi_{\mbox{best}}$. Therefore, $\kpoi_{\mbox{best}}$ is theoretically more greedy than $\kpoi_{\mbox{all}}$.

\subsection{Approximation of q-PoI}
\footnote{From this section, we only consider bi-objective case, that is, $m=2$.}
Suppose that we have a Pareto-front approximation set $\pfa$, a mean matrix $M$, two covariance matrices $\Sigma_1$ and $ \Sigma_2$ for each objective/coordinate,
\\ where: 
$M = \begin{bmatrix}
    \mu_1^{(1)} & \mu_1^{(2)}\\
    \mu_2^{(1)} & \mu_2^{(2)}
    \end{bmatrix}
$ and 
$\Sigma_i = 
    \begin{bmatrix}
    s_i^{(1)^2}                                    & \Cov(s_i^{(1)},s_i^{(2)}) \\
    \Cov(s_i^{(1)},s_i^{(2)})           & s_i^{(2)^2}
    \end{bmatrix}, i= 1, 2$
    
The parameter of $M$, $\Sigma_1$ and $\Sigma_2$ can compose a predicted batch in a 2-dimensional objective space, where $\hat{\mathbf{Y}} = \big( \hat{\vy}^{(1)}=(\hat{y}_1^{(1)},\hat{y}_2^{(1)}), \hat{\vy}^{(2)}=(\hat{y}_1^{(2)},\hat{y}_2^{(2)})\big)$. That is to say: 
\begin{align*}
\begin{matrix} 
    \hat{y}_1^{(1)} \sim \mathcal{N}(\mu_1^{(1)},s_1^{(1)^2}) \qquad & \independent \qquad  & \hat{y}_2^{(1)} \sim \mathcal{N}(\mu_2^{(1)},s_2^{(1)^2})\\ 
    \notindependent   \qquad & \qquad     & \notindependent \\ 
    \hat{y}_1^{(2)} \sim \mathcal{N}(\mu_1^{(2)},s_1^{(2)^2}) \qquad & \independent \qquad  & \hat{y}_2^{(2)} \sim \mathcal{N}(\mu_2^{(2)},s_2^{(2)^2})  
\end{matrix}
\numberthis
\end{align*}

The Monte Carlo method to approximate \kpoi by using an acceptance-rejection method \cite{hastings1970monte} is illustrated in Algorithm \ref{alg:mc_kpoi}. 
The idea of the MC method is composed of three main steps: 
\begin{enumerate}
	\item Randomly generate two solutions according to the mean matrix $M$ and two covariance matrices $\boldsymbol{\Sigma}$ (line 4 - line 6). 
	\item Dominance check w.r.t different \kpoi definitions in Section \ref{subsec:kpoi_definitions} and update the corresponding counters (line 7 - line 12). 
	\item Calculate \kpoi by computing the average occurrence points that dominate the $\pfa$.
\end{enumerate}

\begin{algorithm}[!h]
\KwIn{
Mean matrix $M = [\boldsymbol{\mu}_1, \boldsymbol{\mu}_2]^\top$, 
covariance matrices $\boldsymbol{\Sigma} = \{\Sigma_1,\Sigma_2\}$,
number of samples $n_{sample}$,
A Pareto-front approximation set $\pfa$.
}
\KwOut{$\kpoi_{\mbox{case}}$ for $\mbox{case}\in \{\mbox{best,\, worst,\, all,\, one,\, mean} \}$
}
$\text{flag}_{\mbox{case}}=0$ for $\mbox{case}\in \{\mbox{best,\, worst,\, all,\, one,\, mean} \}$ \;
\For(\tcc*[f]{Main loop}){$i=1$ \KwTo $n_{sample}$}{
Random generate a vector $\tilde{\vy}_1 = (\tilde{y}_1^{(1)}, \tilde{y}_1^{(2)}) \sim 
\boldsymbol{\mathcal{N}}(M_1, \Sigma_1)$ \;
Random generate a vector $\tilde{\vy}_2 = (\tilde{y}_2^{(1)}, \tilde{y}_2^{(2)}) \sim
\boldsymbol{\mathcal{N}}(M_2, \Sigma_2)$ \; 
Compose the two solutions $\tilde{\vy}^{(1)}=(\tilde{y}_1^{(1)}, \tilde{y}_2^{(1)})^\top$ and $\tilde{\vy}^{(2)}=(\tilde{y}_1^{(2)}, \tilde{y}_2^{(2)})^\top$ \;
$\text{flag}_{\mbox{best}} = \text{flag}_{\mbox{best}} + \mathrm{I}\big( (\mbox{max}(\tilde{\vy}_1),\mbox{max}(\tilde{\vy}_2))^\top \prec \pfa \big)$\;
$\text{flag}_{\mbox{worst}} = \text{flag}_{\mbox{worst}} + \mathrm{I}\big( (\mbox{min}(\tilde{\vy}_1),\mbox{min}(\tilde{\vy}_2))^\top \prec \pfa \big)$\;
$\text{flag}_{\mbox{all}} = \text{flag}_{\mbox{all}} + \mathrm{I} ( \tilde{\vy}^{(1)} \prec \pfa ) \land  \mathrm{I} ( \tilde{\vy}^{(2)} \prec \pfa ) $\;
$\text{flag}_{\mbox{one}} = \text{flag}_{\mbox{one}} +  \mathrm{I} ( \tilde{\vy}^{(1)} \prec \pfa )  \lor  \mathrm{I} ( \tilde{\vy}^{(2)} \prec \pfa ) $\;
$\text{flag}_{\mbox{mean}} = \text{flag}_{\mbox{mean}} + \frac{1}{2}\left(\mathrm{I}(\tilde{\vy}^{(1)} \prec \pfa) +\mathrm{I}(\tilde{\vy}^{(2)} \prec \pfa) \right) $\;
}
$\kpoi_{\mbox{case}} = \text{flag}_{\mbox{case}} / n_{sample}$ for $\mbox{case}\in \{\mbox{best,\, worst,\, all,\, one,\, mean} \}$\; 
\caption{{\bf Monte Carlo approximation for $\kpoi$ of $q=2$} \label{alg:mc_kpoi}}
\end{algorithm}

\subsection{Explicit Computational Formulas for q-PoI based on MCDF}
In this subsection, we present both  explicit computational formulas for the proposed five q-PoIs, and focus on the bi-objective (i.e., $m=2$) optimization problems for $q=2$. All the following formulas assume $q=2$ if there is no special statement.

For consistent lucidity, we briefly describe the partitioning method for bi-objective optimization problems in \cite{yang2019efficient,Michael2016book}.  
For a sorted Pareto-front set $\pfa = \big( \vy^{(1)}, \vy^{(2)}, \cdots, \vy^{(n)} \big)$ by descending order in $f_2$, 
we augment $\pfa$ with two sentinels: $\mathbf{y}^{(0)} = (r_1, -\infty)$ and $\mathbf{y}^{(n+1)} = (-\infty, r_2)$.  Then the non-dominated space of $\pfa$ can be represented by the stripes $S^{(i)}$, which are now defined by:
\begin{align}
S^{(i)} :=& \left(\left( \begin{array}{c}l_1^{(i)}\\l_2^{(i)}\end{array} \right),
\left(\begin{array}{c}u_2^{(i)}\\u_2^{(i)}\end{array}\right) \right) = 
\left(\left(\begin{array}{c}y_1^{(i)}\\-\infty\end{array}\right), \left(\begin{array}{c}y_1^{(i-1)}\\y_2^{(i)}\end{array}\right) \right), i=1, \dots, n+1  
\end{align}

\begin{example}
Suppose a Pareto-front approximation set $\pfa = \{ \vy^{(1)} = (3,1)^\top, \vy^{(2)} =$ $
(2,1.5)^\top, \vy^{(3)} = (1,2.5)^\top \}$ and a reference point $\mathbf{r} = (4,4)^\top$, as shown in Fig. \ref{fig:partition-2d-min}. The sentinels of $\pfa$ are $\vy^{(0)} = (4, -\infty)^\top$ and $\vy^{(4)} = (-\infty, 4)^\top$. Each stripe $S_j$ can be represented by a lower bound point $(l_1^{(j)},l_2^{(j)})^\top$ and by a upper bound point $(u_1^{(j)},u_2^{(j)})^\top$. 
\end{example}
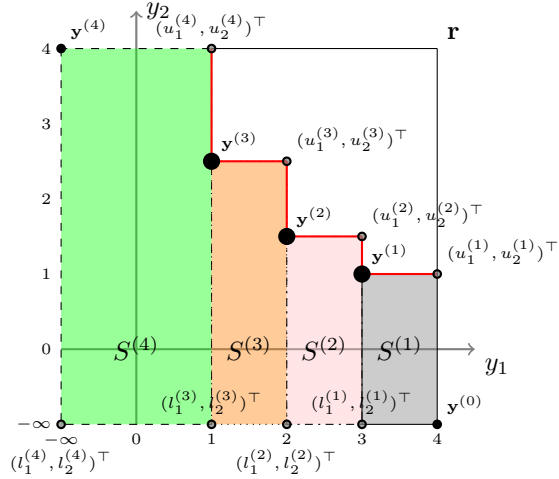
\begin{figure}
    \centering
    \begin{tikzpicture}[domain=0:2,thick,fill opacity=.4,draw opacity=1]
	\draw [gray,thick,->] (-1,0) -- (4.5,0);
	\node [below right,fill opacity=1] at (4.5,0) {$y_1$};
	\draw [gray,thick,->] (0,-1) -- (0,4.5);
	\node [right,fill opacity=1] at (0,4.5) {$y_2$};
		
	\draw [thin] (1, 4) -- (4, 4);
	\draw [thin] (4, 1) -- (4, 4);
	
	\draw [fill=green, dashed,thin] (-1,4) rectangle (1,-1);
	\draw [fill=orange, dotted, thin] (1,2.5) rectangle (2,-1);
	\draw [fill=pink,    dashdotted, thin] (2,1.5) rectangle (3,-1);	
	\draw [fill=gray,   thin] (3,1) rectangle (4,-1);
	
	\draw [red, thick] (1, 4) -- (1, 2.5);
	\draw [red, thick] (1, 2.5) -- (2, 2.5);	
	\draw [red, thick] (2, 2.5) -- (2, 1.5);	
	\draw [red, thick] (2, 1.5) -- (3, 1.5);	
	\draw [red, thick] (3, 1.5) -- (3, 1);	
	\draw [red, thick] (3, 1) -- (4, 1);	
	
	\draw [color=black, fill=black,fill opacity=1] (-1, 4) circle (0.05);
	\node [above right, fill opacity=1] at (-1, 4) {\tiny $\mathbf{y}^{(4)}$};	
	\draw [color=black, fill=black,fill opacity=1] (1, 2.5) circle (0.1);
	\node [above right, fill opacity=1] at (1,2.5) {\tiny $\mathbf{y}^{(3)}$};	
	\draw [color=black, fill=black,fill opacity=1] (2, 1.5) circle (0.1);
	\node [above right, fill opacity=1] at (2,1.5) {\tiny $\mathbf{y}^{(2)}$};
	\draw [color=black, fill=black,fill opacity=1] (3, 1) circle (0.1);
	\node [above right, fill opacity=1] at (3,1) {\tiny $\mathbf{y}^{(1)}$};
	\draw [color=black, fill=black,fill opacity=1] (4, -1) circle (0.05);
	\node [above right, fill opacity=1] at (4, -1) {\tiny $\mathbf{y}^{(0)}$};
	
	\draw [color=black, fill=gray, fill opacity=0.7] (-1, -1) circle (0.05);
	\node [below, fill opacity=1] at (-1, -1.2) {\tiny $(l_1^{(4)},l_2^{(4)})^\top$};
	\draw [color=black, fill=gray, fill opacity=0.7] (1, 4) circle (0.05);
	\node [above, fill opacity=1] at (1, 4) {\tiny $(u_1^{(4)},u_2^{(4)})^\top$};
	
	\draw [color=black, fill=gray, fill opacity=0.7] (1, -1) circle (0.05);
	\node [above, fill opacity=1] at (1, -1) {\tiny $(l_1^{(3)},l_2^{(3)})^\top$};
	\draw [color=black, fill=gray, fill opacity=0.7] (2, 2.5) circle (0.05);
	\node [above right, fill opacity=1] at (2, 2.5) {\tiny $(u_1^{(3)},u_2^{(3)})^\top$};
	
	\draw [color=black, fill=gray, fill opacity=0.7] (2, -1) circle (0.05);
	\node [below , fill opacity=1] at (2, -1.2) {\tiny $(l_1^{(2)},l_2^{(2)})^\top$};
	\draw [color=black, fill=gray, fill opacity=0.7] (3, 1.5) circle (0.05);
	\node [above right, fill opacity=1] at (3, 1.5) {\tiny $(u_1^{(2)},u_2^{(2)})^\top$};
	
	\draw [color=black, fill=gray, fill opacity=0.7] (3, -1) circle (0.05);
	\node [above , fill opacity=1] at (3, -1) {\tiny $(l_1^{(1)},l_2^{(1)})^\top$};
	\draw [color=black, fill=gray, fill opacity=0.7] (4,1) circle (0.05);
	\node [above right, fill opacity=1] at (4,1) {\tiny $(u_1^{(1)},u_2^{(1)})^\top$};
	
	\node [below,fill opacity=1] at (3.5, 0.3) {$S^{(1)}$};
	\node [below,fill opacity=1] at (2.5, 0.3) {$S^{(2)}$};
	\node [below,fill opacity=1] at (1.5, 0.3) {$S^{(3)}$};
	\node [below,fill opacity=1] at (0, 0.3)   {$S^{(4)}$};
	\node [above right,fill opacity=1] at (4, 4) {$\mathbf {r}$};
	\node [below,fill opacity=1] at (0, -1)   {\tiny $0$};
	\node [below,fill opacity=1] at (1, -1)   {\tiny $1$};
	\node [below,fill opacity=1] at (2, -1)   {\tiny $2$};
	\node [below,fill opacity=1] at (3, -1)   {\tiny $3$};
	\node [below,fill opacity=1] at (4, -1)   {\tiny $4$};
	\node [left,fill opacity=1] at (-1, 0)   {\tiny $0$};
	\node [left,fill opacity=1] at (-1, 1)   {\tiny $1$};
	\node [left,fill opacity=1] at (-1, 2)   {\tiny $2$};
	\node [left,fill opacity=1] at (-1, 3)   {\tiny $3$};
	\node [left,fill opacity=1] at (-1, 4)   {\tiny $4$};
	\node [left,fill opacity=1] at (-1, -1)   {\tiny $-\infty$};
	\node [below,fill opacity=1] at (-1, -1)   {\tiny $-\infty$};	
\end{tikzpicture}
    \caption{Partitioning of the integration region into stripes}
    \label{fig:partition-2d-min}
\end{figure}

To calculate q-PoI exactly, having a multivariate cumulative density function of normal distributions (MCDF) is a prerequisite. 

\begin{definition}[\mbox{MCDF} function]\footnote{The definition of MCDF here is for general cases, and all the formulas of q-PoI are restricted to bivariate normal distributions without an additional specific statement in this paper.}
Given a multivariate normal distribution $\mathbf{X} \sim \boldsymbol{\mathcal{N}} (\boldsymbol{\mu}, \Sigma)$ with the parameters of a mean vector $\boldsymbol{\mu} = (\mu^{(1)}, \mu^{(2)}, \cdots, \mu^{(q)})^\top$ and a covariance matrix $\Sigma$ with a size of $q\times q$, the multivariate cumulative probability function of a vector $\mathbf{x}$ is defined as: 
\begin{align*}
\mbox{MCDF}(&\mathbf{x}, \boldsymbol{\mu}, \Sigma)
:= 
\mathbbm{P}(\mathbf{X} \leq \mathbf{x}) 
        = \mathbbm{P}(X_1 \leq {x}_1, X_2 \leq {x}_2, \cdots, X_q \leq {x}_q) \\
&= \int_{-\infty}^{{x}_1} \int_{-\infty}^{{x}_2}\cdots\int_{-\infty}^{{x}_q} \boldsymbol{\pdf}_{\boldsymbol{\mu},\Sigma}(\vy) d\vy 
= \int_{(-\infty, -\infty, \cdots, -\infty)}^{({x}_1, {x}_2, \cdots, {x}_q)}\boldsymbol{\pdf}_{\boldsymbol{\mu},\Sigma}(\vy) d\vy .
\numberthis
\label{def:MCDF}
\end{align*}
\end{definition}

To evaluate the integrals in Eq. (\ref{def:MCDF}), one can adopt multiple numerical estimation methods, including an adaptive quadrature on a transformation of the t-density for bivariate and trivariate distributions \cite{drezner1990bivariatenormalintegral,drezner1994computation,genz2004numerical}, a quasi-Monte Carlo integration algorithm for higher dimensional distributions ($m \leq 4$) \cite{genz1999numerical,genz2002comparison}, and other methods as described in \cite{genz2009computation,botev2017normal}. 
The build-in function of $mvncdf$ in MATLAB is employed for the computation of $\mbox{MCDF}\big( \mathbf{x}, \boldsymbol{\mu}, \Sigma \big)$.
In this case, to calculate the cumulative probability density for Cartesian product domain of the form $(a,b]\times (c,d]$, we introduce the notation $\Gamma(a,b,c,d,\boldsymbol{\mu}, \Sigma)$ as below
\begin{align*}
&\Gamma(a,b,c,d,\boldsymbol{\mu}, \Sigma) :=
\int_{(a,c)}^{(b,d)}\boldsymbol{\pdf}_{\boldsymbol{\mu},\Sigma}(\vy) d\vy \\
=&\left(\int_{(-\infty,-\infty)}^{(b,d)} + 
\int_{(-\infty,-\infty)}^{(a,c)}  - 
\int_{(-\infty,-\infty)}^{(a,d)} 
 - \int_{(-\infty,-\infty)}^{(b,c)} \right) \boldsymbol{\pdf}_{\boldsymbol{\mu},\Sigma}(\vy) d\vy \numberthis
\label{def:GammaFunction}
\end{align*}

Now, we are ready to provide the explicit formulas for each q-PoI based on the MCDF function. Note that some formulas can be trivially extended to a more general case $q>2$. 

\paragraph{\bf $\kpoi_{\mbox{all}}$}
For the sake of simplification, in this case, we again only present the formula for $q=2$. 
\begin{align*}
& \kpoi_{\mbox{all}} (M, \boldsymbol{\Sigma}, \pfa)=\int_{\mathbb{R}^{2 \times 2}} \big( \mathrm{I}(\hat{\mathbf{y}}^{(1)} \mbox{ impr } \pfa ) \big) \cap \big( \mathrm{I}(\hat{\mathbf{y}}^{(2)} \mbox{ impr } \pfa ) \big) \boldsymbol{\pdf}_{M, \boldsymbol{\Sigma}}(\hat{\mathbf{Y}}) d\hat{\mathbf{Y}} \\
&= \int_{{\vy}_1=(-\infty, -\infty)}^{(\infty, \infty)} 
\int_{\hat{\mathbf{y}}_2=(-\infty, -\infty)}^{(\infty, \infty)}
\big( \mathrm{I}(\hat{\mathbf{y}}^{(1)} \mbox{ impr } \pfa ) \big) \cap \big( \mathrm{I}(\hat{\mathbf{y}}^{(2)} \mbox{ impr } \pfa ) \big) \boldsymbol{\pdf}_{M, \boldsymbol{\Sigma}}(\hat{\mathbf{Y}}) d\hat{\mathbf{Y}}  \\
&= \sum_{j=1}^{n+1}\sum_{jj=1}^{n+1} \Big(
\int_{(l_1^{(j)},l_1^{(jj)})}^{(u_1^{(j)}, u_1^{(jj)})}
\boldsymbol{\pdf}_{\boldsymbol{\mu}_1, \Sigma_1}(\hat{\mathbf{Y}}_1) d\hat{\mathbf{Y}}_1 \times
\int_{(l_2^{(j)},l_2^{(jj)})}^{(u_2^{(j)}, u_2^{(jj)})}
\boldsymbol{\pdf}_{\boldsymbol{\mu}_2, \Sigma_2}(\hat{\mathbf{Y}}_2) d\hat{\mathbf{Y}}_2
\Big) \\
&= \sum_{j=1}^{n+1}\sum_{jj=1}^{n+1} \prod_{i=1}^{m=2}\Gamma(l_i^{(j)},u_i^{(j)},l_i^{(jj)},u_i^{(jj)},\boldsymbol{\mu}_i,\Sigma_i) \numberthis
\label{eq:exact_kpoi_all}
\end{align*}
For more general case of $q>2$, the Eq.~\eqref{eq:exact_kpoi_all} can be derived from the same spirit presented above, while integrations need to permute all the combinations of the $q$ entries $\hat{\mathbf{y}}^{(q)} $ over every rectangular stripe in $ndom(\pfa)$. 

\paragraph{\bf $\kpoi_{\mbox{one}}$}
The formula for $q=2$, in this case, is the following
\begin{align*}
& \kpoi_{\mbox{one}} (M, \boldsymbol{\Sigma}, \pfa)=\int_{\mathbb{R}^{2 \times 2}} \big( \mathrm{I}(\hat{\mathbf{y}}^{(1)} \mbox{ impr } \pfa ) \big) \cup \big( \mathrm{I}(\hat{\mathbf{y}}^{(2)} \mbox{ impr } \pfa ) \big) \boldsymbol{\pdf}_{M, \boldsymbol{\Sigma}}(\hat{\mathbf{Y}}) d\hat{\mathbf{Y}} \\
& = \Big( \sum_{j=1}^{q=2}
\int_{\mathbb{R}^2} \mathrm{I}(\hat{\mathbf{y}}^{(j)} \mbox{ impr } \pfa ) \boldsymbol{\pdf}_{\boldsymbol{\mu}^{(j)}, \boldsymbol{s}^{(j)}}(\hat{\mathbf{y}}^{(j)}) d\hat{\mathbf{y}}^{(j)}\Big) - 
\kpoi_{\mbox{all}} (M, \boldsymbol{\Sigma}, \pfa)
\numberthis
\label{eq:exact_kpoi_one}
\end{align*}
Note that the above formula can not be directly extended to the case of $q>2$. However, a more general formula can be derived from the basic calculations by permuting different combinations of the adjoint distribution of $j$ different entries from the set $\{\hat{\mathbf{y}}^{(q)}\}$ for $j=2$ to $j=q$. In Eq. \eqref{eq:exact_kpoi_one}, $\sum_{j=1}^{q=2}
\int_{\mathbb{R}^2} \mathrm{I}(\hat{\mathbf{y}}^{(j)} \mbox{ impr } \pfa ) \boldsymbol{\pdf}_{\boldsymbol{\mu}^{(j)}, \boldsymbol{s}^{(j)}}(\hat{\mathbf{y}}^{(j)}) d\hat{\mathbf{y}}^{(j)}$ can be computed simply by $2\times \kpoi_{\mbox{mean}}$ with $q=2$ in Eq. \eqref{eq:exact_kpoi_mean}.

\paragraph{\bf $\kpoi_{\mbox{best}}$}
According to the definition of $\kpoi_{\mbox{best}}$ in Eq. (\ref{def:kpoi_for_best}) and the property of integration, $\kpoi_{\mbox{best}}$ when $q =2$ can be explicitly calculated as follows: 
\begin{align*}
 \kpoi_{\mbox{best}}& (M, \boldsymbol{\Sigma}, \pfa) = \int_{\mathbb{R}^{2 \times q}} \mathrm{I}(\mbox{DisC}(\hat{\boldsymbol{Y}}) \mbox{impr } \pfa )  
 \boldsymbol{\pdf}_{M, \boldsymbol{\Sigma}}(\hat{\mathbf{Y}}) d\hat{\mathbf{Y}} \\
&= \int_{\hat{\mathbf{Y}}_1=(-\infty, -\infty)}^{(\infty, \infty)} 
\int_{\hat{\mathbf{Y}}_2=(-\infty, -\infty)}^{(\infty, \infty)}
\big( \mathrm{I}( \hat{\mathbf{Y}}_1 \lor \hat{\mathbf{Y}}_2 ) \mbox{ impr } \pfa \big)  
\boldsymbol{\pdf}_{M, \boldsymbol{\Sigma}}(\hat{\mathbf{Y}}) d\hat{\mathbf{Y}} \\
&= \sum_{j=1}^{n+1} \Big(
\left(\int_{(-\infty,l_1^{(j)})}^{(l_1^{(j)},u_1^{(j)})} + \int_{(l_1^{(j)},-\infty)}^{(u_1^{(j)},u_1^{(j)})}\right)
\boldsymbol{\pdf}_{\boldsymbol{\mu}_1, {\Sigma}_1}(\hat{\mathbf{Y}}_1) d\hat{\mathbf{Y}}_1 \times \\
& \quad \qquad \enskip
\left(\int_{(-\infty,l_2^{(j)})}^{(l_2^{(j)},u_2^{(j)})} + \int_{(l_2^{(j)},-\infty)}^{(u_2^{(j)},u_2^{(j)})}\right) 
\boldsymbol{\pdf}_{\boldsymbol{\mu}_2, {\Sigma}_2}(\hat{\mathbf{Y}}_2) d\hat{\mathbf{Y}}_2 
 \Big)\\
&= \sum_{j=1}^{n+1} \prod_{i=1}^{m=2} \Big( \mbox{MCDF}\big( (u_i^{(j)},u_i^{(j)}), \boldsymbol{\mu}_i, \Sigma_i \big) - \mbox{MCDF}\big( (l_i^{(j)},l_i^{(j)}), \boldsymbol{\mu}_i, \Sigma_i \big) \Big)
\numberthis
\label{def:exact_kpoi_best}
\end{align*}

\paragraph{\bf $\kpoi_{\mbox{worst}}$}
Similarly, by using the definition of $\kpoi_{\mbox{worst}}$ in Eq. (\ref{def:kpoi_for_worst}), $\kpoi_{\mbox{worst}}$ when $q = 2$ can also be explicitly computed: 
\begin{align*}
&\kpoi_{\mbox{worst}} (M, \boldsymbol{\Sigma}, \pfa) 
= \int_{\mathbb{R}^{2 \times q}} \big(\mathrm{I}(\mbox{ConJ}(\hat{\boldsymbol{Y}}) \mbox{ impr } \pfa \big)  
\boldsymbol{\pdf}_{M, \boldsymbol{\Sigma}}(\hat{\mathbf{Y}}) d\hat{\mathbf{Y}} \\
&= \sum_{j=1}^{n+1} \Big(
\left(\int_{(l_1^{(j)},u_1^{(j)})}^{(u_1^{(j)},+\infty)} + \int_{(l_1^{(j)},l_1^{(j)})}^{+\infty,u_1^{(j)})}\right)
\boldsymbol{\pdf}_{\boldsymbol{\mu}_1, {\Sigma}_1}(\hat{\mathbf{Y}}_1) d\hat{\mathbf{Y}}_1 \times \\
& \quad \qquad \enskip
\left(\int_{(l_2^{(j)},u_2^{(j)})}^{(u_2^{(j)},+\infty)} + \int_{(l_2^{(j)},l_2^{(j)})}^{+\infty,u_2^{(j)})}\right)
\boldsymbol{\pdf}_{\boldsymbol{\mu}_2, {\Sigma}_2}(\hat{\mathbf{Y}}_2) d\hat{\mathbf{Y}}_2 
 \Big)\\
= &\sum_{j=1}^{n+1}\prod_{i=1}^{m=2}
\Big\{ \left( \mbox{MCDF}\big( (u_i^{(j)},\infty), \boldsymbol{\mu}_i, \Sigma_i \big) + \mbox{MCDF}\big( (\infty, u_i^{(j)}), \boldsymbol{\mu}_i, \Sigma_i \big)  \right) \\
& - \left( 
\mbox{MCDF}\big( (\infty, l_i^{(j)}), \boldsymbol{\mu}_i, \Sigma_i \big)
 + \mbox{MCDF}\big( (l_i^{(j)},\infty), \boldsymbol{\mu}_i, \Sigma_i \big) \right) \\
&- \left(\mbox{MCDF}\big( (u_i^{(j)},u_i^{(j)}), \boldsymbol{\mu}_i, \Sigma_i \big) - \mbox{MCDF}\big( (l_i^{(j)},l_i^{(j)}), \boldsymbol{\mu}_i, \Sigma_i \big) \right) \Big\} \numberthis 
\label{def:exact_kpoi_worst}
\end{align*}
Notice that the last line in \eqref{def:exact_kpoi_worst} contains the same component as \eqref{def:exact_kpoi_best} but with a  subtract symbol.

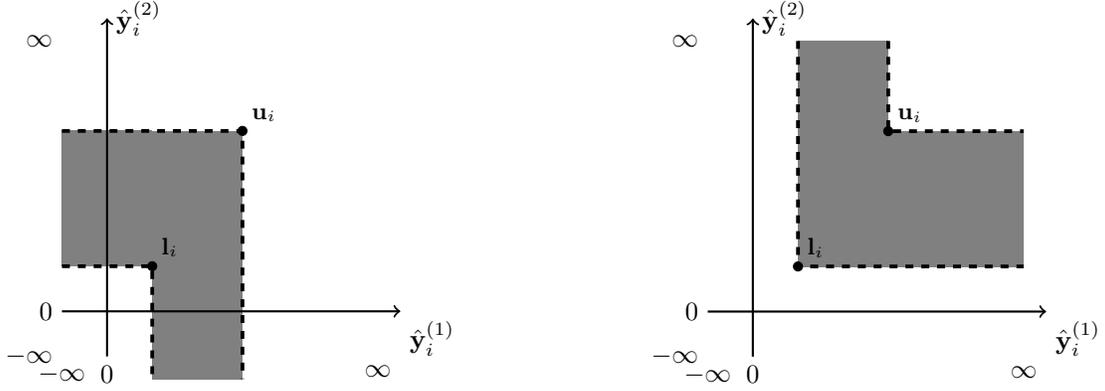
\begin{figure}
\centering
  \begin{subfigure}[b]{0.48\textwidth}
      \centering
      \begin{tikzpicture}[scale=0.6]
        \draw [fill=gray,draw=none] (-1,1) rectangle (1,4);
        \draw [fill=gray,draw=none] (1,-1.5) rectangle (3,4);
        
    	\draw [color=black, fill=black,fill opacity=1] (1, 1) circle (0.1);
    	\node [above right, fill opacity=1] at (1, 1) {\small $\mathbf{l}_i$};
    	
    	\draw [color=black, fill=black,fill opacity=1] (3, 4) circle (0.1);
    	\node [above right, fill opacity=1] at (3,4) {\small $\mathbf{u}_i$};
    	
    	\draw [black, line width=0.5mm, dashed] (-1, 1) -- (1, 1) -- (1, -1.5);
    	\draw [black, line width=0.5mm, dashed] (-1, 4) -- (3, 4) -- (3, -1.5);
    
    	\node [below,fill opacity=1] at (0, -1)   { $0$};
    	\node [below,fill opacity=1] at (6, -1)   { $\infty$};
    	\node [left,fill opacity=1] at (-1, 0)   { $0$};
    	\node [left,fill opacity=1] at (-1, 6)   { $\infty$};
    	\node [left,fill opacity=1] at (-1, -1)   { $-\infty$};
    	\node [below,fill opacity=1] at (-1, -1)   { $-\infty$};
    
    	\draw [thick,->] (-1,0) -- (6.5,0);
    	\node [below right,fill opacity=1] at (6.5,0) {$\hat{\mathbf{y}}^{(1)}_i$};
    	\draw [thick,->] (0,-1) -- (0,6.5);
    	\node [right,fill opacity=1] at (0,6.5) {$\hat{\mathbf{y}}^{(2)}_i$};
    \end{tikzpicture}
    \caption{Range of integral of $\kpoi_{\mbox{best}}$ for $i-$th dimension.}
    \label{subfig:integralDomainQPOIBest}
  \end{subfigure}
  \hfill
  \begin{subfigure}[b]{0.48\textwidth}
      \centering
      \begin{tikzpicture}[scale=0.6]
        \draw [fill=gray,draw=none] (1,1) rectangle (3,6);
        \draw [fill=gray,draw=none] (1,1) rectangle (6,4);
        
    	\draw [color=black, fill=black,fill opacity=1] (1, 1) circle (0.1);
    	\node [above right, fill opacity=1] at (1, 1) {\small $\mathbf{l}_i$};
    	
    	\draw [color=black, fill=black,fill opacity=1] (3, 4) circle (0.1);
    	\node [above right, fill opacity=1] at (3,4) {\small $\mathbf{u}_i$};
    	
    	\draw [black, line width=0.5mm, dashed] (1, 6) -- (1, 1) -- (6, 1);
    	\draw [black, line width=0.5mm, dashed] (3, 6) -- (3, 4) -- (6, 4);
    
    	\node [below,fill opacity=1] at (0, -1)   { $0$};
    	\node [below,fill opacity=1] at (6, -1)   { $\infty$};
    	\node [left,fill opacity=1] at (-1, 0)   { $0$};
    	\node [left,fill opacity=1] at (-1, 6)   { $\infty$};
    	\node [left,fill opacity=1] at (-1, -1)   { $-\infty$};
    	\node [below,fill opacity=1] at (-1, -1)   { $-\infty$};
    
    	\draw [thick,->] (-1,0) -- (6.5,0);
    	\node [below right,fill opacity=1] at (6.5,0) {$\hat{\mathbf{y}}^{(1)}_i$};
    	\draw [thick,->] (0,-1) -- (0,6.5);
    	\node [right,fill opacity=1] at (0,6.5) {$\hat{\mathbf{y}}^{(2)}_i$};
    \end{tikzpicture}
    \caption{Range of integral of $\kpoi_{\mbox{worst}}$ for $i-$th dimension.}
    \label{subfig:integralDomainQPOIworst}
  \end{subfigure}
  \caption{Range of integral for $\kpoi_{\mbox{best}}$ and $\kpoi_{\mbox{worst}}$, where $i$ stands for $i$-th objective, $\mathbf{l}_i=(l_1^{(1)},l_1^{(2)}), \mathbf{u}_i=(u_1^{(1)},u_1^{(2)})$ and grey areas represent the range of integral in $\hat{\mathbf{Y}}_i$.}
  \label{fig:intregralDomain}
\end{figure}

\emph{\textbf{Remark:}}
When $q>2$, the calculations for $\kpoi_{\mbox{best}}$ and for $\kpoi_{\mbox{worst}}$ can be computed in a similar manner. One needs to consider all possible ranges of integration carefully.
The idea of $\kpoi_{\mbox{best}}$ and $\kpoi_{\mbox{worst}}$ is to treat $m\times q$ multivariate normal distributions $\hat{\mathbf{Y}}$ as a $m\times 1$ normal distribution $\hat{\mathbf{Y}}'$ and calculate the PoI of the $\hat{\mathbf{Y}}'$ in the best and worst senses as Definition (\ref{def:kpoi_for_best}) and (\ref{def:kpoi_for_worst}) says, respectively. In $\kpoi_{\mbox{best}}$ and $\kpoi_{\mbox{worst}}$, we first calculate the PoI of a joint multivariate normal distribution for each dimension and then multiply the PoI of each dimension. 
The Range of the integral in $\kpoi_{\mbox{best}}$ and $\kpoi_{\mbox{worst}}$ for each dimension are shown in Figure~\ref{fig:intregralDomain} by grey areas.

\paragraph{\bf $\kpoi_{\mbox{mean}}$}
In this formulation, we consider the case $q\geq2$. 
\begin{align*}
\kpoi_{\mbox{mean}} (M, \boldsymbol{\Sigma}, \pfa) 
& = \frac{1}{q} \sum_{i=1}^{q} \sum_{j=1}^{n+1}
\int_{l_1^{(j)}}^{u_1^{(j)}}\pdf_{\mu_1^{(i)},s_1^{(i)}} d y_1
\int_{l_2^{(j)}}^{u_2^{(j)}}\pdf_{\mu_2^{(i)},s_2^{(i)}} d y_2
\numberthis
\label{eq:exact_kpoi_mean}
\end{align*}

\emph{\textbf{Remark:}} The explicit formulas of the proposed five PoI variants can be easily extended into high-dimensional case. See the details in Appendix.

\subsection{Computational Complexity in Bi-objective Case}
Assuming the computation of a q-dimensional cumulative probability density function takes $O(q)$ time units. The computational complexities of $\kpoi_{\mbox{best}}$ and $\kpoi_{\mbox{worst}}$ are bounded by $O( 2 n \log n) = O(n \log n)$ for every $q$, as both of them only need the evaluation of the MCDF on a $\mathbb{R}^{2\times 2}$ space. In practice, $\kpoi_{\mbox{worst}}$ needs more execution time as it evaluates four times more for the MCDF in every iteration, in comparison with $\kpoi_{\mbox{best}}$. It is easy to conclude the computational complexity of $\kpoi_{\mbox{mean}}$ is $O(q n \log n )$. The computational complexity of $\kpoi_{\mbox{all}}$ is $O(n^q \log n)$ as the number of the combinations of $q$ entries in $\hat{\mathbf{Y}}$ is $n^q$. Following the same idea, $\kpoi_{\mbox{one}}$ requires $\sum_{j=1}^q{q \choose j} n^j=(n+1)^q-1$ times
calculations for q-dimensional cumulative probability density function. Therefore, the computational complexity of $\kpoi_{\mbox{one}}$ is $O\big( (n+1)^q \log n \big) = O (n^q \log n)$.

\emph{\textbf{Remark:}} In practice, a trick for reducing time complexity is to compute the integrals in the stripes that locate between $\mu-3\sigma$ and $\mu+3\sigma$. This trick is not used in the computational speed test part to scientifically analyze the `computational complexity' but is used in experiments to save algorithms' execution time.

\subsubsection*{Computational Speed Test}
Five different exact q-PoIs in Section \ref{subsec:kpoi_definitions} are assessed on {\sc convexSpherical} and {\sc concaveSpherical} Pareto-front approximation sets~\cite{emmerich2011computing}. The results are compared for validation with the MC integration in Algorithm \ref{alg:mc_kpoi}. The MC method is allowed to run for 100,000 iterations. All the experiments are executed on the same computer: Intel(R) i7-4940MX CPU @ 3.10GHz, RAM 32GB. The operating system is Windows 10 (64-bit), and the software platform is MATLAB 9.9.0.1467703 (R2020b).

Table \ref{tab:speedTest} shows the empirical speed experiments for the exact \kpoi method and the MC method. Both the exact calculation method and the MC method are executed without parallel computing\footnote{Any parallel technique can be utilized to speed-up execution times of exact calculation method and the MC method.}. Pareto front sizes are $|\pfa|\in{\lbrace10, 100, 1000\rbrace}$. The parameters in $\GP$ are: $M=[4\enskip9; 8\enskip7]^\top$ and $M=[1\enskip5; 5\enskip1]^\top$ for {\sc convexSpherical} and {\sc concaveSpherical}, respectively; the standard deviation matrix and covariance matrices $\boldsymbol{\Sigma}$ are used for both {\sc convexSpherical} and {\sc concaveSpherical} $\pfa$: \\
$
\Lambda = \begin{bmatrix}
    s_1^{(1)} & s_1^{(2)}  \\
    s_2^{(1)} & s_2^{(2)}
    \end{bmatrix} = 
    \begin{bmatrix}
    2.5 & 2.5 \\
    2.5 & 2.5
    \end{bmatrix},
$
$\Sigma_i = 
    \begin{bmatrix}
    s_i^{(1)^2}                                    & 
    \rho_is_i^{(1)}s_i^{(2)} \\
    \rho_is_i^{(1)}s_i^{(2)}                  & s_i^{(2)^2}
    \end{bmatrix},  i\in\{1,2\}$
, where $\rho_1 = -\rho_2 = 0.5$ and $\rho=\Cov(s^{(1)},s^{(2)})/(s^{(1)}s^{(2)})$.
The average running time of ten repetitions is computed and shown in Table \ref{tab:speedTest}. The result confirms that $\kpoi_{\mbox{mean}}$ processes the lowest running time, which does not require covariance. On the other hand, both  $\kpoi_{\mbox{one}}$ and $\kpoi_{\mbox{all}}$ require a large amount of execution time. The running time of these two q-PoIs is increased by a factor of $|\pfa|^2$. The running times of $\kpoi_{\mbox{worst}}$ and $\kpoi_{\mbox{best}}$ are confirmed with an increase of a factor $|\pfa|$. The comparison of $\kpoi_{best,worst}$ and $\kpoi_{\mbox{mean}}$ indicates that the constant $C$ of $\kpoi_{best,worst}$ is roughly 10 times of $\kpoi_{\mbox{mean}}$ because the CDF of multi-variate normal distribution (MCDF) requires more computational time than that of normal distribution's CDF. Note that this study should not serve as a speed comparison, as the MC method is not precise. 
\begin{table}
\caption{Average running time (s)}
\label{tab:speedTest}
\begin{adjustbox}{width=\columnwidth,center}
\begin{tabular}{c|c|ccccc|c}
\toprule
\multirow{2}{*}{$\pfa$ Type} & \multirow{2}{*}{$|\pfa|$} & \multicolumn{5}{c|}{Exact Calculation}                                             & \multirow{2}{*}{MC} \\
                         &                           & $\kpoi_{\mbox{all}}$ & $\kpoi_{\mbox{one}}$ & $\kpoi_{\mbox{best}}$ & $\kpoi_{\mbox{worst}}$ & $\kpoi_{\mbox{mean}}$ &                              \\
\hline
\multirow{3}{*}{Convex}  & 10                        & 0.1017        & 0.1031        & 0.0195         & 0.0247          & 0.0014         & 62.0334                      \\
                         & 100                       & 7.6448        & 7.6463        & 0.1536         & 0.0787          & 0.0015         & 72.6362                      \\
                         & 1000                      & 655.6621      & 655.6648      & 1.3512         & 0.6463          & 0.0026         & 908.3802                     \\
\hline 
\multirow{3}{*}{Concave} & 10                        & 0.0819        & 0.0825        & 0.0153         & 0.0124          & 0.0003         & 49.3340                      \\
                         & 100                       & 6.7376        & 6.7379        & 0.1238         & 0.0542          & 0.0006         & 57.7963                      \\
                         & 1000                      & 661.5397      & 661.5414      & 1.1952         & 0.5232          & 0.0016         & 821.9510        \\
\bottomrule
\end{tabular}
\end{adjustbox}
\end{table}

\subsection*{Accuracy Comparison}
\begin{figure}[!h]
      \centering
      \begin{subfigure}[b]{0.48\textwidth}
          \centering
          \includegraphics[width=\textwidth]{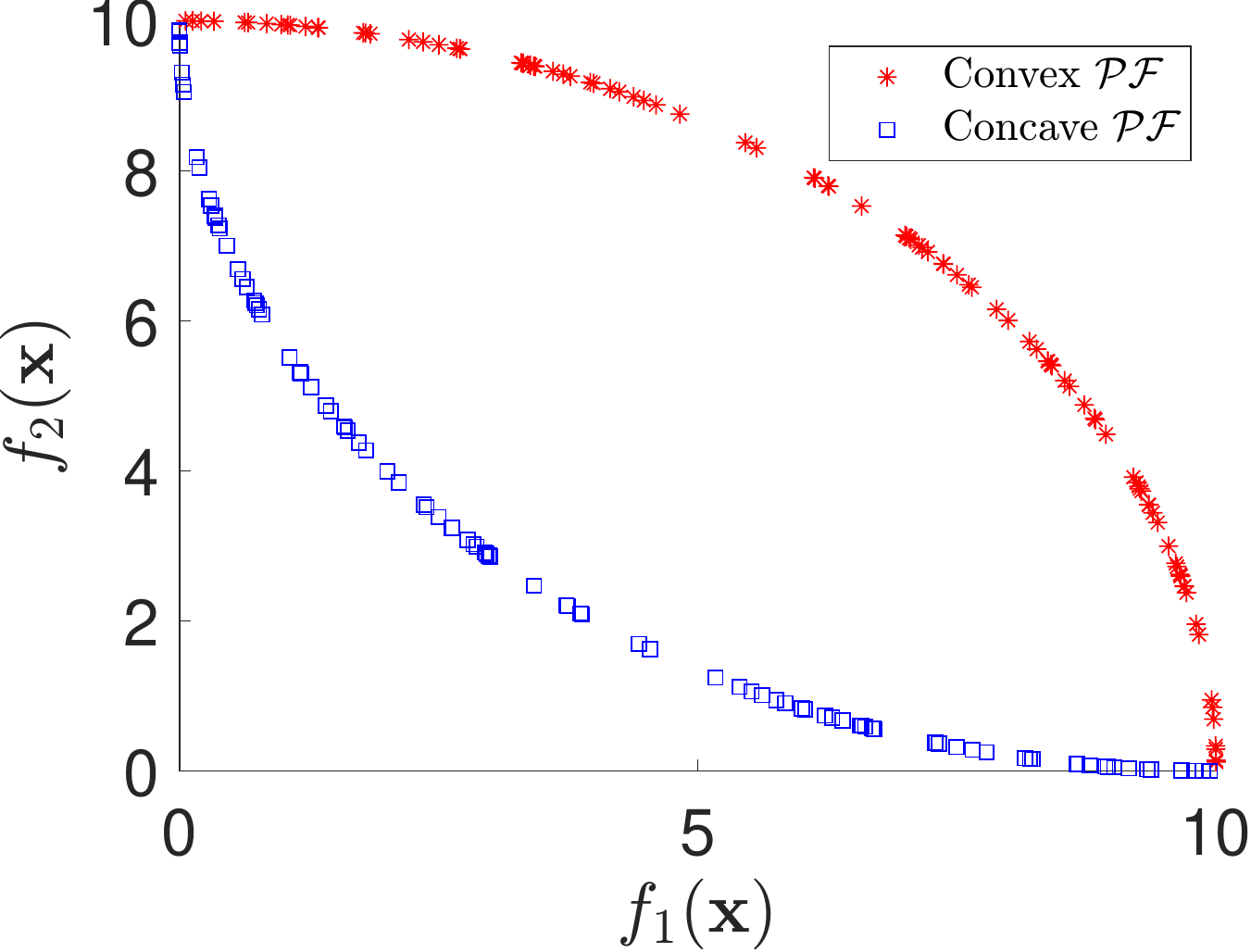}
          \caption{Convex and concave $\pfa.$}
          \label{subfig:convexVSconcave}
      \end{subfigure}
      \hfill
      \begin{subfigure}[b]{0.48\textwidth}
          \centering
          \includegraphics[width=\textwidth]{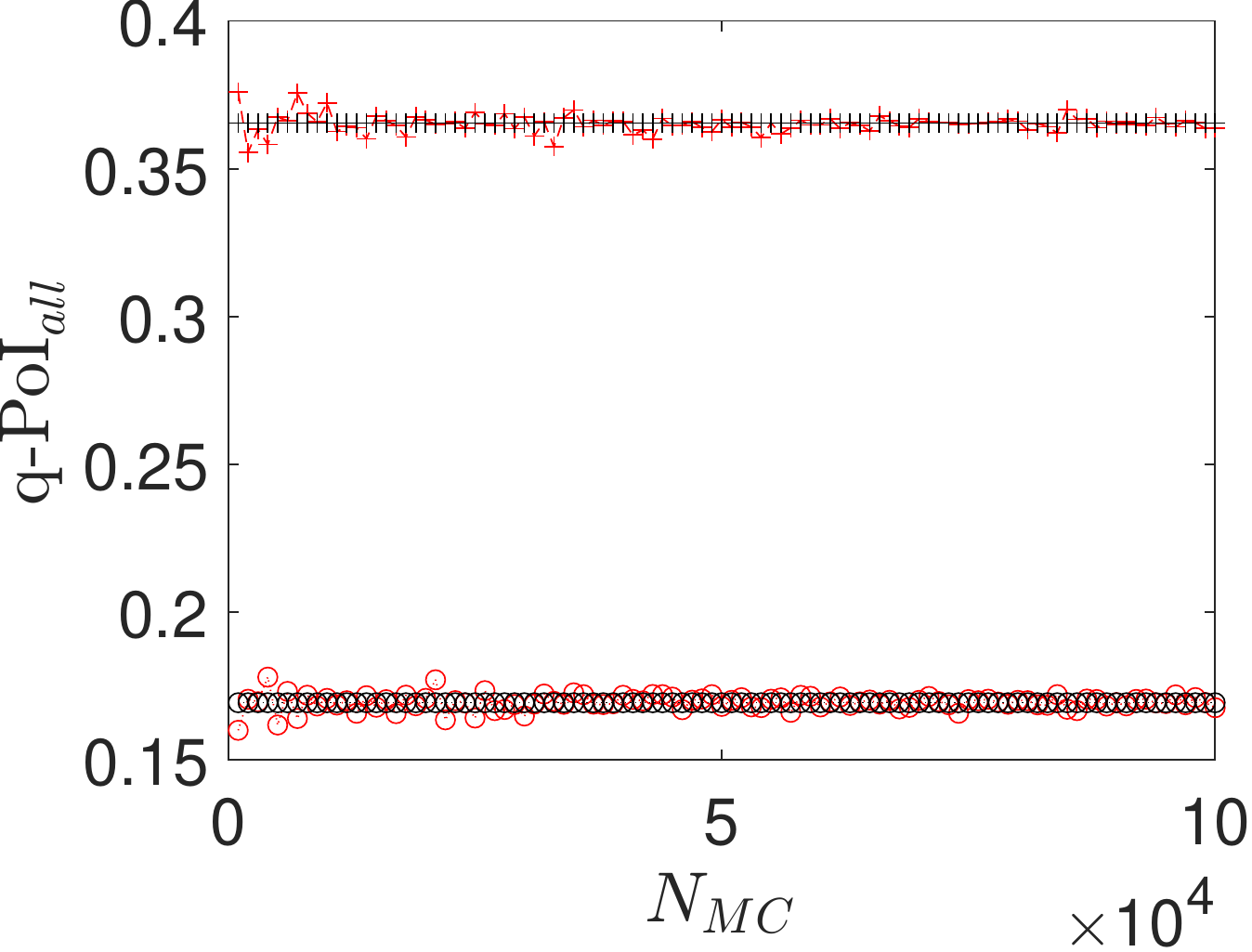}
          \caption{The comparison of MC and ET in computing $\kpoi_{\mbox{all}}$ .}
          \label{subfig:mc_vs_et_qpoi_all}
      \end{subfigure}
      \hfill
      \begin{subfigure}[b]{0.48\textwidth}
          \centering
          \includegraphics[width=\textwidth]{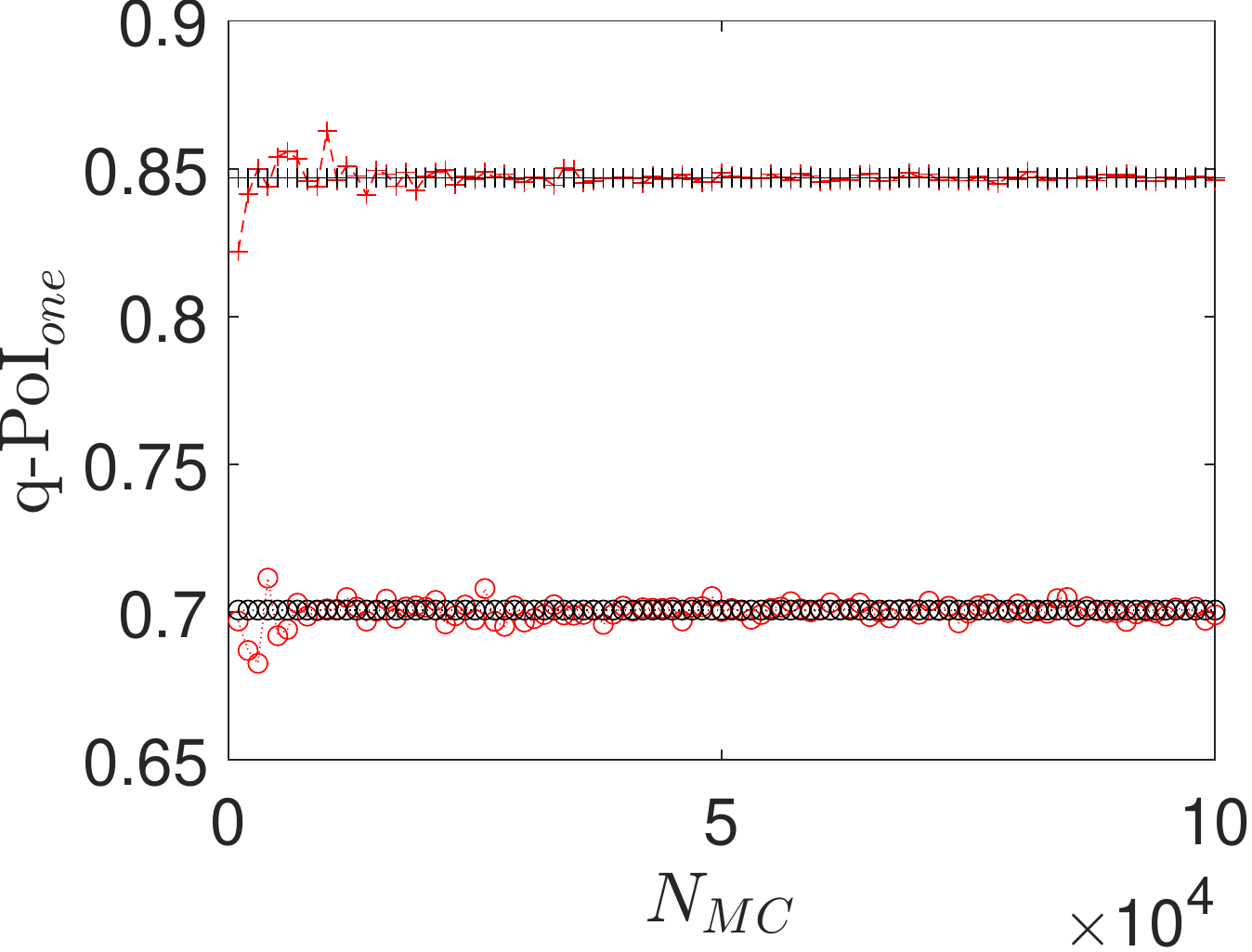}
          \caption{The comparison of MC and ET in computing $\kpoi_{\mbox{one}}$ .}
          \label{subfig:mc_vs_et_qpoi_one}
      \end{subfigure}
      \hfill
      \begin{subfigure}[b]{0.48\textwidth}
          \centering
          \includegraphics[width=\textwidth]{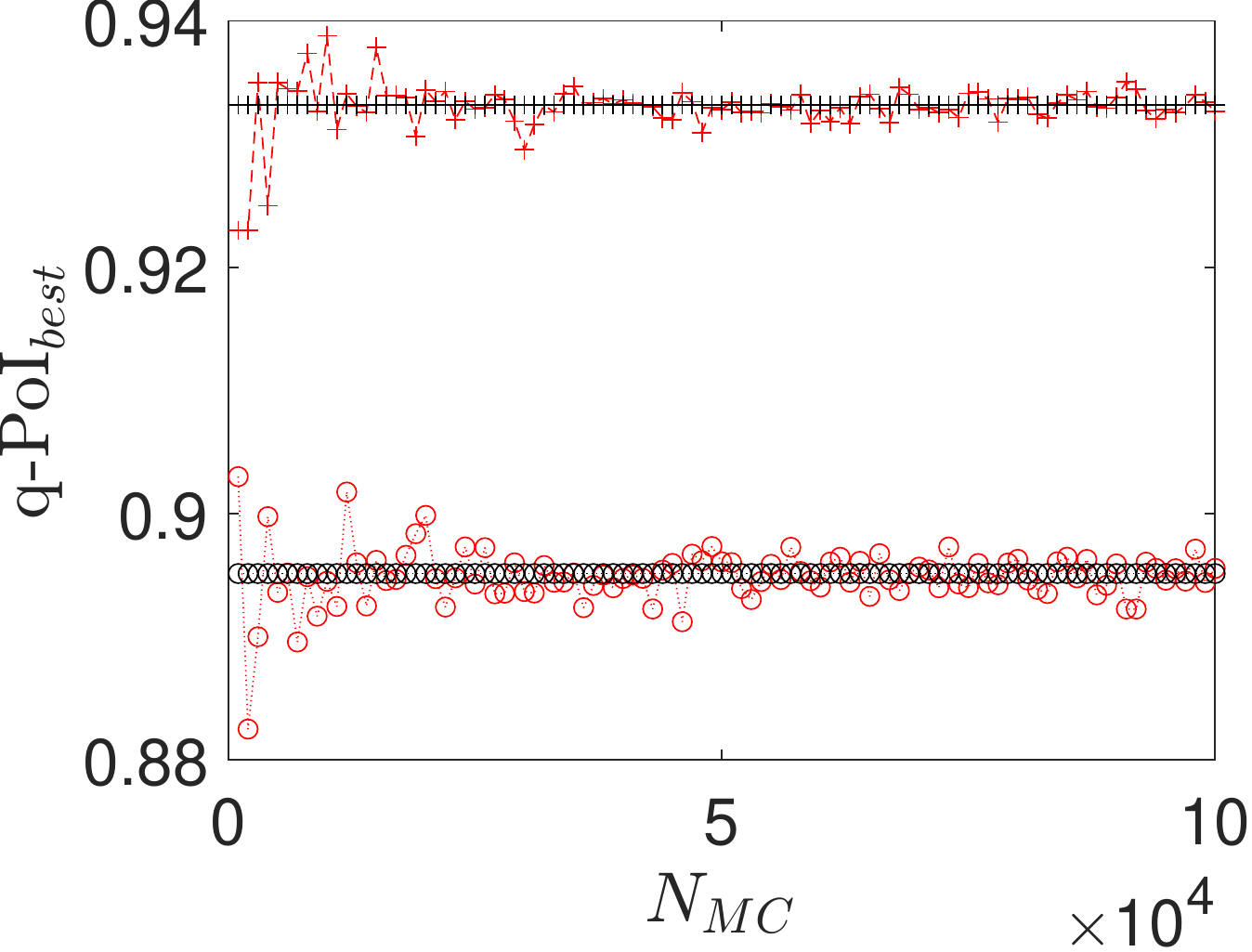}
          \caption{The comparison of MC and ET in computing $\kpoi_{\mbox{worst}}$ .}
          \label{subfig:mc_vs_et_qpoi_worst}
      \end{subfigure}
      \hfill 
      \begin{subfigure}[b]{0.48\textwidth}
          \centering
          \includegraphics[width=\textwidth]{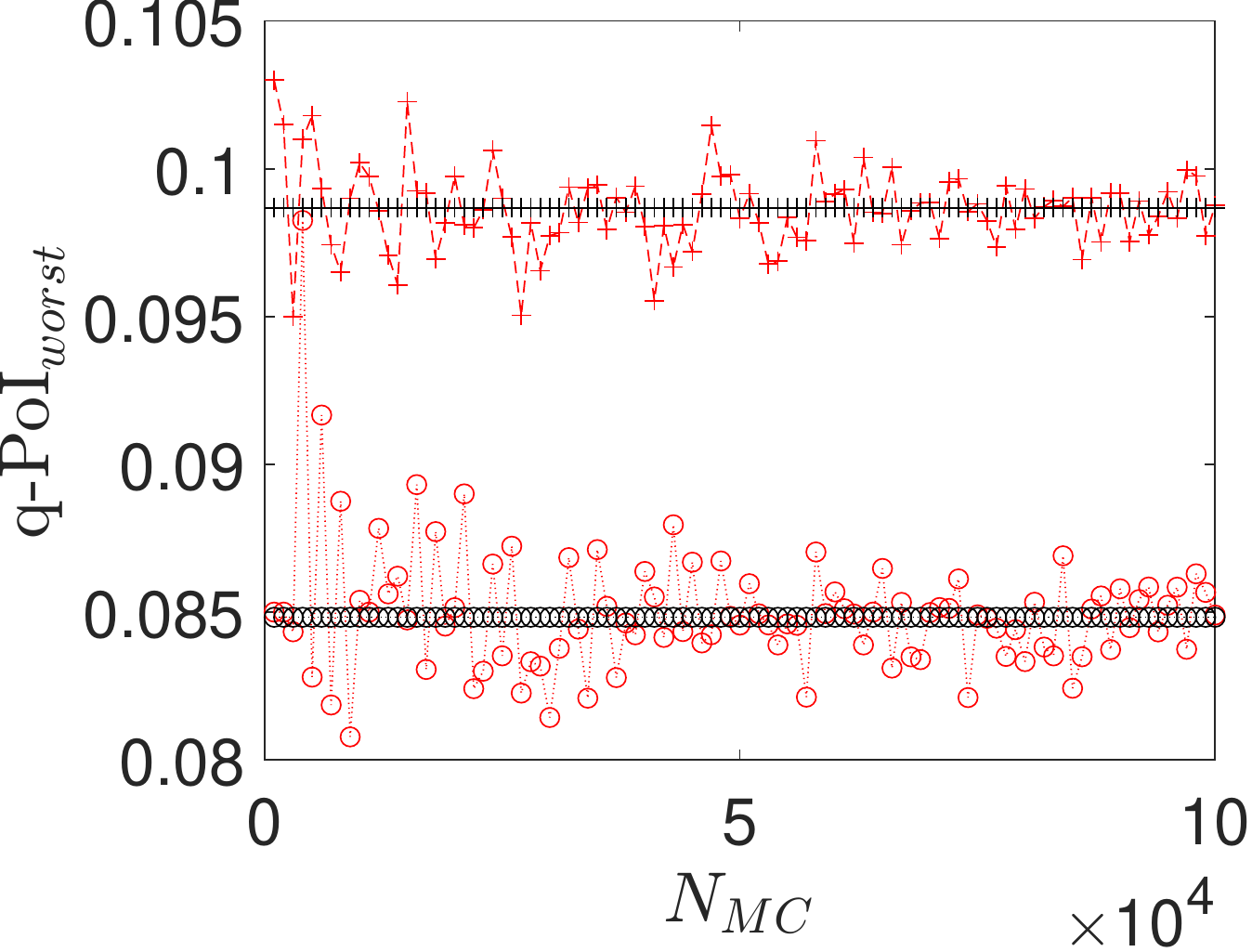}
          \caption{The comparison of MC and ET in computing $\kpoi_{\mbox{best}}$ .}
          \label{subfig:mc_vs_et_qpoi_best}
      \end{subfigure}
      \hfill
      \begin{subfigure}[b]{0.48\textwidth}
          \centering
          \includegraphics[width=\textwidth]{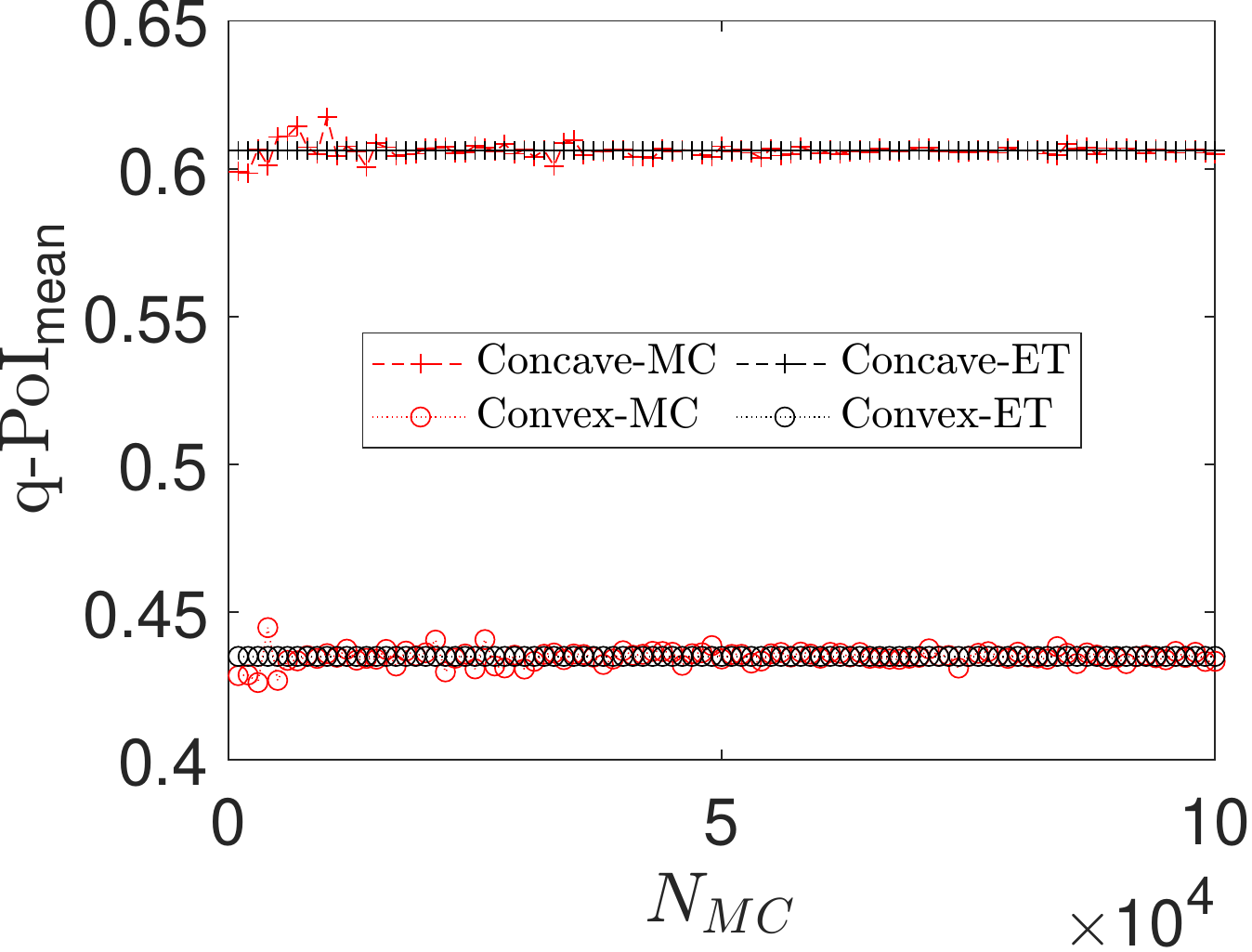}
          \caption{The comparison of MC and ET in computing $\kpoi_{\mbox{mean}}$ .}
          \label{subfig:mc_vs_et_qpoi_avg}
      \end{subfigure}
      \caption{The studies of accuracy comparison between the Monte Carlo (MC) method and exact computation (ET), where $N_{MC}$ represents the number of iterations in the MC method. \emph{Note} that the y-axis limit differs in (b), (c), (d), (e) and (f).  }
     \label{fig:accuracy_mc_vs_et}
\end{figure}
Figure~\ref{fig:accuracy_mc_vs_et} shows the randomly generated {\sc convexSpherical} and {\sc concaveSpherical} Pareto fronts of $|\pfa|=100$ for the 2-D case from \cite{emmerich2011computing} in the Fig. \ref{subfig:convexVSconcave}, and the convergence figures of the MC integration of the five different q-PoIs in the remaining subfigures. The parameters of the evaluated batch $\hat{\mathbf{Y}}$ are mean matrices $M=[4\enskip9;8\enskip7]^\top$ and $M=[1\enskip5;5\enskip1]^\top$ for {\sc convexSpherical} and {\sc concaveSpherical} Pareto-front approximation sets, respectively. For both types of the Pareto-front approximation sets, the standard deviation and coefficient are the same, namely, $\Lambda=[2.5\enskip2.5;2.5\enskip2.5]^\top$ and $\rho_1=-\rho_2=0.5$. The results show that the $\kpoi_i|_{i\in\{\mbox{all, one, mean}\}}$ values based on the MC method are similar to the exact method after 50,000 iterations. However, the MC method requires more iterations to get a sufficiently accurate value. 

\subsection{Influence of Covariance Matrices on q-PoIs}
\begin{figure}
      \centering
      \begin{subfigure}[b]{0.3\textwidth}
          \centering
          \includegraphics[width=\textwidth]{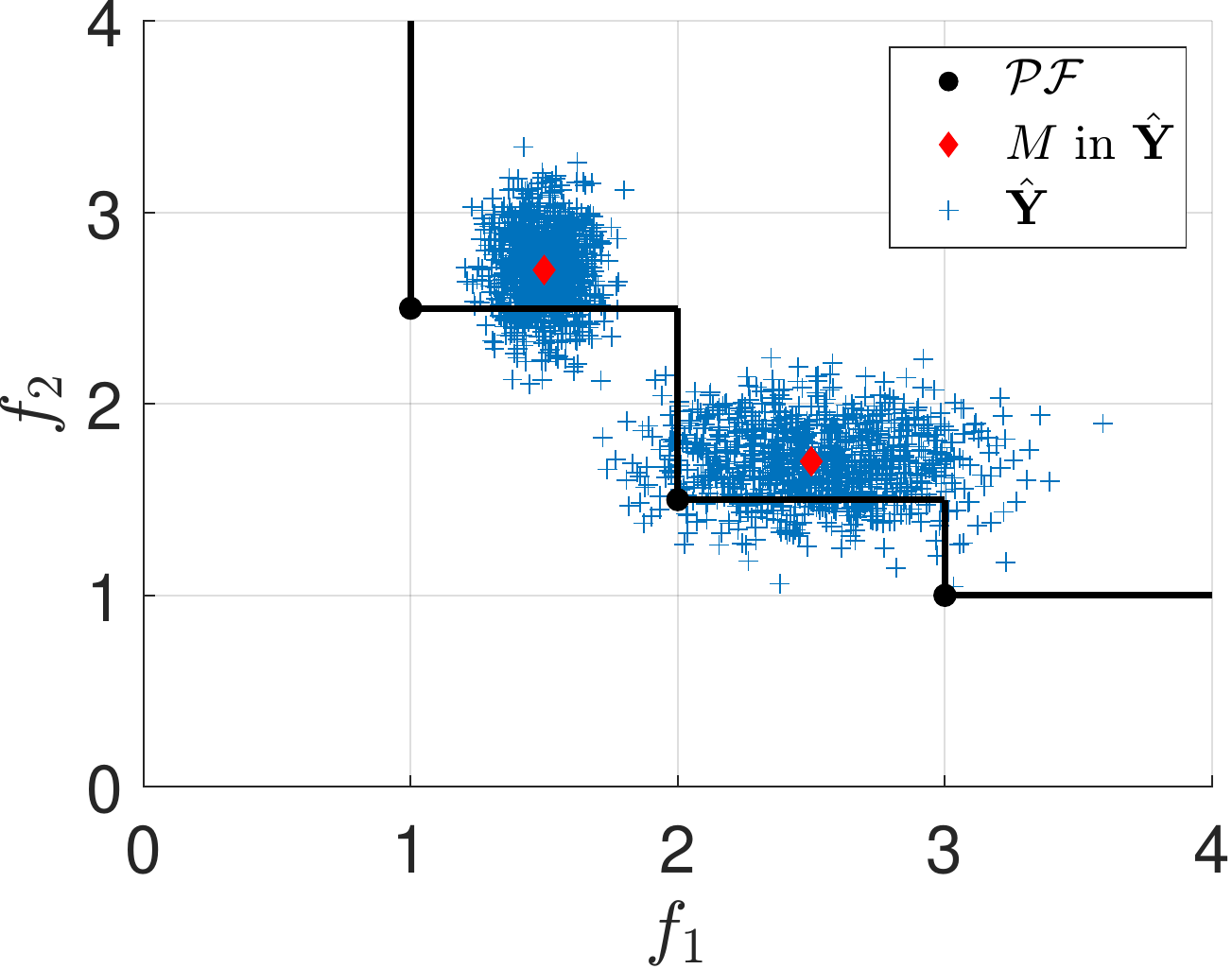}
          \caption{$\pfa \text{ and } \hat{\mathbf{Y}}\sim\mathbfcal{N}(M, \alpha^2\mathbf{\Sigma})$ in Case I, where $M=[1.5 \enskip 2.7; 2.5 \enskip 1.7]^\top$ and $\alpha=0.1$.}
          \label{fig:pf_1}
      \end{subfigure}
      \hfill
      \begin{subfigure}[b]{0.3\textwidth}
          \centering
          \includegraphics[width=\textwidth]{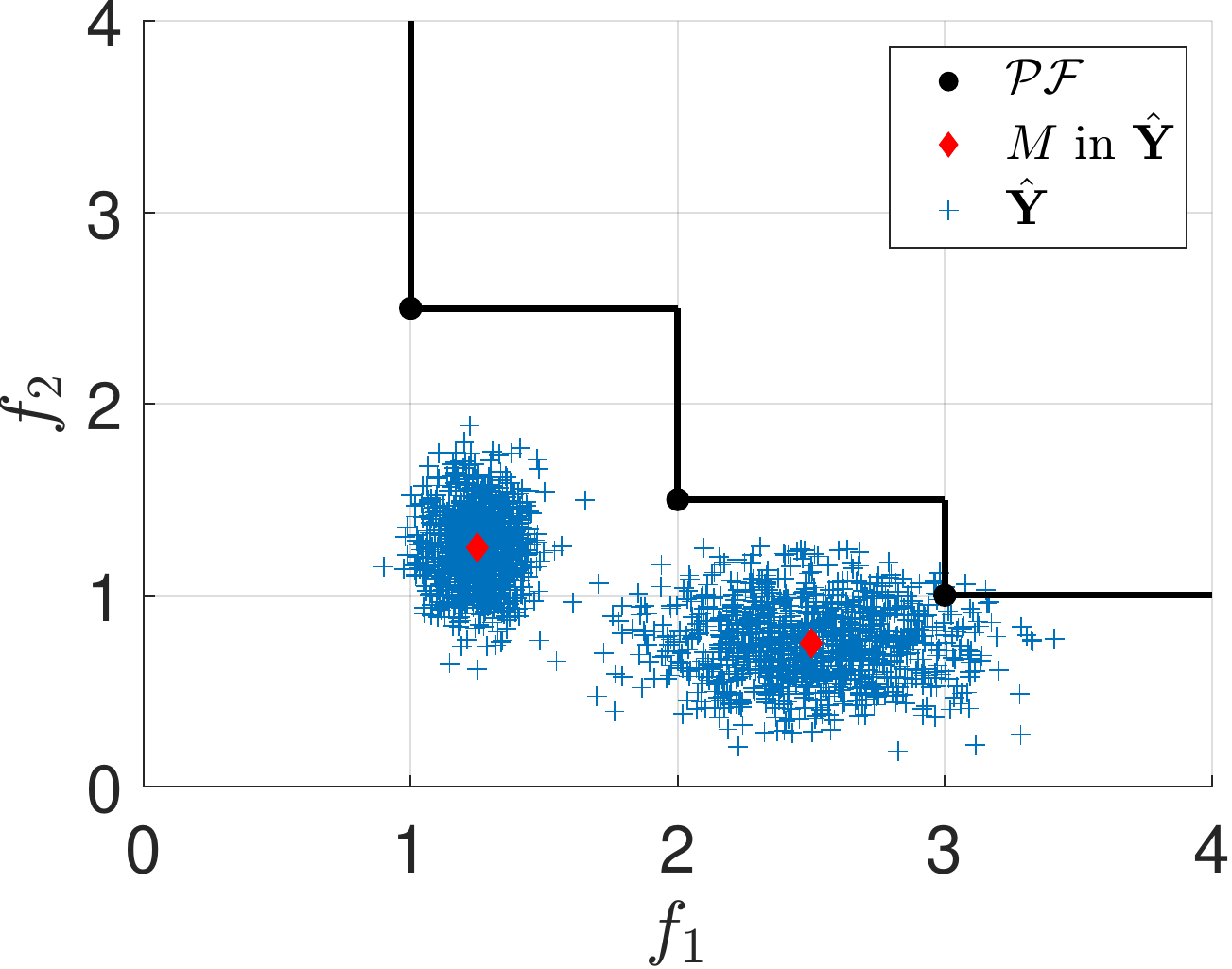}
          \caption{$\pfa \text{ and } \hat{\mathbf{Y}}\sim\mathbfcal{N}(M, \alpha^2\mathbf{\Sigma})$ in Case II, where $M=[1.25 \enskip 1.25; 2.5  \enskip 0.75]^\top$ and $\alpha=0.1$.}
          \label{fig:pf_2}
      \end{subfigure}
      \hfill
      \begin{subfigure}[b]{0.3\textwidth}
          \centering
          \includegraphics[width=\textwidth]{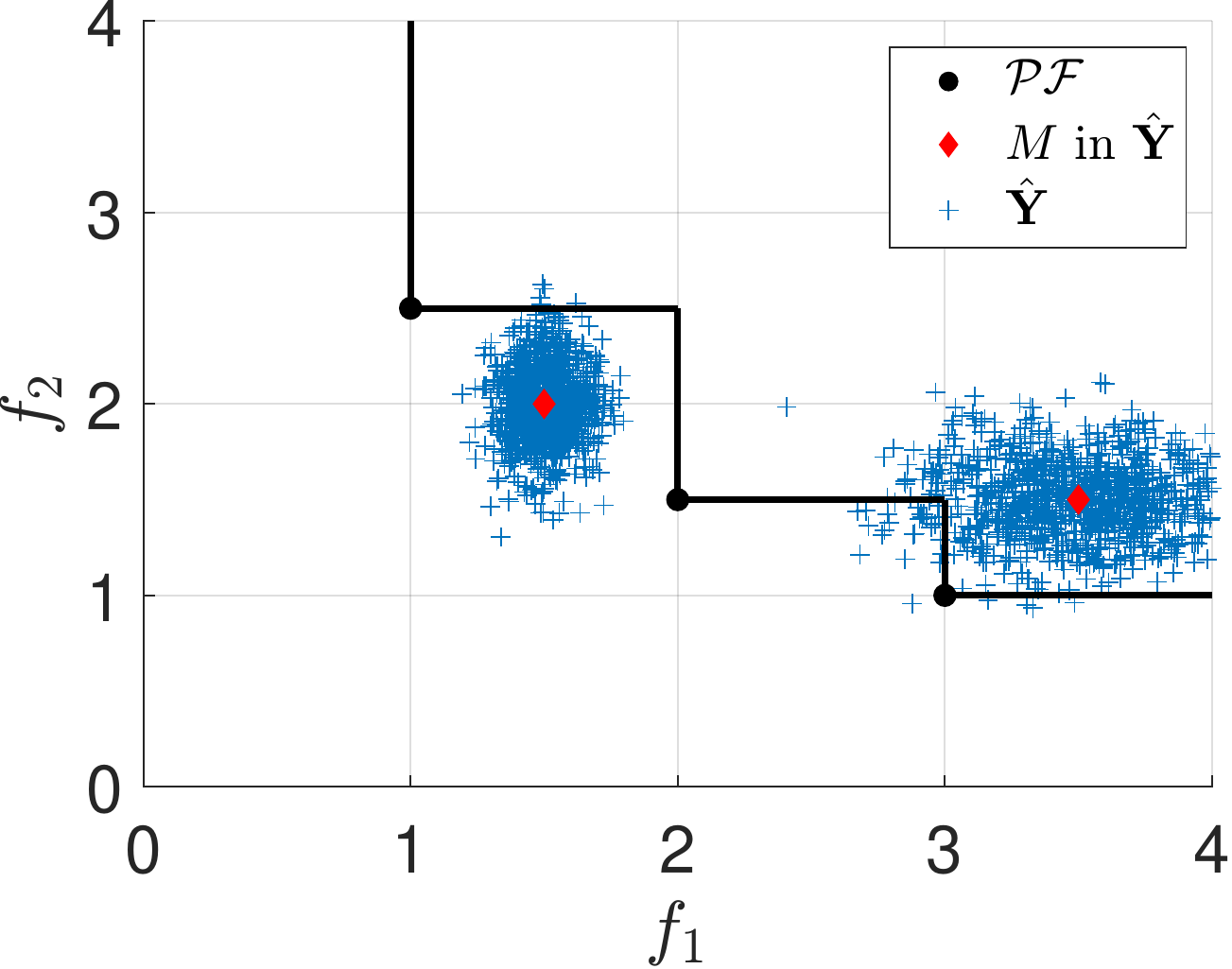}
           \caption{$\pfa \text{ and } \hat{\mathbf{Y}}\sim\mathbfcal{N}(M, \alpha^2\mathbf{\Sigma})$ in Case III, where $M=[1.5 \enskip 2; 3.5 \enskip 1.5]^\top$ and $\alpha=0.1$.}
          \label{fig:pf_3}
      \end{subfigure}
         \caption{The distribution of $\hat{\mathbf{Y}\sim\mathbfcal{N}(M, 0.01\mathbf{\Sigma})}$ with 1,000 sampled points by using the same covariance matrix and three different $M$.}
         \label{fig:threePfCases}
\end{figure}

 As reported in~\cite{Emmerich2020,greedyisgood}, the position of $M$ leads to variance monotonic properties of PoI. Here we investigate similar behaviors for q-PoIs. When two points are considered in an objective space, there are three different cases: Case I -- both two points in the mean matrix $M$ are located in the dominated space; Case II -- both two points in $M$ are located in the non-dominated space; Case III -- one point in $M$ is located in the non-dominated space, and the other one is located in the dominated space. Therefore, the behaviours of q-PoIs under varying covariance matrices $\boldsymbol{\Sigma}$ in a batch $\hat{\mathbf{Y}}$ are analyzed in three different mean matrices $M$, for $M=[1.5 \enskip 2.7; 2.5 \enskip 1.7]^\top$, $M=[1.25 \enskip 1.25; 2.5  \enskip 0.75]^\top$, $M = [1.5 \enskip 2; 3.5 \enskip 1.5]^\top$ corresponding to Case I in Fig. \ref{fig:pf_1}, Case II in Fig. \ref{fig:pf_2}, and Case III in Fig. \ref{fig:pf_3}, respectively. A Pareto approximation set is designed to be $\pfa=[1 \enskip2.5; 2 \enskip1.5; 3 \enskip1]^\top$. The standard deviation matrix is $\Lambda = [s_1^{(1)} \enskip s_1^{(2)}; s_2^{(1)} \enskip s_2^{(2)}] = [1\enskip 3; 2 \enskip 2]$, and the covariance matrices $\boldsymbol{\Sigma}$ can be achieved by $\Lambda$ and $\rho_1=0.5, \rho_2=-0.5$. 

    
\begin{figure}
      \centering
      \begin{subfigure}[b]{0.48\textwidth}
          \centering
          \includegraphics[width=\textwidth,height=4.2cm]{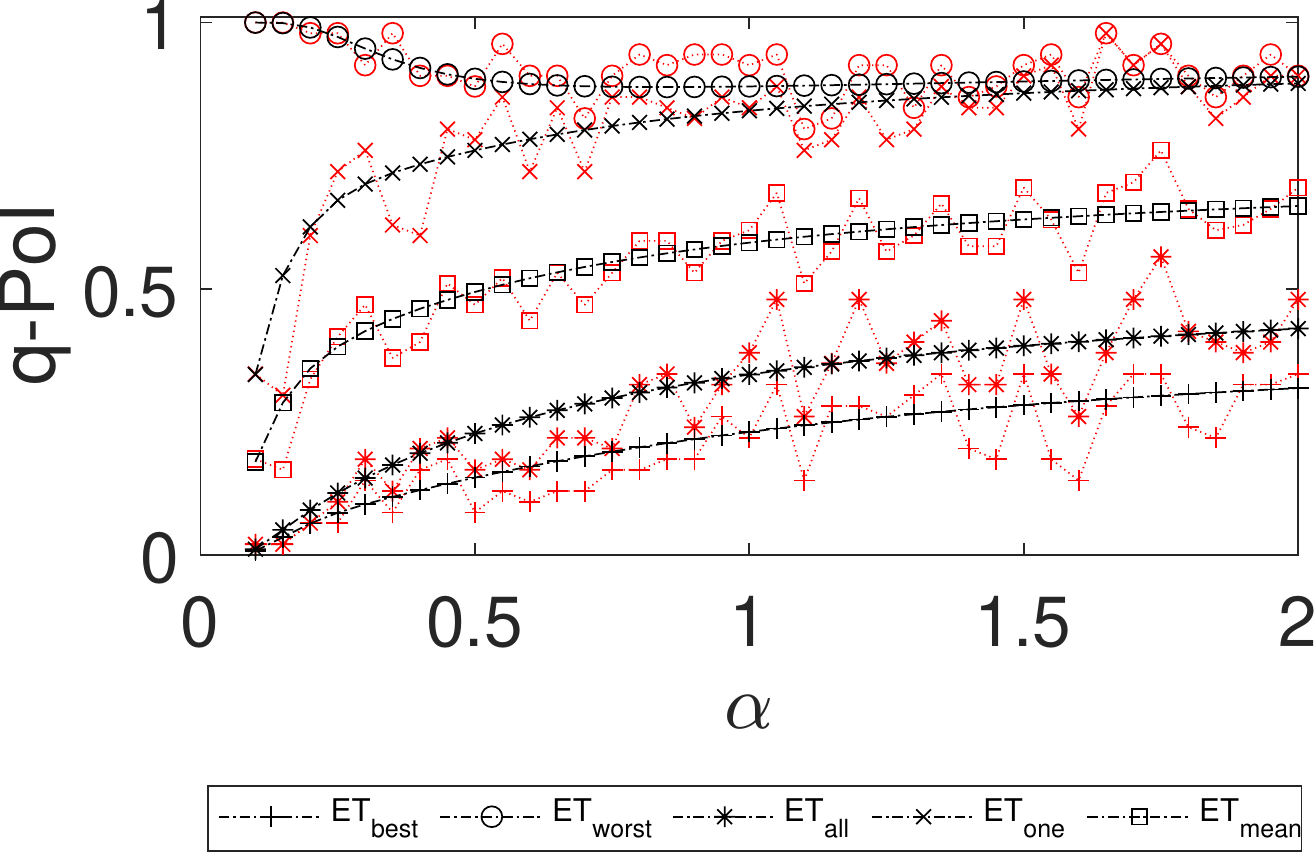}
          \caption{The behaviors of q-PoIs w.r.t $\alpha$ in Case I}
          \label{subfig:sInCaseI}
      \end{subfigure}
      \hfill
      \begin{subfigure}[b]{0.48\textwidth}
          \centering
          \includegraphics[width=\textwidth,height=4.2cm]{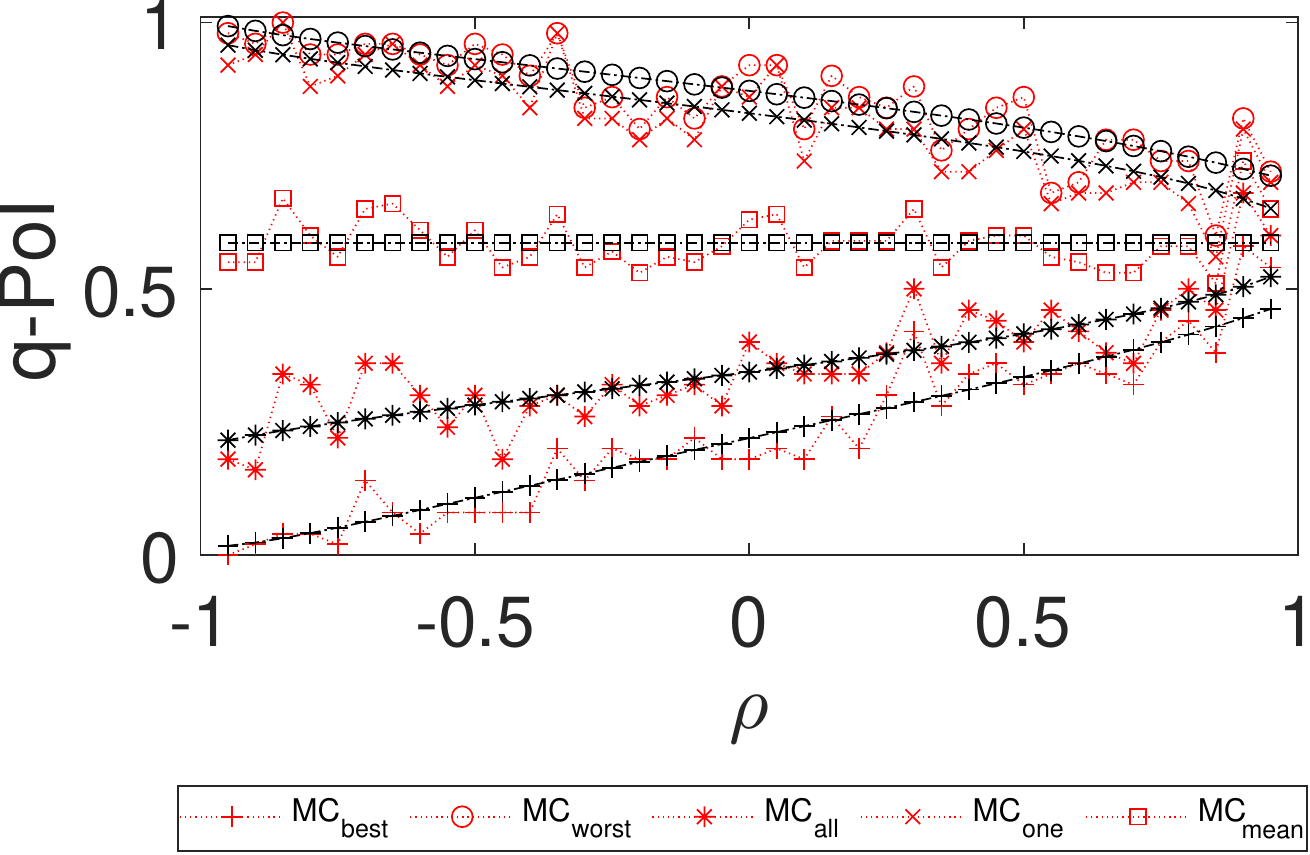}
          \caption{The behaviors of q-PoIs w.r.t $\rho$ in Case I}
          \label{subfig:rhoInCaseI}
      \end{subfigure}
      \hfill
      \begin{subfigure}[b]{0.48\textwidth}
          \centering
          \includegraphics[width=\textwidth,height=4.2cm]{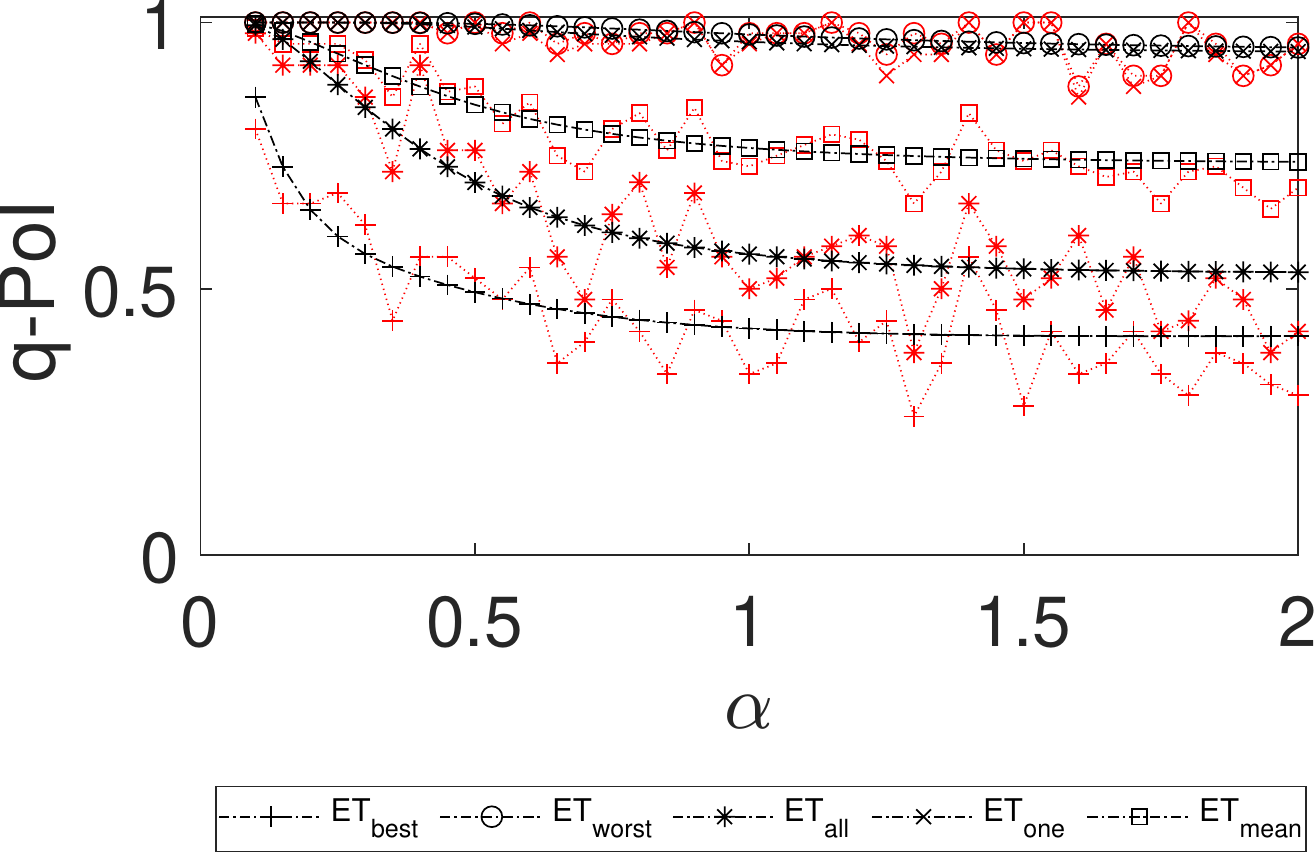}
          \caption{The behaviors of q-PoIs w.r.t $\alpha$ in Case II}
          \label{subfig:sInCaseII}
      \end{subfigure}
      \hfill
      \begin{subfigure}[b]{0.48\textwidth}
          \centering
          \includegraphics[width=\textwidth,height=4.2cm]{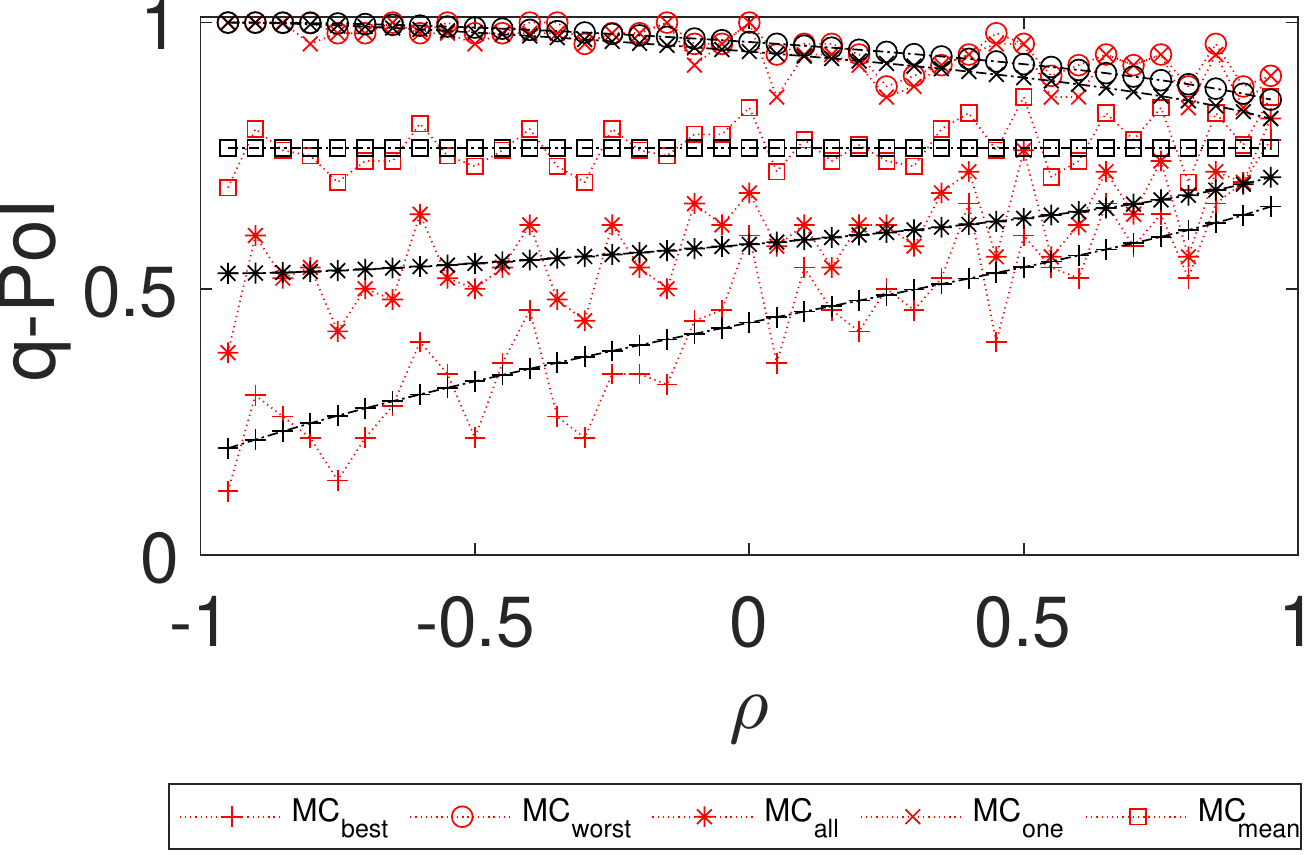}
          \caption{The behaviors of q-PoIs w.r.t $\rho$ in Case II}
          \label{subfig:rhoInCaseII}
      \end{subfigure}
      \hfill 
      \begin{subfigure}[b]{0.48\textwidth}
          \centering
          \includegraphics[width=\textwidth,height=4.2cm]{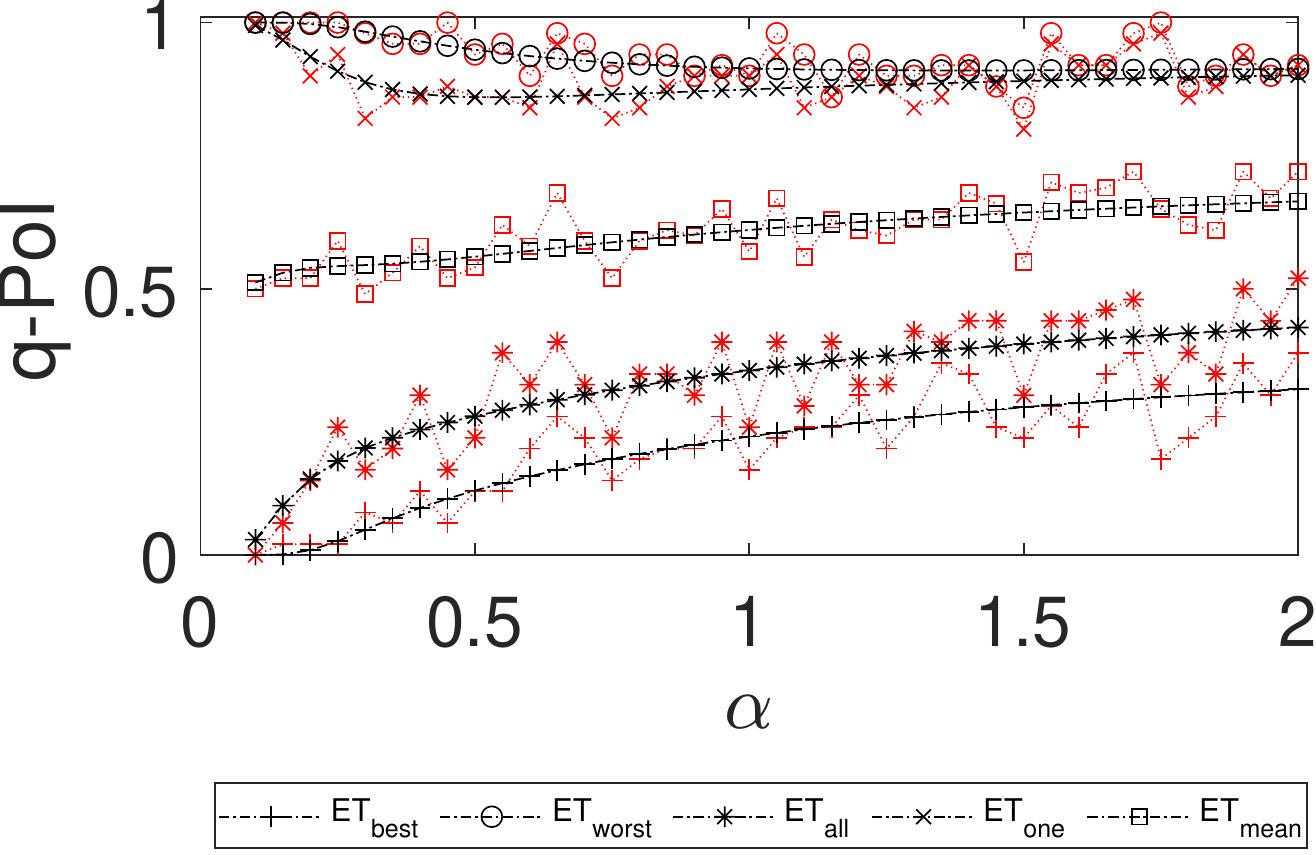}
          \caption{The behaviors of q-PoIs w.r.t $\alpha$ in Case III}
          \label{subfig:sInCaseIII}
      \end{subfigure}
      \hfill
      \begin{subfigure}[b]{0.48\textwidth}
          \centering
          \includegraphics[width=\textwidth,height=4.2cm]{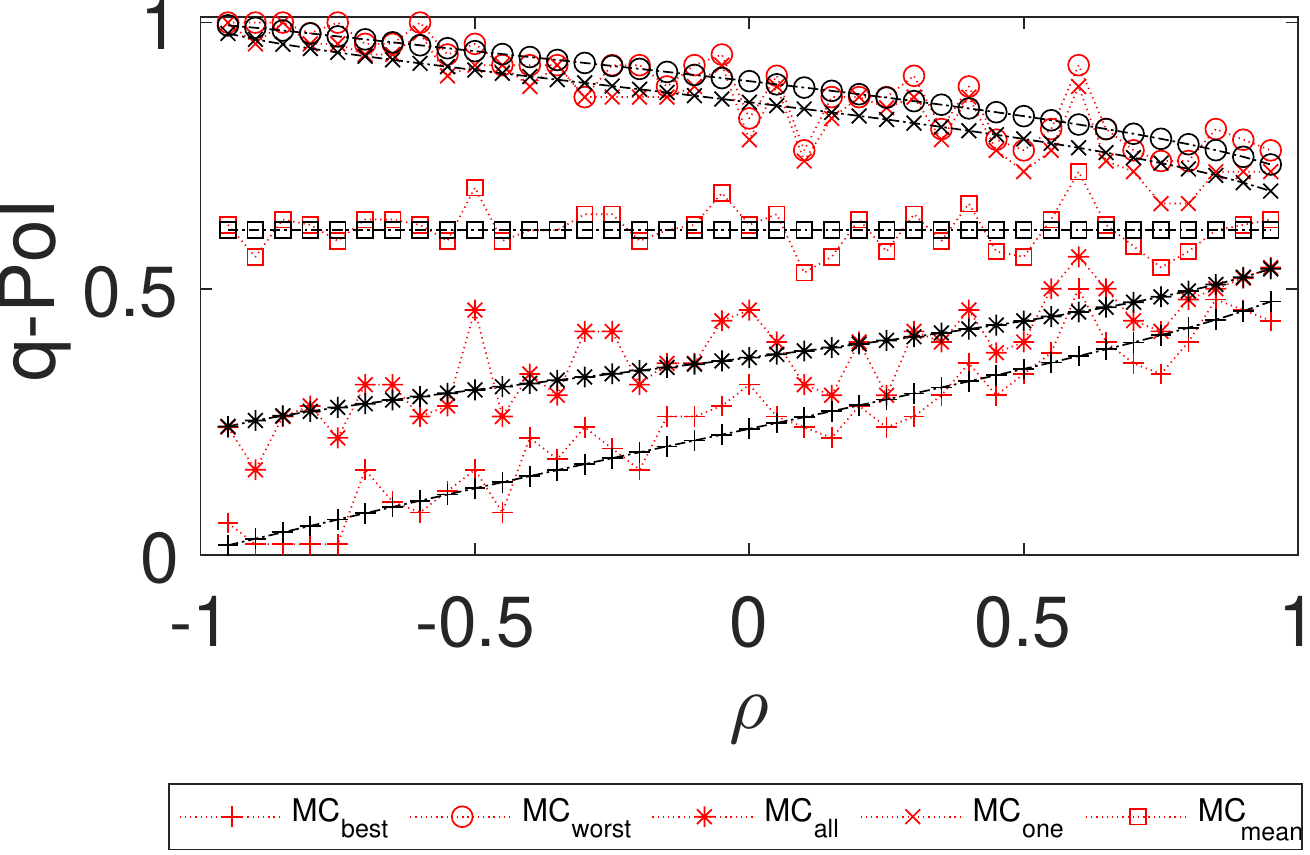}
          \caption{The behaviors of q-PoIs w.r.t $\rho$ in Case III}
          \label{subfig:rhoInCaseIII}
      \end{subfigure}
      \caption{The behavior studies of five different q-PoIs w.r.t. standard deviation matrix $\Sigma$ and correlation coefficient $\rho$ in three different cases.}
     \label{fig:monotonicity}
\end{figure}

Fig. \ref{fig:monotonicity} shows the position-dependent behavior of q-PoIs w.t.r standard deviation $s$ and correlation coefficient $\rho$ in three different cases. In the left column, correlation coefficients are constant, but the standard deviation matrix $\Lambda$ is varied by multiplying a factor $\alpha = [0.1, 2.0]$ with stepsize $0.05$. In the right column, where the $\Lambda$ is constant, the correlation coefficients $\rho_1=\rho_2$ change from $-0.95$ to $0.95$ with a step size of 0.05. The five q-PoIs are computed by both exact formulas (represented by black curves) and by the MC method (represented by red curves), where the number of iterations in the MC method is $1000$. The first, the second and the third rows are the corresponding results of Case I (Figure~\ref{subfig:sInCaseI} and ~\ref{subfig:rhoInCaseI}), Case II (Figure~\ref{subfig:sInCaseII} and ~\ref{subfig:rhoInCaseII}) and Case III (Figure~\ref{subfig:sInCaseIII} and ~\ref{subfig:rhoInCaseIII}), respectively. 

In the left column of Figure~\ref{fig:monotonicity}, when both points in $M$ are dominated by the $\pfa$ (Case I), all q-PoIs increase w.r.t. a standard deviation matrix $\Lambda$, except for $\kpoi_{\mbox{worst}}$, because a large variance of a point in the dominated space would increase the probability of a sampled point falling into a non-dominated space. For the reason of $\kpoi_{\mbox{worst}}$ decreases in Case I is that $\mbox{ConJ}(M)$ dominates $\pfa$ when $\Lambda$ is a zero matrix. Therefore, an increasing $\Lambda$ decreases $\kpoi_{\mbox{worst}}$. In Case II, when both points in $M$ are not dominated by $\pfa$, all q-PoIs  decrease w.r.t. a standard deviation matrix $\Lambda$, because points in $M$ are located in the non-dominated space of $\pfa$, and a large variance will lead to widespread distributions of $\hat{\mathbf{Y}}$. In Case III, each \kpoi either decreases or increases w.r.t $\Lambda$, except for $\kpoi_{\mbox{one}}$, of which the curve is convex. This is reasonable because a point in $M$ dominates the $\pfa$, and another point in $M$ is dominated by the $\pfa$. The distributions of the left red and right red points in Fig. \ref{fig:pf_3} cover more dominated and non-dominated space, respectively, when $\Lambda$ increases. Since $\kpoi_{\mbox{one}}$ is defined as the \poi of \emph{at least one} point dominates a $\pfa$, there is a trade-off balancing between the effects of two multivariate normal distributions. Therefore, a stationary point regarding $\Lambda$ exists in $\kpoi_{\mbox{one}}$ in Case III. Comparing $\kpoi_{\mbox{worst}}$ and $\kpoi_{\mbox{one}}$, the difference between these two q-PoIs can be clearly distinguished when at least one point in $\hat{\mathbf{Y}}$ is located close to $\pfa$ (i.e., Case I and Case III). 

In the right column of Figure~\ref{fig:monotonicity}, we find that $\kpoi_{\mbox{worst}}$ and $\kpoi_{\mbox{one}}$ decrease w.r.t. $\rho_1, \rho_2$, while $\kpoi_{\mbox{best}}$ and $\kpoi_{\mbox{all}}$ increases in all the three cases. Unsurprisingly, a correlation coefficient does not influence $\kpoi_{\mbox{mean}}$. Note that the values of all the q-PoIs at $\rho=0$ represent their corresponding \kpoi values without correlations. The influence of $\rho$ on $\kpoi_{\mbox{all}}$ and on $\kpoi_{\mbox{best}}$ is positively correlated because a positive covariance leads the coordinate values of two solutions closer, and vice vera. One can also conclude that $\rho$ has a negative influence on $\kpoi_{\mbox{one}}$ since Eq. \eqref{eq:exact_kpoi_one} shows that $\rho$ is negatively correlated. 

Both $\kpoi_{\mbox{all}}$ and $\kpoi_{\mbox{best}}$ have a stricter requirement in their definitions than the other three q-PoIs. Compared with $\kpoi_{\mbox{all}}$, $\kpoi_{\mbox{best}}$ has a more `relaxed-exploration' property and a greedy characteristic because of its strict requirement in the definition. This is why the values of $\kpoi_{\mbox{best}}$ are always smaller than that of $\kpoi_{\mbox{all}}$. Comparing $\kpoi_{\mbox{worst}}$ and $\kpoi_{\mbox{one}}$, $\kpoi_{\mbox{worst}}$ enhances exploration caused by a more relaxed requirement of $\mathrm{I(.)}$ by its definition\footnote{PoI's explorative property can be also enhanced by introducing an $\epsilon$-improvement with a carefully designed~\cite{Emmerich2020}. The proposed k-PoI can also incorporate the $\epsilon$ strategy for the same purpose.}. This explains why the values of $\kpoi_{\mbox{worst}}$ are larger than those of $\kpoi_{\mbox{one}}$.

%

\section{Empirical Experiments}
\subsection{Benchmarks}
In this paper, experiments on 20 artificial bi-objective test problems are performed, including ZDT1-3 \cite{zitzler2000comparison}, GSP-1 with $\gamma=0.4$, GSP-2~\cite{Emmerich2007gsp} with $\gamma=1.8$, MaF1, MaF5, and MaF12~\cite{Cheng2017}, mDTLZ1-4~\cite{mdtlz}, WOSGZ1-8~\cite{wosgz}. The dimensions of decision space are 3, 5, 10, and 15 for GSP1-2, ZDT1-3, mDTLZ1-4, and WOSGZ1-8, respectively. The reference points are $[11,11]$ for ZDT1-3, $[1.2, 1.2]$ for mDTLZ1-4 and WOSGZ1-8, and $[5,5]$ for the other problems. 

\subsection{Algorithm Configuration}
The five proposed q-PoIs are compared with other indicator-based MOBGO algorithms (original PoI, two parallel techniques of PoI by using Kriging Believer and Constant Liar with a `mean' liar strategy in Alg.~\ref{alg:mobgo}, and q-EHVI~\cite{daulton2020differentiable}), 
two state-of-the-art multiple-point MOBGO (ParEGO~\cite{knowles2006parego}, MOEA/D-EGO~\cite{zhang2009expensive}), and three recently proposed multiple-point surrogated-assisted multi-objective optimization algorithms (TSEMO~\cite{bradford2018efficient}, DGEMO~\cite{dgemo}, and MOEA/D-ASS~\cite{moea/d-ass}). The platform of TSEMO, ParEGO, MOED/D-EGO, and DGEMO is Python, and the platform of the other algorithms is MATLAB in this paper~\footnote{The Python source code is available on \url{https://github.com/yunshengtian/DGEMO}, and the source code of MOEA/D-ASS is available on \url{https://github.com/ZhenkunWang/MOEAD-ASS}}.

In all the experiments, the number of DoE ($\eta$) is $\min\{6\times d, 60\}$ and the maximum function evaluation is $T_c = \min\{ \eta \times 9, \eta + (200 - \eta) \times 2  \} $.
In indicator-based MOBGO algorithms, the acquisition function is optimized by CMA-ES (of 1 restart and at most 2000 iterations) to search for optimal $\mathrm{X}^*$ due to its favorable performance on BBOB function testbed~\cite{hansen2009benchmarking}. The optimizers of the other algorithms in this paper are NSGA-II, CMA-ES, MOEA/D, GA, and a so-called `discovery optimization'~\cite{discovery}, respectively, in TSEMO~\cite{tsemo}, a batch-version ParEGO~\cite{dgemo}, MOEA/D-EGO\cite{zhang2009expensive}, MOEA/D-ASS~\cite{moea/d-ass}, and DGEMO~\cite{dgemo}. The maximum iteration of NSGA-II and MOEA/D is $2\times2000$. Among all the optimizers, the 'discovery optimization'~\cite{discovery} is the only optimization algorithm that requires the gradient and the Hessian matrix of the predictions of Gaussian Processes. 

\subsection{Experimental Results}

We evaluate the Pareto-front approximation sets in this section using the HV indicator. In the experimental studies, Wilcoxon’s rank-sum test at a 0.05 significance level was implemented between an algorithm and its competitors to test the statistical significance. In the following tables, "+", "$\approx$", and "-" denote that an algorithm in the first row performs better than, worse than, and similar to its competitors in the first column, respectively.

\input{tab/hv_summary}

The results of all the test algorithms are summarized in Table~\ref{tab:HVComparison1}. Since all the test algorithms failed to locate a Pareto-front approximation set that dominates the reference point on mDTLZ1 and mDTLZ3 problems, the results of these two problems are not visualized or counted in the pairwise Wilcoxon's Rank-Sum test. From Table~\ref{tab:HVComparison1}, it can be observed that DGEMO and MOEA/D-ASS yield the best and the second best results among all the test algorithms w.r.t. the mean and the standard deviation (std.) of HV values. 
Between DGEMO and MOEA/D-ASS, DGEMO outperforms MOEA/D-ASS w.r.t. mean HV because DGEMO incorporates the diversity knowledge from both design and objective spaces in the batch selection. Additionally, the DGEMO's optimizer is based on a first-order approximation of the Pareto front, which can discover piecewise continuous regions of the Pareto front rather than individual points on the Pareto front to be captured~\cite{discovery}. 
 

Table~\ref{tab:sumRankSum_all} show the performance of pairwise Wilcoxon's Rank-Sum test ($+/\approx/-$) matrix among all indicator-based MOBGO algorithms test algorithms on 18 benchmarks. The sum of Wilcoxon's Rank-Sum test (sum of $+/\approx/-$) indicates $\kpoi_{\mbox{one}}$ performs best, as it significantly outperforms 78 pairwise instances between algorithms and problems. Additionally, the PoI variants ($\kpoi_i|i \in {\mbox{all}, \mbox{one}, \mbox{best}, \mbox{worst}}$) that consider correlations between multiple point predictions outperform PoI and the other two parallel techniques of PoI (KB-PoI and CL-PoI) in most cases. 

Table~\ref{tab:HVComparison1} and Table~\ref{tab:sumRankSum_lowDim} show that q-EHVI performs best on low-dimensional problems (ZDT1-3, GSP problems, and MaF problems) among all the indicator-based MOBGO algorithms. $\kpoi_{\mbox{best}}$ yields the second best results on the low-dimensional test problems. On high-dimensional problems (mDTLZ and WOSGZ problems) of which boundaries of the Pareto fronts are difficult-to-approximate (DtA), $\kpoi_{\mbox{one}}$ yields the best results w.r.t. mean HV (see Table~\ref{tab:HVComparison1}) and the pairwise Wilcoxon's Rank-Sum test in  Table~\ref{tab:sumRankSum_highDim}. Comparing all the indicator-based MOBGO algorithms, the poor performance of q-EHVI on high-dimensional problems is explained as follows. Compared to q-EHVI, the objective area involved in the computation of the PoI and its variants is larger. The computation of HV-based acquisition functions only covers the non-dominated space that dominates the reference point. Otherwise, the HV will be infinity. Introducing a reference point makes it impossible to explore boundary non-dominated solutions dominated by the reference point, that is, the space $ndom(\pfa, \infty^m) \setminus ndom(\pfa, \mathbf{r})$. On the other hand, PoI and its variants don't have this limitation, as their computations cover an entire non-dominated space. Therefore, PoI and its variants are easier to locate the boundary non-dominated solutions for problems with DtA Pareto-front boundaries. Another possible reason is that the computational error caused by the MC method can deteriorate the performance of the CMA-ES optimizer.

\begin{figure}
      \centering
      \begin{subfigure}[b]{1\textwidth}
           \centering
           \includegraphics[width=\textwidth] {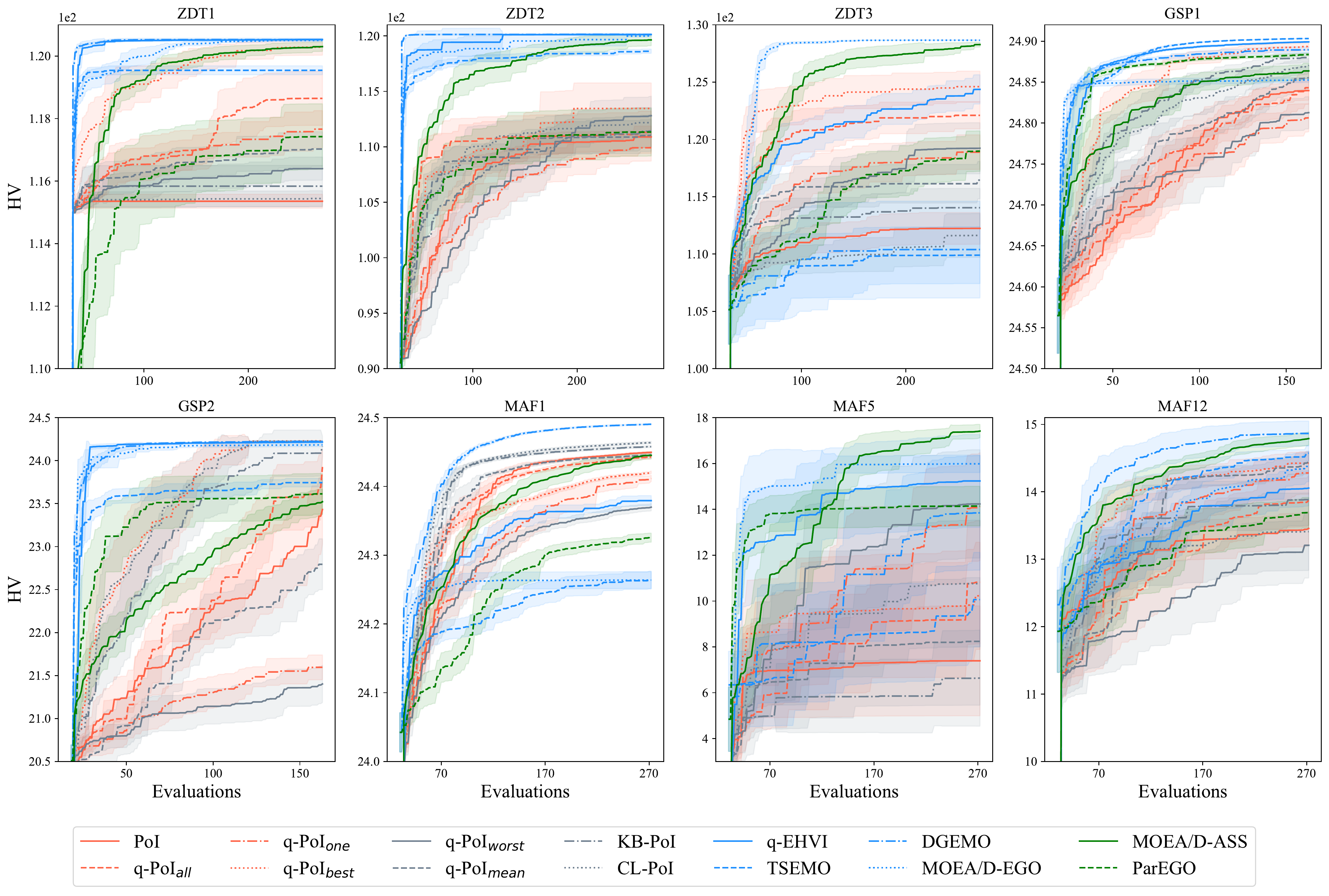}
           \caption{HV convergence plot on low-dimensional problems.}
            \label{fig:hv_low_dimension}
      \end{subfigure}
      \hfill
      \begin{subfigure}[b]{1\textwidth}
          \centering
          \includegraphics[width=\textwidth] {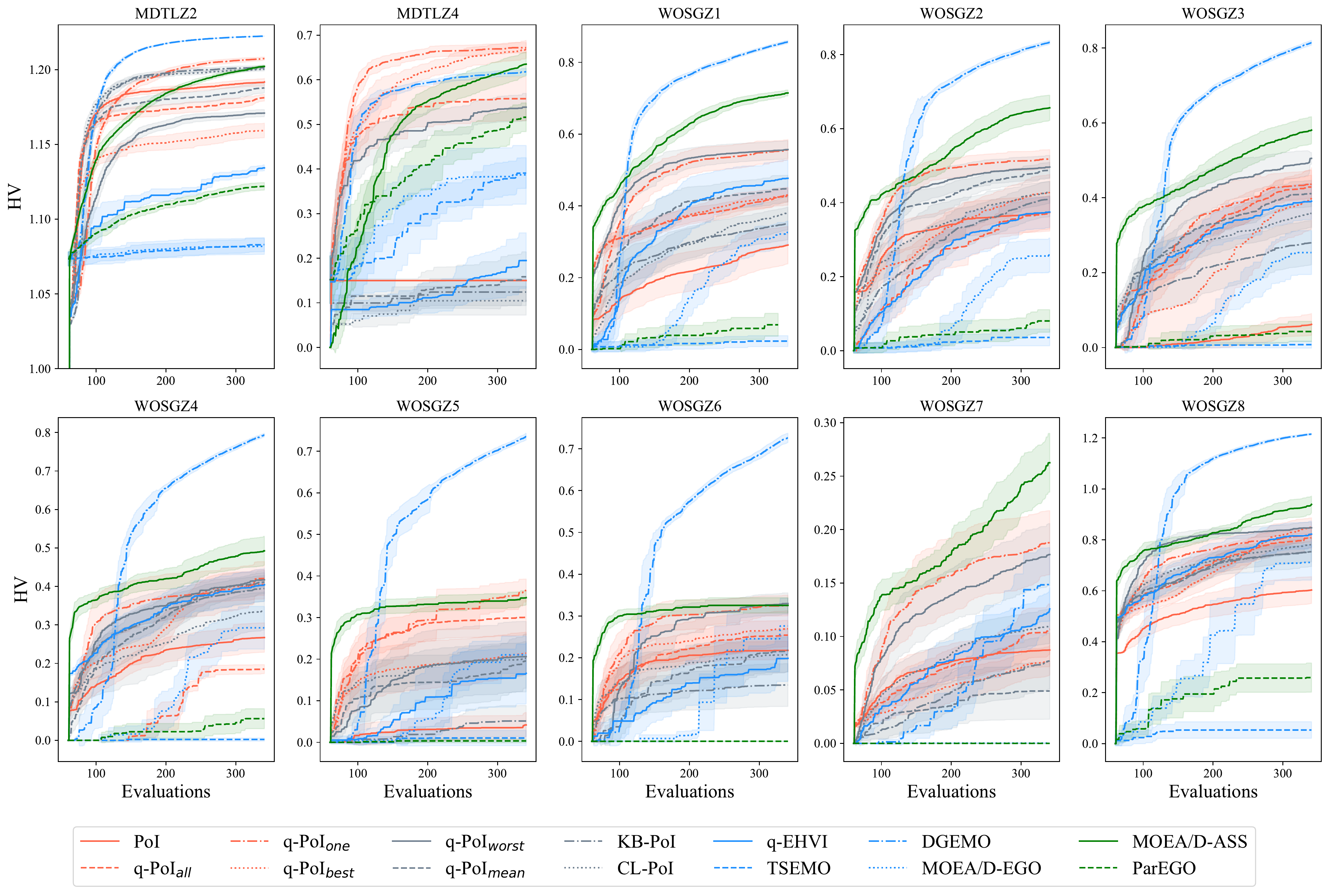}
          \caption{HV convergence plot on low-dimensional problems.}
          \label{fig:hv_high_dimension}
      \end{subfigure}
      \caption{HV convergence plots of 18 algorithms on the benchmarks are visualized in terms of the average HV over 15 independent runs.}
     \label{fig:HVConvergence}
\end{figure}

Figure~\ref{fig:HVConvergence} exhibits the average HV convergence curves of 15 independent runs of 14 algorithms on the 18 test problems. At the beginning optimization stage, DGEMO converges much faster than the other algorithms in low-dimensional problems but converges slowly in high-dimensional problems. This is because it is more difficult to quantify a credible diversity knowledge in both design and objective spaces for problems with DtA boundaries when the number of samples is small. 
The convergence of q-EHVI is fast on low-dimensional problems but is slow on high-dimensional problems. The reason is mainly because of its greedy property in theory, compared with PoI and q-PoIs. Among all the indicator-based MOBGO algorithms, $\kpoi_{\mbox{best}}$ converges second fast in low-dimensional problems, and $\kpoi_{\mbox{one}}$ converges fastest in high-dimensional problems. The reason relates to the strictness of the acquisition function. The more strict condition to fulfill is, the more greedy the acquisition function will be, and vice versa. In low-dimensional problems, $\kpoi_{\mbox{best}}$ is the most greedy due to its strict requirement to fulfill. However, this greedy strategy is efficient on low-dimensional problems. Therefore, $\kpoi_{\mbox{all}}$ and $\kpoi_{\mbox{best}}$ converge much faster than the other indicator-based MOBGO algorithms on low-dimensional test problems (see Figure~\ref{fig:hv_low_dimension}). When the test problem is complex w.r.t. the number of decision variables and the property of DtA PF boundaries, the exploration acquisition function is more effective as it is easier to jump out of the local optima. This is also the reason why $\kpoi_{\mbox{worst}}$ converges the fastest on WOSGZ1 problem (see Figure~\ref{fig:hv_high_dimension}).

\begin{figure}[]
    \centering
      \includegraphics[width=1\textwidth] {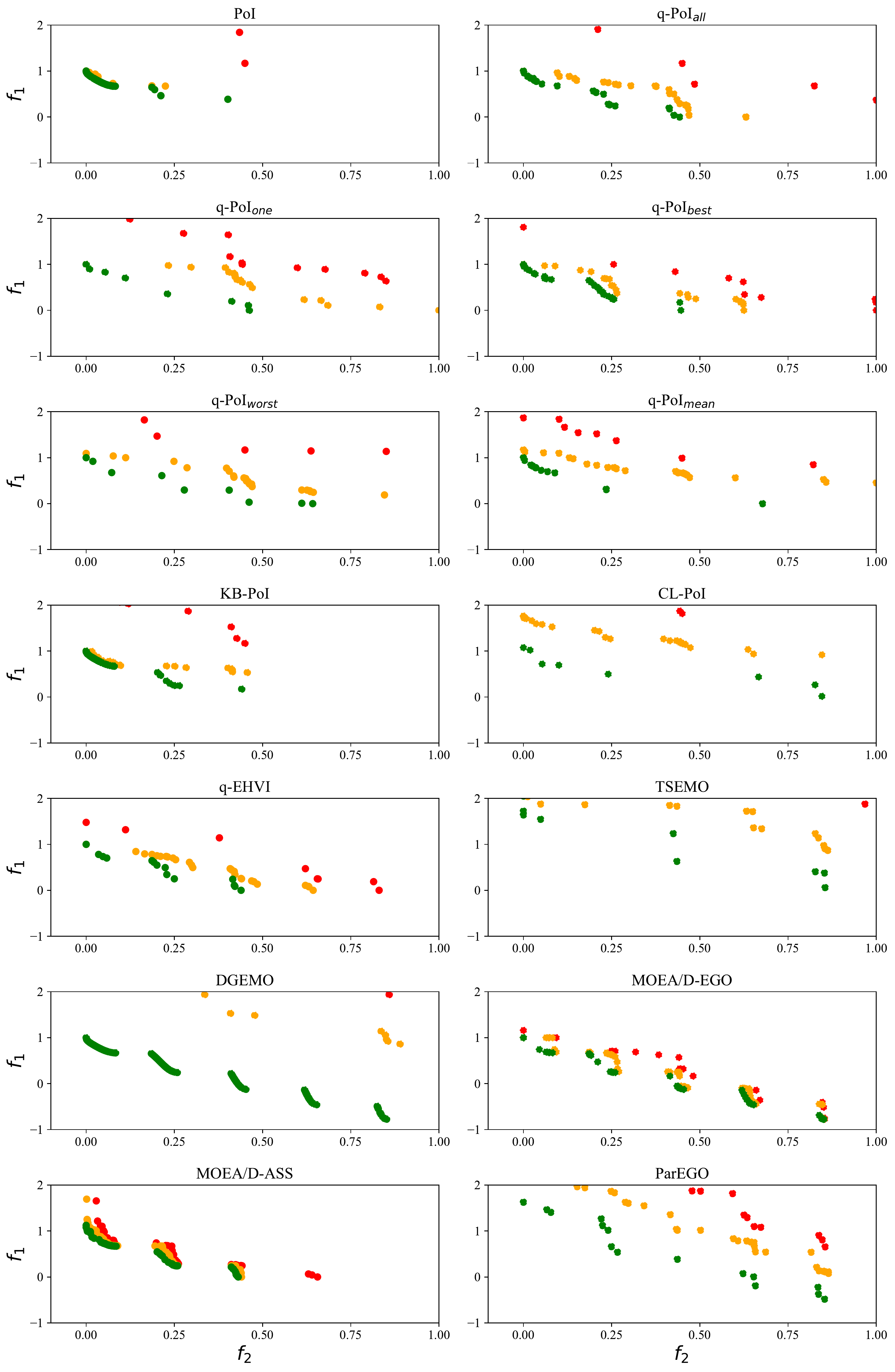}
      \caption{The best, median, and the worst empirical attainment curves on ZDT3, where \protect\greendot, \protect\orangedot, and \protect\reddot \enskip represent the best, the worst, and the median Pareto-front approximation sets over 15 independent runs, respectively.}
      \label{fig:eaf_zdt3}
\end{figure}

\begin{figure}[]
    \centering
      \includegraphics[width=\textwidth] {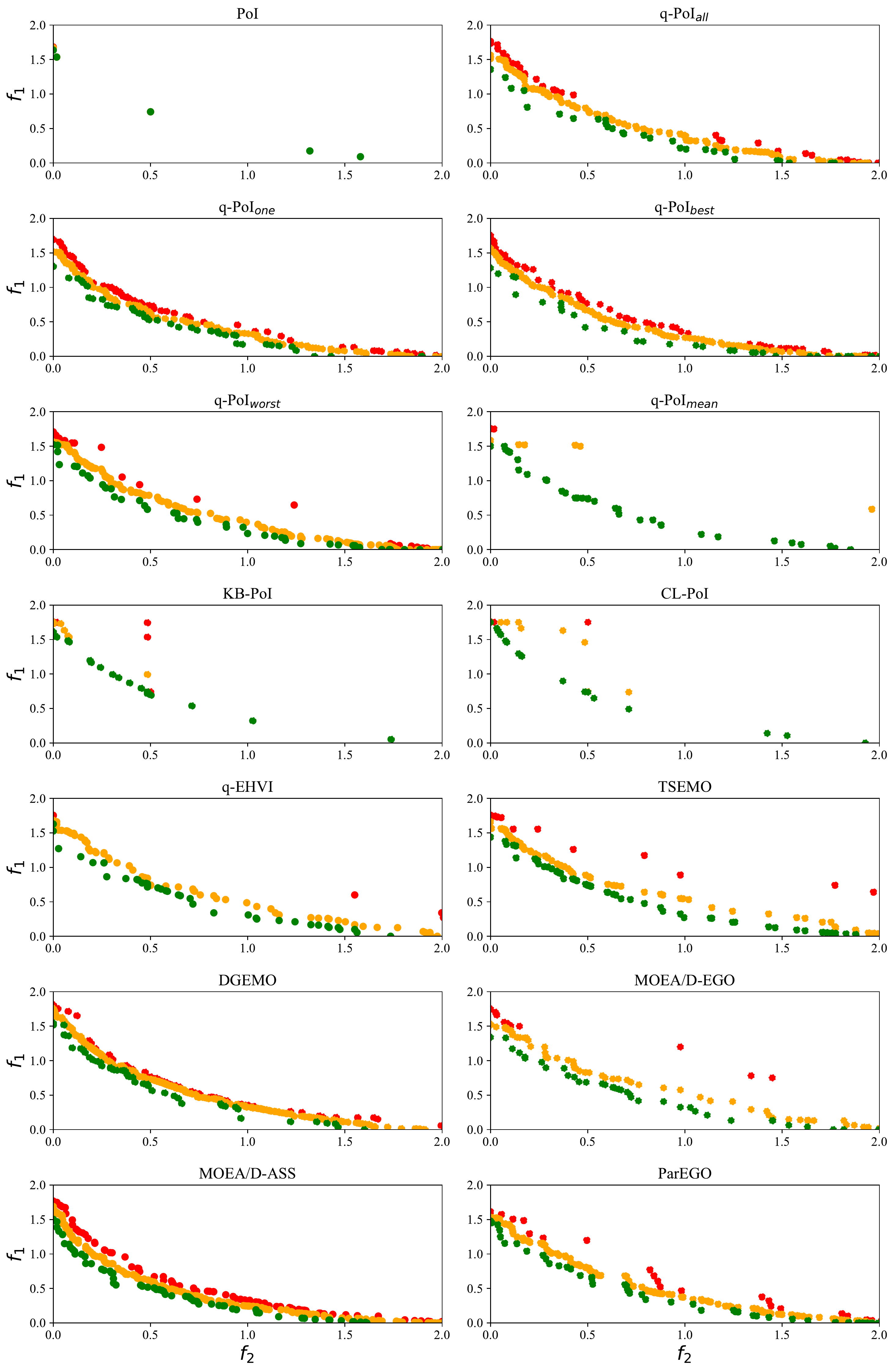}
      \caption{The best, median, and the worst empirical attainment curves on mDTLZ4, where \protect\greendot, \protect\orangedot, and \protect\reddot \enskip represent the best, the worst, and the median Pareto-front approximation sets over 15 independent runs, respectively. }
      \label{fig:eaf_mdtlz4}
\end{figure}

\begin{figure}
    \centering
      \includegraphics[width=\textwidth] {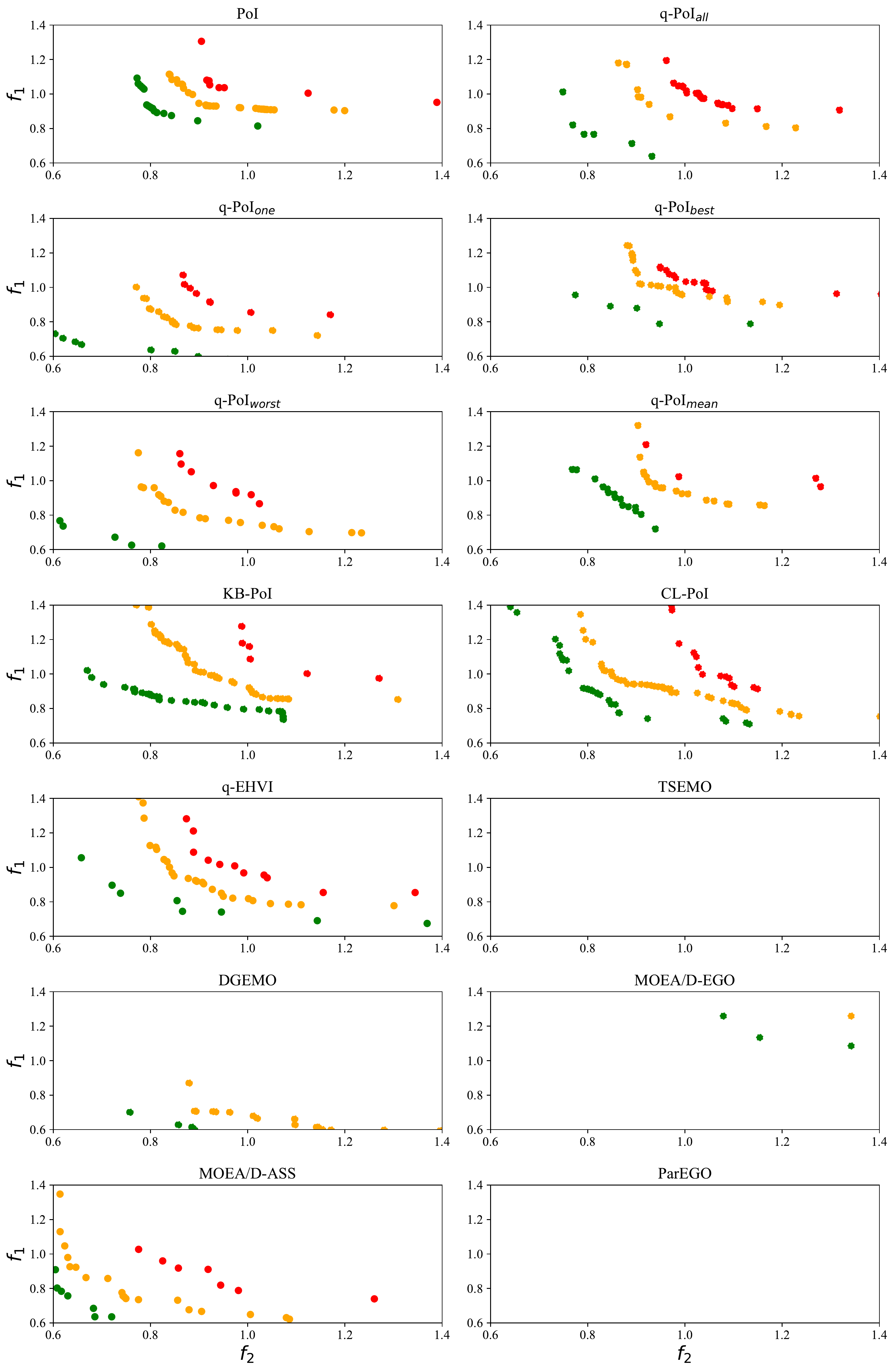}
      \caption{The best, median, and the worst empirical attainment curves on WOSGZ7, where \protect\greendot, \protect\orangedot, and \protect\reddot \enskip represent the best, the worst, and the median Pareto-front approximation sets over 15 independent runs, respectively. }
      \label{fig:eaf_wosgz7}
\end{figure}

Figure~\ref{fig:eaf_zdt3}, Figure~\ref{fig:eaf_mdtlz4} and Figure~\ref{fig:eaf_wosgz7} show the best, median, and worst empirical attainment curves of the Pareto-front approximation set on ZDT3, mDTLZ4, and WOSGZ7, respectively, by using the empirical first-order attainment function in~\cite{eaf}. The ranges of $f_1$ and $f_2$ are the same for all the algorithm's attainment curves on a specific problem. On the discontinued ZDT3 problem (see Figure~\ref{fig:eaf_zdt3}), DGEMO finds all the Pareto fronts due to the utilization of diversity knowledge from both design and objective spaces. The best attainment curve of $\kpoi_{\mbox{worst}}$ finds more Pareto fronts among all the indicator-based MOBGO algorithms because its requirement is most relaxed, and it is much more explorative than the other acquisition functions. In Figure~\ref{fig:eaf_mdtlz4}, it is easy to observe that the acquisition functions that incorporate correlation information yield better results than the other acquisition functions that don't use the correlation information. WOSGZ7 problem is more difficult than WOSGZ1-6 because of the imbalance between the middle and the boundary regions of the Pareto front~\cite{wosgz}. This problem entails more difficulty for optimization algorithms in fulfilling breadth diversity. Thus, an optimization algorithm with more exploration-property works better on WOSGZ7, which is the reason why $\kpoi_{\mbox{one}}$ and $\kpoi_{\mbox{worst}}$ work much better than the other indicator-based algorithms and DGEMO. The explanation for the poor performance of DGEMO on WOSGZ7 is that credible diversity knowledge of this kind of problem requires more samples. This also explains the reason for the fast convergence of DGEMO at the end of the optimization stage on WOSGZ7 in Figure~\ref{fig:hv_high_dimension}. 

\begin{table}[]
\caption{The pairwise Wilcoxon's Rank-Sum test  ($+/\approx/-$) matrix at a 0.05 significance level was performed among nine indicator-based MOBGO algorithms, where algorithms in all the columns (except the first column) are compared with the algorithms in the first column.}
\label{tab:sumRankSum_all}
\begin{adjustbox}{width=\columnwidth,center}
    \begin{tabular}{c|ccccccccc}
    \toprule
        ~ & PoI   & $\kpoi_{\mbox{all}}$ & $\kpoi_{\mbox{one}}$ & $\kpoi_{\mbox{best}}$ & $\kpoi_{\mbox{worst}}$ & $\kpoi_{\mbox{mean}}$ & KB-PoI & CL-PoI  & q-EHVI  \\ \hline
        PoI & 0/0/0 & 9/6/3 & 14/1/3 & 11/5/2 & 12/2/4 & 9/6/3 & 8/9/1 & 9/8/1 & 12/4/2  \\ 
        $\kpoi_{\mbox{all}}$ & 3/7/8 & 0/0/0 & 8/5/5 & 8/6/4 & 7/4/7 & 5/7/6 & 5/5/8 & 5/6/7 & 6/7/5  \\ 
        $\kpoi_{\mbox{one}}$ & 3/1/14 & 5/5/8 & 0/0/0 & 6/5/7 & 2/9/7 & 3/7/8 & 3/3/12 & 3/3/12 & 5/5/8  \\ 
        $\kpoi_{\mbox{best}}$& 2/5/11 & 4/6/8 & 7/5/6 & 0/0/0 & 7/4/7 & 3/8/7 & 2/5/11 & 3/6/9 & 5/6/7  \\ 
        $\kpoi_{\mbox{worst}}$& 4/2/12 & 8/3/7 & 7/9/2 & 7/4/7 & 0/0/0 & 6/4/8 & 5/3/10 & 4/4/10 & 8/3/7  \\ 
        $\kpoi_{\mbox{mean}}$ & 4/5/9 & 6/7/5 & 8/7/3 & 7/8/3 & 9/3/6 & 0/0/0 & 5/8/5 & 4/7/7 & 7/7/4  \\ 
        KB-PoI & 1/9/8 & 9/4/5 & 12/3/3 & 11/5/2 & 10/3/5 & 5/8/5 & 0/0/0 & 3/12/3 & 8/7/3  \\ 
        CL-PoI & 1/8/9 & 7/6/5 & 14/1/3 & 9/6/3 & 10/4/4 & 7/7/4 & 3/13/2 & 0/0/0 & 8/6/4  \\ 
        q-EHVI & 2/4/12 & 5/7/6 & 8/5/5 & 7/6/5 & 7/3/8 & 4/7/7 & 3/7/8 & 4/7/7 & 0/0/0  \\ \midrule
    Sum of $+/\approx/-$ & 20/41/83	 & 53/44/47	 & \textbf{78/36/30}	 & 66/45/33	 & 64/32/48	 & 42/54/48	 & 34/53/57	 & 35/53/56	 & 59/45/40 \\
    \bottomrule
    \end{tabular}
\end{adjustbox}
\end{table}

\clearpage
\section{Conclusion}
This paper proposed five alternative acquisition functions by generalizing \poi from a single point into a batch with multiple points. For each proposed \kpoi, explicit computational formulas and the MC method for approximation are provided. Computational complexities of the exact computational formula are given. Among the five proposed q-PoIs, $\kpoi_{\mbox{one}}$ processes the highest computational complexity as it has to compute the sum of \poi for each single solution and $\kpoi_{\mbox{all}}$, while $\kpoi_{\mbox{mean}}$ only considers the diagonal elements in covariance matrix $\boldsymbol{\Sigma}$. All the proposed \kpoi leave spaces for parallel techniques in MOBGO to evaluate multiple solutions in each batch simultaneously. Some of the behaviors of q-PoIs and their connections to the standard deviation matrix and correlation coefficient are analyzed for the five proposed q-PoI variants in three cases. 

This paper also compares the performance of the five proposed acquisition functions with that of PoI, two parallel techniques for PoI, q-EHVI, and five other batch-selection surrogate-assisted multi-objective optimization algorithms on 20 benchmarks. The experiments show that the acquisition functions with more greedy characteristics (q-EHVI, $\kpoi_{\mbox{all}}$, and $\kpoi_{\mbox{best}}$) perform much better than the other acquisition functions on low-dimensional test problems. For high-dimensional  problems, the acquisition functions ($\kpoi_{\mbox{one}}$ and $\kpoi_{\mbox{worst}}$) with more explorative properties outperform the other acquisition functions, especially on the problems with DtA Pareto front boundaries. DGEMO outperforms the other test algorithms, but its convergence is slower than most of the proposed q-PoI variants in this paper on the problems with DtA Pareto front boundaries. 

The experimental results confirm that strict requirements in q-PoIs' definitions can refine algorithms' exploitation property. Due to the strict condition in $\kpoi_{\mbox{best}}$ and $\kpoi_{\mbox{all}}$, the algorithm behaves greedily. This exploitative behavior allows MOBGO to quickly converge to the true Pareto front at the early optimization stage. Still, it renders MOBGO performance when the solutions of the Pareto front are highly biased or the problem's landscape has DtA PF boundaries. On the other hand, relaxed requirements of $\kpoi_{one}$ and $\kpoi_{worst}$ can enhance the exploration property of PoI, which purely focuses on exploration. 
Strategically, one can use strict acquisition functions $\kpoi_{\mbox{best}}$ at the early stage of iteration and then switch to a relaxed acquisition function $\kpoi_{\mbox{one}}$ or $\kpoi_{\mbox{worst}}$ when the HV convergence velocity starts to slow down during the optimization processes on low-dimensional problems. On high-dimensional problems, especially those with DtA Pareto front boundaries, it is recommended to use $\kpoi_{\mbox{one}}$ or $\kpoi_{\mbox{worst}}$. 

The proposed algorithms can be applied to many real-world applications involving expensive simulations, including box-type boom designing problem~\cite{karder2018gecco}, material flow optimization problems~\cite{Affenzeller2015}, algorithm selection problems~\cite{werth2020surrogate}, and hyperparameter tuning problems in the machine learning field, just to name a few. For future work, it is worthwhile to study the multi-objective acquisition functions that consider the correlations among each coordinate using the multi-output Gaussian process, as all of the existing multi-objective acquisition functions simply assume the independence among each objective. It is also interesting to incorporate the diversity control mechanism in indicator-based MOBGO algorithms due to its effectiveness shown in DGEMO. An intuitive solution is to use truncated normal distribution in PoI and its variants to split objective space into several subregions and then search for the optima in each subregion.




\section*{Appendix}
\subsection*{A.1 Explicit formulas of PoIs for high-dimensional problems}

Note that the explicit formulas for computing q-PoIs in the case of $m$ objectives for $m>2$ can be easily generalized from the definition and the formulas for $m=2$ provided in Section \ref{subsec:kpoi_definitions}.
For completeness, we list the formulas for the case $m>2$:
\begin{align*}
&\kpoi_{\mbox{all}}(M, \boldsymbol{\Sigma}, \pfa) 
= 
\int_{\mathbb{R}^{m \times 2}} \mathrm{I}(\hat{\boldsymbol{y}}^{(1)} 
\mbox{ impr } \pfa ) \cap  \mathrm{I}(\hat{\boldsymbol{y}}^{(2)} \mbox{ impr } \pfa ) 
\boldsymbol{\pdf}_{M, \boldsymbol{\Sigma}}(\hat{\mathbf{Y}}) d\hat{\mathbf{Y}} \\
&= \sum_{j=1}^{N_m}\sum_{jj=1}^{N_m} \Big(
\int_{(l_1^{(j)},l_1^{(jj)})}^{(u_1^{(j)}, u_1^{(jj)})}
\boldsymbol{\pdf}_{\boldsymbol{\mu}_1, \Sigma_1}(\hat{\mathbf{Y}}_1) d\hat{\mathbf{Y}}_1 \times \cdots \times 
\int_{(l_m^{(j)},l_m^{(jj)})}^{(u_m^{(j)}, u_m^{(jj)})}
\boldsymbol{\pdf}_{\boldsymbol{\mu}_m, \Sigma_m}(\hat{\mathbf{Y}}_m) d\hat{\mathbf{Y}}_m
\Big) \\
&= \sum_{j=1}^{N_m}\sum_{jj=1}^{N_m} \prod_{i=1}^{m}\Gamma(l_i^{(j)},u_i^{(j)},l_i^{(jj)},u_i^{(jj)},\boldsymbol{\mu}_i,\Sigma_i) \numberthis
\end{align*}

\begin{align*}
&\kpoi_{\mbox{one}}(M, \boldsymbol{\Sigma}, \pfa) 
= 
\int_{\mathbb{R}^{m \times 2}} \mathrm{I}(\hat{\boldsymbol{y}}^{(1)} \mbox{ impr } \pfa ) \cup 
\mathrm{I}(\hat{\boldsymbol{y}}^{(2)} \mbox{ impr } \pfa ) 
\boldsymbol{\pdf}_{M, \boldsymbol{\Sigma}}(\hat{\mathbf{Y}}) d\hat{\mathbf{Y}} \\
& = \Big( \sum_{j=1}^{q=2}
\int_{\mathbb{R}^m} \mathrm{I}(\hat{\mathbf{y}}^{(j)} \mbox{ impr } \pfa ) \boldsymbol{\pdf}_{\boldsymbol{\mu}^{(j)}, \boldsymbol{s}^{(j)}}(\hat{\mathbf{y}}^{(j)}) d\hat{\mathbf{y}}^{(j)}\Big) -
\kpoi_{\mbox{all}} (M, \boldsymbol{\Sigma}, \pfa)
\numberthis
\label{eq:exact_kpoi_one_highD}
\end{align*}

\begin{align*}
&\kpoi_{\mbox{worst}} (M, \boldsymbol{\Sigma}, \pfa) 
= \int_{\mathbb{R}^{m \times 2}} \big(\mathrm{I}(\mbox{ConJ}(\hat{\boldsymbol{Y}}) \mbox{ impr } \pfa \big) 
\boldsymbol{\pdf}_{M, \boldsymbol{\Sigma}}(\hat{\mathbf{Y}}) d\hat{\mathbf{Y}} \\
&= \sum_{j=1}^{N_m} \Big(
\left(\int_{(l_1^{(j)},u_1^{(j)})}^{(u_1^{(j)},+\infty)} + \int_{(l_1^{(j)},l_1^{(j)})}^{+\infty,u_1^{(j)})}\right)
\boldsymbol{\pdf}_{\boldsymbol{\mu}_1, {\Sigma}_1}(\hat{\mathbf{Y}}_1) d\hat{\mathbf{Y}}_1 \times \cdots \times \\
& \quad \qquad \enskip
\left(\int_{(l_m^{(j)},u_m^{(j)})}^{(u_m^{(j)},+\infty)} + \int_{(l_m^{(j)},l_m^{(j)})}^{+\infty,u_m^{(j)})}\right)
\boldsymbol{\pdf}_{\boldsymbol{\mu}_m, {\Sigma}_m}(\hat{\mathbf{Y}}_m) d\hat{\mathbf{Y}}_m
 \Big)\\
& = \sum_{j=1}^{N_m}\prod_{i=1}^{m}
\Big\{ \left( \mbox{MCDF}\big( (u_i^{(j)}, \infty), \boldsymbol{\mu}_i, \Sigma_i \big) + \mbox{MCDF}\big( (\infty,   u_i^{(j)}), \boldsymbol{\mu}_i, \Sigma_i \big)  \right) \\
& \quad \qquad \quad  - \left( 
\mbox{MCDF}\big( (\infty, l_i^{(j)}), \boldsymbol{\mu}_i, \Sigma_i \big)
 + \mbox{MCDF}\big( (l_i^{(j)},\infty), \boldsymbol{\mu}_i, \Sigma_i \big) \right) \\
&\quad \qquad \quad - \left(\mbox{MCDF}\big( (u_i^{(j)},u_i^{(j)}), \boldsymbol{\mu}_i, \Sigma_i \big) - \mbox{MCDF}\big( (l_i^{(j)},l_i^{(j)}), \boldsymbol{\mu}_i, \Sigma_i \big) \right) \Big\} \numberthis 
\label{def:exact_kpoi_worst_high}
\end{align*}

\begin{align*}
&\kpoi_{\mbox{best}} (M, \boldsymbol{\Sigma}, \pfa) = \int_{\mathbb{R}^{m \times 2}} \mathrm{I}(\mbox{DisC}(\hat{\boldsymbol{Y}}) \mbox{impr } \pfa )
\boldsymbol{\pdf}_{M, \boldsymbol{\Sigma}}(\hat{\mathbf{Y}}) d\hat{\mathbf{Y}} \\
&= \int_{\hat{\mathbf{Y}}_1=(-\infty, -\infty)}^{(\infty, \infty)}\cdots 
\int_{\hat{\mathbf{Y}}_m=(-\infty, -\infty)}^{(\infty, \infty)}
\big( \mathrm{I}( \hat{\mathbf{Y}}_1 \lor \cdots \lor \hat{\mathbf{Y}}_m ) \mbox{ impr } \pfa \big)  
\boldsymbol{\pdf}_{M, \boldsymbol{\Sigma}}(\hat{\mathbf{Y}}) d\hat{\mathbf{Y}} \\
&= \sum_{j=1}^{N_m} \prod_{i=1}^{m} \Big( \mbox{MCDF}\big( (u_i^{(j)},u_i^{(j)}), \boldsymbol{\mu}_i, \Sigma_i \big) - \mbox{MCDF}\big( (l_i^{(j)},l_i^{(j)}), \boldsymbol{\mu}_i, \Sigma_i \big) \Big)
\numberthis
\label{def:exact_kpoi_best_high}
\end{align*}

\begin{align*}
\kpoi_{\mbox{mean}} (M, \boldsymbol{\Sigma}, \pfa) 
& = \frac{1}{q} \sum_{i=1}^{q} \sum_{j=1}^{N_m}
\int_{l_1^{(j)}}^{u_1^{(j)}}\pdf_{\mu_1^{(i)},s_1^{(i)}} d y_1 \cdots 
\int_{l_m^{(j)}}^{u_m^{(j)}}\pdf_{\mu_m^{(i)},s_m^{(i)}} d y_m
\numberthis
\label{eq:exact_kpoi_mean_high}
\end{align*}

In the formulas above, $N_m$ is the number of decomposed non-dominated areas in m-dimensional objective space. By using the methods in~\cite{yang2019efficient}, this number can be reduced into $n+1$ and $2n+1$ for bi- and tri-objective spaces, respectively. 

Please also note that all the q-PoIs formulas for high-dimensional objective spaces in this appendix assume that no overlapped stripes/cells/boxes exist in a dominated space. Therefore, the grid decomposition method and the decomposition methods for $m=2,3$ in~\cite{yang2019efficient} work on these q-PoI formulas. However, it is not recommended to use the grid decomposition method due to its high computational complexity introduced. The decomposition method for $m\geq 4$ mentioned in \cite{yang2019efficient} can not be directly applied to the q-PoI formulas in this appendix. One has to subtract the integral of the overlapped cells by using the decomposition method that exists in overlapped cells. 
\subsection*{A.2 Tables}
\begin{table}[!h]
\caption{Low-dimensional case (ZDT1-3, GSP1-2, MaF1/5/12): The pairwise Wilcoxon's Rank-Sum test  ($+/\approx/-$) matrix at a 0.05 significance level was performed among \textbf{nine indicator-based MOBGO algorithms}, where algorithms in all the columns (except the first column) are compared with the algorithms in the first column.}
\label{tab:sumRankSum_lowDim}
\begin{adjustbox}{width=\columnwidth,center}
    \begin{tabular}{c|ccccccccc}
    \toprule
        ~ & PoI   & $\kpoi_{\mbox{all}}$ & $\kpoi_{\mbox{one}}$ & $\kpoi_{\mbox{best}}$ & $\kpoi_{\mbox{worst}}$ & $\kpoi_{\mbox{mean}}$ & KB-PoI & CL-PoI  & q-EHVI  \\ 
        \midrule
                PoI & 0/0/0 & 4/3/1 & 4/1/3 & 5/2/1 & 4/1/3 & 4/2/2 & 4/4/0 & 4/4/0 & 7/0/1  \\ 
         $\kpoi_{\mbox{all}}$ & 1/4/3 & 0/0/0 & 0/3/5 & 5/2/1 & 0/3/5 & 2/3/3 & 3/3/2 & 3/3/2 & 5/2/1  \\ 
         $\kpoi_{\mbox{one}}$ & 3/1/4 & 5/3/0 & 0/0/0 & 6/1/1 & 1/3/4 & 3/4/1 & 3/2/3 & 3/2/3 & 5/2/1  \\ 
         $\kpoi_{\mbox{best}}$ & 1/2/5 & 1/2/5 & 1/1/6 & 0/0/0 & 1/1/6 & 1/3/4 & 1/2/5 & 1/2/5 & 4/1/3  \\ 
         $\kpoi_{\mbox{worst}}$ & 3/1/4 & 5/3/0 & 4/3/1 & 6/1/1 & 0/0/0 & 5/1/2 & 4/1/3 & 3/3/2 & 8/0/0  \\ 
         $\kpoi_{\mbox{mean}}$ & 2/2/4 & 3/3/2 & 1/4/3 & 4/3/1 & 3/0/5 & 0/0/0 & 4/2/2 & 2/3/3 & 6/0/2  \\ 
        KB-PoI & 0/4/4 & 2/3/3 & 3/2/3 & 5/2/1 & 3/1/4 & 2/2/4 & 0/0/0 & 2/4/2 & 5/1/2  \\ 
        CL-PoI & 0/4/4 & 2/3/3 & 4/1/3 & 5/2/1 & 2/3/3 & 3/3/2 & 2/5/1 & 0/0/0 & 5/1/2  \\ 
        q-EHVI & 1/0/7 & 1/2/5 & 1/2/5 & 3/1/4 & 0/0/8 & 2/0/6 & 2/1/5 & 2/1/5 & 0/0/0  \\ 
        \midrule
        Sum of $+/\approx/-$ & 11/18/35 & 23/22/19 & 18/17/29 & 39/14/11 & 14/12/38 & 22/18/24 & 23/20/21 & 20/22/22 & \textbf{45/7/12} \\ 
    \bottomrule
    \end{tabular}
\end{adjustbox}
\end{table}

\begin{table}[!h]
\caption{High-dimensional case (mDTLZ1-4, WOSGZ1-8): The pairwise Wilcoxon's Rank-Sum test  ($+/\approx/-$) matrix at a 0.05 significance level was performed among \textbf{nine indicator-based MOBGO algorithms}, where algorithms in all the columns (except the first column) are compared with the algorithms in the first column.}
\label{tab:sumRankSum_highDim}
\begin{adjustbox}{width=\columnwidth,center}
    \begin{tabular}{c|ccccccccc}
    \toprule
        ~ & PoI   & $\kpoi_{\mbox{all}}$ & $\kpoi_{\mbox{one}}$ & $\kpoi_{\mbox{best}}$ & $\kpoi_{\mbox{worst}}$ & $\kpoi_{\mbox{mean}}$ & KB-PoI & CL-PoI  & q-EHVI  \\ 
        \midrule
        PoI & 0/0/0 & 5/3/2 & 10/0/0 & 6/3/1 & 8/1/1 & 5/4/1 & 4/5/1 & 5/4/1 & 5/4/1  \\ 
        $\kpoi_{\mbox{all}}$ & 2/3/5 & 0/0/0 & 8/2/0 & 3/4/3 & 7/1/2 & 3/4/3 & 2/2/6 & 2/3/5 & 1/5/4  \\ 
        $\kpoi_{\mbox{one}}$ & 0/0/10 & 0/2/8 & 0/0/0 & 0/4/6 & 1/6/3 & 0/3/7 & 0/1/9 & 0/1/9 & 0/3/7  \\ 
        $\kpoi_{\mbox{best}}$ & 1/3/6 & 3/4/3 & 6/4/0 & 0/0/0 & 6/3/1 & 2/5/3 & 1/3/6 & 2/4/4 & 1/5/4  \\ 
        $\kpoi_{\mbox{worse}}$ & 1/1/8 & 3/0/7 & 3/6/1 & 1/3/6 & 0/0/0 & 1/3/6 & 1/2/7 & 1/1/8 & 0/3/7  \\ 
        $\kpoi_{\mbox{mean}}$ & 2/3/5 & 3/4/3 & 7/3/0 & 3/5/2 & 6/3/1 & 0/0/0 & 1/6/3 & 2/4/4 & 1/7/2  \\ 
        KB-PoI & 1/5/4 & 7/1/2 & 9/1/0 & 6/3/1 & 7/2/1 & 3/6/1 & 0/0/0 & 1/8/1 & 3/6/1  \\ 
        CL-PoI & 1/4/5 & 5/3/2 & 10/0/0 & 4/4/2 & 8/1/1 & 4/4/2 & 1/8/1 & 0/0/0 & 3/5/2  \\ 
        q-EHVI & 1/4/5 & 4/5/1 & 7/3/0 & 4/5/1 & 7/3/0 & 2/7/1 & 1/6/3 & 2/6/2 & 0/0/0  \\ 
        \midrule
        Sum of $+/\approx/-$  & 9/23/48 & 30/22/28 & \textbf{60/19/1} & 27/31/22 & 50/20/10 & 20/36/24 & 11/33/36 & 15/31/34 & 14/38/28 \\ 
    \bottomrule
    \end{tabular}
\end{adjustbox}
\end{table}

\bibliographystyle{elsart-num}
\bibliography{new_bib.bib}


\end{document}

%% file: package.tex
\usepackage{booktabs}
\usepackage{multirow}
\usepackage{longtable}
\usepackage[counterclockwise, figuresright]{rotating}
\usepackage{graphicx}
\usepackage{adjustbox}
\usepackage[normalem]{ulem}

\usepackage{lipsum}
\usepackage{algorithmicx}
\usepackage[linesnumbered,ruled,vlined]{algorithm2e}
\usepackage{algpseudocode}

\usepackage{tikz}
\usepackage{bm}

\usepackage{bbm}

\usepackage{graphicx}
\usepackage{pgfplots}
\usepackage{tikz}
\usepackage{amssymb}
\usepackage{graphics}
\usepackage{graphicx}
\usepackage{url}
\usepackage{theorem}
\usepackage{booktabs}
\usepackage{multirow}

\usepackage {longtable}
\usepackage{tikz}
\usetikzlibrary{patterns}
\pgfplotsset{compat=newest}
\usepgfplotslibrary{units}
\usepackage{booktabs}
\usepackage{multirow}
\usepackage{caption}
\usepackage{subcaption}
\usepackage{times}
\usepackage{fancyhdr,graphicx,amsmath,amssymb}
\pgfplotsset{
colormap={whitered}{color(0cm)=(white); color(1cm)=(green!75!red)}
}

\usepackage{amsmath}
\usepackage[toc]{appendix}
\numberwithin{equation}{section}

\renewcommand{\theequation}{\arabic{section}-\arabic{equation}}        
\def\Item$#1${\item $\displaystyle#1$ 
    \hfill\refstepcounter{equation}(\theequation)}                     

\usepackage{setspace}

\usepackage{mathtools}

\SetKwFunction{FRecurs}{FnRecursive}%

\usepackage{pgfplots}
\pgfplotsset{compat=newest}
\usetikzlibrary{plotmarks}
\usepgfplotslibrary{patchplots}
\usepackage{grffile}
\usepackage{amsmath}
\usepackage{commath}

\usepackage{adjustbox}

\usepackage{rotating}
\usepackage{pdflscape}

\usepackage{hyperref}
\hypersetup{
    colorlinks=true,
    linkcolor=blue,
    filecolor=magenta,      
    urlcolor=cyan,
}
 
\usepackage{esvect}

%% file: definitions.tex
\newcommand{\pfa}{\mathcal{PF}}            		
\newcommand{\pdf}{\xi} 							
\newcommand{\vy}{\mathbf{y}} 					

\usepackage{mathrsfs}

\theoremstyle{definition}
\newtheorem{definition}{Definition}[section]

\theoremstyle{example}
\newtheorem{example}{Example}[section]

\DeclareMathAlphabet\mathbfcal{OMS}{cmsy}{b}{n} 
\newcommand{\argmax}{\text{arg\,max}}
\newcommand{\af}{\mathscr{A}}          
\newcommand{\M}{\mathcal{M}}          
\newcommand{\E}{\operatorname{E}}             
\newcommand{\Cov}{\operatorname{Cov}}
\newcommand{\poi}{\mbox{PoI }}					
\newcommand{\kpoi}{\mbox{q-PoI }}					
\newcommand{\ehvi}{\mbox{EHVI }} 				
\newcommand{\GP}{\mathcal{GP}}                  
\newcommand{\X}{\ensuremath{\mathcal{X}}}       
\newcommand\conProb[2]{\ensuremath{p(#1\;|\;#2)}}     

\newcommand{\BF}[1]{
	\relax
	\ifmmode
	\ifcat\noexpand#1\relax 
		\boldsymbol{#1}     
	\else
		\mathbf{#1}
	\fi
	\else
		\textbf{#1}
	\fi
}

\newcommand\numberthis{\addtocounter{equation}{1}\tag{\theequation}}  			

\makeatletter
\newcommand*{\centernot}{%
  \mathpalette\@centernot
}

\def\@centernot#1#2{%
  \mathrel{%
    \rlap{%
      \settowidth\dimen@{$\m@th#1{#2}$}%
      \kern.5\dimen@
      \settowidth\dimen@{$\m@th#1=$}%
      \kern-.5\dimen@
      $\m@th#1\not$%
    }%
    {#2}%
  }%
}
\makeatother
\newcommand{\independent}{\perp\mkern-9.5mu\perp}
\newcommand{\notindependent}{\centernot{\independent}}

\definecolor{indiagreen}{rgb}{0.07, 0.53, 0.03}
\newcommand{\reddot}{\raisebox{0.6pt}{\tikz{\fill[red!100] (0,0) circle (0.08)}}}
\newcommand{\greendot}{\raisebox{0.6pt}{\tikz{\fill[indiagreen!100] (0,0) circle (0.08)}}}
\newcommand{\orangedot}{\raisebox{0.6pt}{\tikz{\fill[orange!100] (0,0) circle (0.08)}}}

%% file: tab/hv_summary.tex
\clearpage

\begin{sidewaystable}[]
\caption{Empirical Comparisons w.r.t HV.}
\label{tab:HVComparison1}
\centering
\begin{adjustbox}{width=20cm ,center, angle = 0} 
    \begin{tabular}{c|c|ccccccccc|ccccc}
    \toprule
                        Problem &   HV & PoI     & $\kpoi_{\mbox{all}}$  & $\kpoi_{\mbox{one}}$  & $\kpoi_{\mbox{best}}$  & $\kpoi_{\mbox{worst}}$  & $\kpoi_{\mbox{mean}}$  & {KB-PoI}   &{CL-PoI}   & {q-EHVI}     & {TSEMO}    & {DGEMO}     & {MOEA/D-EGO} & {MOEA/D-ASS} & {ParEGO} \\
    \midrule
\multirow{5}{*}{ZDT1}   & min    & 115.14503        & 115.77463          & 116.28355          & 118.59081           & 115.42380            & 115.85917           & 115.14964         & 115.15875         & \textbf{120.48504} & 118.91994         & \textbf{120.34474} & 120.31127           & 119.65315           & 113.71321       \\
                        & max    & 116.81854        & \textbf{120.61182} & 120.51466          & 120.59439           & 118.80349            & 119.32360           & 120.42198         & 117.13656         & 120.56857          & 119.85883         & \textbf{120.65306} & 120.58538           & 120.60104           & 119.79789       \\
                        & median & 115.16934        & 117.80598          & 117.22570          & 120.48251           & 116.34874            & 116.67106           & 115.31758         & 115.24741         & \textbf{120.52853} & 119.66739         & \textbf{120.54432} & 120.51459           & 120.40045           & 117.66237       \\
                        & mean   & 115.35709        & 118.64729          & 117.66193          & 120.28644           & 116.40030            & 117.02513           & 115.83834         & 115.44276         & \textbf{120.52854} & 119.54490         & \textbf{120.52809} & 120.50140           & 120.30248           & 117.42270       \\
                        & std.   & 0.46313          & 1.79193            & 1.23041            & 0.53826             & 0.76735              & 1.14907             & 1.37190           & 0.51547           & \textbf{0.02122}   & 0.29587           & 0.09555            & \textbf{0.06778}    & 0.31476             & 2.17707         \\
    \midrule
\multirow{5}{*}{ZDT2}   & min    & 108.41310        & 106.89833          & 104.87284          & 106.23683           & 104.75706            & 107.77524           & 109.11479         & 110.00000         & \textbf{120.00000} & 118.03388         & \textbf{120.01520} & 119.61446           & 116.78006           & 103.65996       \\
                        & max    & 116.03197        & 120.14233          & 114.64048          & 119.93597           & 119.61804            & 119.99851           & 120.08633         & 120.23037         & \textbf{120.23416} & 119.20350         & \textbf{120.32285} & 120.16556           & 120.27915           & 117.94353       \\
                        & median & 110.29497        & 110.02323          & 110.00000          & 112.90758           & 112.35153            & 110.00000           & 110.32229         & 111.43758         & \textbf{120.14257} & 118.53896         & \textbf{120.11913} & 119.99999           & 120.07684           & 111.42846       \\
                        & mean   & 110.89832        & 111.31122          & 109.91951          & 113.46063           & 112.80778            & 111.02350           & 111.45247         & 112.17912         & \textbf{120.12729} & 118.61616         & \textbf{120.13637} & 119.99545           & 119.63587           & 111.33057       \\
                        & std.   & 2.01997          & 3.92365            & 2.50280            & 4.76649             & 3.65075              & 2.98120             & 3.00725           & 2.82967           & \textbf{0.06296}   & 0.39413           & \textbf{0.10741}   & 0.12187             & 1.08313             & 4.47137         \\
    \midrule
\multirow{5}{*}{ZDT3}   & min    & 107.61774        & 116.69323          & 113.60479          & \textbf{120.49864}  & 108.23819            & 111.46628           & 107.80324         & 107.73242         & 120.00000          & 100.34768         & 99.27106           & \textbf{128.58877}  & 127.00861           & 113.11570       \\
                        & max    & 116.57698        & 125.55716          & 123.36041          & \textbf{128.44041}  & 128.00195            & 123.81246           & 118.75809         & 119.67210         & 128.34440          & 118.92054         & \textbf{128.77537} & 128.70522           & 128.68665           & 125.05329       \\
                        & median & 113.52522        & 121.73447          & 119.35698          & 125.24085           & 118.22064            & 115.61672           & 113.92904         & 110.42525         & \textbf{124.53642} & 110.34954         & 109.14357          & \textbf{128.66144}  & 128.46879           & 118.97199       \\
                        & mean   & 112.23817        & 122.10332          & 118.89049          & \textbf{124.60458}  & 119.23066            & 116.44989           & 114.04081         & 111.62820         & 124.36701          & 109.90154         & 110.39033          & \textbf{128.65648}  & 128.27652           & 118.97475       \\
                        & std.   & 2.86557          & 2.45026            & 3.12578            & 2.81117             & 4.61439              & 3.38666             & 3.48458           & 3.96786           & \textbf{2.63866}   & 5.13158           & 8.74329            & \textbf{0.03856}    & 0.43856             & 3.65352         \\
    \midrule
\multirow{5}{*}{GSP-1}  & min    & 24.78310         & 24.75344           & 24.74722           & 24.87687            & 24.75470             & 24.79914            & 24.85997          & 24.83532          & \textbf{24.89678}  & \textbf{24.90239} & 24.87792           & 24.83675            & 24.82629            & 24.87927        \\
                        & max    & 24.86403         & 24.89280           & 24.86918           & 24.90100            & 24.87424             & 24.88838            & 24.89649          & 24.89728          & \textbf{24.90152}  & \textbf{24.90409} & 24.90238           & 24.86241            & 24.88967            & 24.88986        \\
                        & median & 24.84693         & 24.86252           & 24.81018           & 24.89770            & 24.80560             & 24.86417            & 24.88191          & 24.86665          & \textbf{24.89920}  & \textbf{24.90329} & 24.88890           & 24.85351            & 24.86661            & 24.88313        \\
                        & mean   & 24.83969         & 24.84320           & 24.80647           & 24.89566            & 24.81274             & 24.85544            & 24.88002          & 24.87061          & \textbf{24.89916}  & \textbf{24.90323} & 24.88944           & 24.85255            & 24.86371            & 24.88369        \\
                        & std.   & 0.02411          & 0.04169            & 0.03545            & 0.00641             & 0.04044              & 0.02876             & 0.01170           & 0.01704           & \textbf{0.00137}   & \textbf{0.00041}  & 0.00763            & 0.00594             & 0.01956             & 0.00346         \\
    \midrule
\multirow{5}{*}{GSP-2}  & min    & 21.62841         & 23.05366           & 21.16330           & \textbf{24.22289}   & 20.64925             & 21.79885            & 23.26606          & 24.22049          & 24.20442           & 23.63179          & \textbf{24.20770}  & 24.04059            & 23.06899            & 22.56791        \\
                        & max    & 24.22882         & 24.22552           & 22.31575           & 24.23455            & 22.42779             & 23.73875            & 24.23412          & \textbf{24.23535} & 24.22268           & 23.92806          & \textbf{24.23494}  & 24.20554            & 23.94250            & 24.12619        \\
                        & median & 23.55838         & 24.22191           & 21.61177           & \textbf{24.22949}   & 21.32804             & 22.84138            & 24.23107          & 24.22535          & 24.21698           & 23.69532          & \textbf{24.22579}  & 24.19739            & 23.56759            & 23.67661        \\
                        & mean   & 23.43142         & 23.92663           & 21.59596           & \textbf{24.22975}   & 21.40222             & 22.79435            & 24.16649          & 24.22672          & 24.21558           & 23.74396          & \textbf{24.22494}  & 24.18290            & 23.52211            & 23.61278        \\
                        & std.   & 0.83896          & 0.42627            & 0.29860            & \textbf{0.00334}    & 0.44908              & 0.61373             & 0.24910           & 0.00452           & 0.00491            & 0.10252           & \textbf{0.00828}   & 0.04158             & 0.30080             & 0.49082         \\
    \midrule
\multirow{5}{*}{MaF1}   & min    & 24.44395         & 24.43011           & 24.39512           & 24.41442            & 24.34744             & 24.43731            & 24.45102          & \textbf{24.45486} & 24.34809           & 24.22594          & \textbf{24.48677}  & 24.21248            & 24.41758            & 24.29784        \\
                        & max    & 24.45312         & 24.45215           & 24.42527           & 24.42834            & 24.38795             & 24.45606            & 24.46424          & \textbf{24.46825} & 24.40361           & 24.31138          & \textbf{24.49244}  & 24.30784            & 24.45709            & 24.35433        \\
                        & median & 24.44991         & 24.44470           & 24.41118           & 24.41900            & 24.37264             & 24.44607            & 24.45803          & \textbf{24.46469} & 24.38009           & 24.26409          & \textbf{24.49079}  & 24.26038            & 24.44617            & 24.32305        \\
                        & mean   & 24.44960         & 24.44289           & 24.41167           & 24.42054            & 24.36983             & 24.44604            & 24.45780          & \textbf{24.46367} & 24.37937           & 24.26382          & \textbf{24.49038}  & 24.26379            & 24.44518            & 24.32548        \\
                        & std.   & 0.00295          & 0.00572            & 0.00933            & 0.00400             & 0.01112              & 0.00498             & \textbf{0.00284}  & 0.00370           & 0.01517            & 0.02677           & \textbf{0.00164}   & 0.02542             & 0.01079             & 0.01369         \\
    \midrule
\multirow{5}{*}{MaF5}   & min    & 4.99210          & 4.90357            & 4.78350            & 4.49367             & 4.63685              & 4.62885             & \textbf{4.98309}  & 4.96789           & 4.97845            & 4.21004           & 4.91500            & 14.33572            & \textbf{16.06217}   & 10.39669        \\
                        & max    & 17.21314         & 17.87484           & 17.55057           & 15.55907            & 17.36472             & 17.80466            & \textbf{17.96383} & 17.79883          & 17.56295           & 15.61226          & 17.86876           & 16.94406            & \textbf{18.12792}   & 16.06160        \\
                        & median & 4.99841          & 10.26329           & 16.40577           & 10.89141            & 16.57681             & 4.98166             & 4.99909           & 12.58571          & \textbf{16.84995}  & 10.83461          & 16.62098           & 16.19690            & \textbf{17.63682}   & 14.32153        \\
                        & mean   & 7.39246          & 10.81622           & 14.12557           & 10.22810            & 14.23553             & 8.24190             & 6.63787           & 10.76061          & \textbf{15.23755}  & 10.06386          & 13.84725           & 15.98673            & \textbf{17.41411}   & 14.14249        \\
                        & std.   & 4.95975          & 5.55852            & 4.86901            & 4.83974             & 4.88930              & 5.76168             & 4.33819           & 5.73301           & \textbf{4.18842}   & 4.26116           & 5.17367            & 0.78569             & \textbf{0.60251}    & 1.78766         \\
    \midrule
\multirow{5}{*}{MaF12}  & min    & 12.29540         & 12.09839           & 13.11170           & \textbf{13.87125}   & 11.19373             & 13.77272            & 12.12714          & 11.19667          & 12.62125           & 14.14035          & 13.83735           & 13.18278            & \textbf{14.44380}   & 12.88543        \\
                        & max    & 14.31010         & 14.78046           & \textbf{14.92758}  & 14.88899            & 14.52885             & 14.76182            & 14.76002          & 14.89184          & 14.82729           & 14.85499          & \textbf{15.35306}  & 14.88684            & 15.20315            & 14.28985        \\
                        & median & 13.57286         & 14.09620           & 14.38516           & 14.35152            & 13.58159             & \textbf{14.49131}   & 14.13178          & 13.60929          & 14.24934           & 14.56024          & \textbf{15.00409}  & 14.51359            & 14.78393            & 13.83739        \\
                        & mean   & 13.45678         & 13.87346           & 14.28025           & 14.42725            & 13.20671             & \textbf{14.44932}   & 13.88373          & 13.40946          & 14.05550           & 14.57274          & \textbf{14.86567}  & 14.37650            & 14.78957            & 13.69051        \\
                        & std.   & 0.52329          & 0.76260            & 0.65405            & 0.33467             & 1.00392              & \textbf{0.32176}    & 0.79363           & 1.19060           & 0.54431            & 0.23177           & 0.38210            & 0.47079             & \textbf{0.20370}    & 0.46769         \\
\bottomrule
\end{tabular}
\end{adjustbox}
\end{sidewaystable}

\begin{sidewaystable}[]
\ContinuedFloat
\caption{Empirical Comparisons w.r.t HV (Continued).}
\label{tab:HVComparison2}
\centering
\begin{adjustbox}{width=20cm ,center, angle = 0}    
    \begin{tabular}{c|c|ccccccccc|ccccc}
    \toprule
                        Problem &   HV & PoI     & $\kpoi_{\mbox{all}}$  & $\kpoi_{\mbox{one}}$  & $\kpoi_{\mbox{best}}$  & $\kpoi_{\mbox{worst}}$  & $\kpoi_{\mbox{mean}}$  & {KB-PoI}   &{CL-PoI}   & {q-EHVI}     & {TSEMO}    & {DGEMO}     & {MOEA/D-EGO} & {MOEA/D-ASS} & {ParEGO} \\
    \midrule
\multirow{5}{*}{mDTLZ1} & min    & 0.00000          & 0.00000            & 0.00000            & 0.00000             & 0.00000              & 0.00000             & 0.00000           & 0.00000           & 0.00000            & 0.00000           & 0.00000            & 0.00000             & 0.00000             & 0.00000         \\
                        & max    & 0.00000          & 0.00000            & 0.00000            & 0.00000             & 0.00000              & 0.00000             & 0.00000           & 0.00000           & 0.00000            & 0.00000           & 0.00000            & 0.00000             & 0.00000             & 0.00000         \\
                        & median & 0.00000          & 0.00000            & 0.00000            & 0.00000             & 0.00000              & 0.00000             & 0.00000           & 0.00000           & 0.00000            & 0.00000           & 0.00000            & 0.00000             & 0.00000             & 0.00000         \\
                        & mean   & 0.00000          & 0.00000            & 0.00000            & 0.00000             & 0.00000              & 0.00000             & 0.00000           & 0.00000           & 0.00000            & 0.00000           & 0.00000            & 0.00000             & 0.00000             & 0.00000         \\
                        & std.   & 0.00000          & 0.00000            & 0.00000            & 0.00000             & 0.00000              & 0.00000             & 0.00000           & 0.00000           & 0.00000            & 0.00000           & 0.00000            & 0.00000             & 0.00000             & 0.00000         \\
    \midrule
\multirow{5}{*}{mDTLZ2} & min    & 1.18247          & 1.17364            & \textbf{1.20386}   & 1.13538             & 1.16227              & 1.17192             & 1.19717           & 1.19466           & 1.12891            & 1.06826           & \textbf{1.22216}   & 1.06720             & 1.19868             & 1.11195         \\
                        & max    & 1.19715          & 1.18787            & \textbf{1.21162}   & 1.17687             & 1.17886              & 1.19507             & 1.20662           & 1.20399           & 1.14050            & 1.10199           & \textbf{1.22257}   & 1.11098             & 1.20429             & 1.12973         \\
                        & median & 1.19424          & 1.18143            & \textbf{1.20772}   & 1.16128             & 1.17038              & 1.19125             & 1.20068           & 1.20076           & 1.13422            & 1.08004           & \textbf{1.22242}   & 1.08276             & 1.20316             & 1.12274         \\
                        & mean   & 1.19175          & 1.18117            & \textbf{1.20742}   & 1.15936             & 1.17092              & 1.18787             & 1.20129           & 1.20034           & 1.13431            & 1.08298           & \textbf{1.22242}   & 1.08184             & 1.20217             & 1.12195         \\
                        & std.   & 0.00510          & 0.00456            & \textbf{0.00225}   & 0.01015             & 0.00491              & 0.00696             & 0.00261           & 0.00242           & 0.00409            & 0.00971           & \textbf{0.00011}   & 0.01110             & 0.00200             & 0.00520         \\
    \midrule
\multirow{5}{*}{mDTLZ3} & min    & 0.00000          & 0.00000            & 0.00000            & 0.00000             & 0.00000              & 0.00000             & 0.00000           & 0.00000           & 0.00000            & 0.00000           & 0.00000            & 0.00000             & 0.00000             & 0.00000         \\
                        & max    & 0.00000          & 0.00000            & 0.00000            & 0.00000             & 0.00000              & 0.00000             & 0.00000           & 0.00000           & 0.00000            & 0.00000           & 0.00000            & 0.00000             & 0.00000             & 0.00000         \\
                        & median & 0.00000          & 0.00000            & 0.00000            & 0.00000             & 0.00000              & 0.00000             & 0.00000           & 0.00000           & 0.00000            & 0.00000           & 0.00000            & 0.00000             & 0.00000             & 0.00000         \\
                        & mean   & 0.00000          & 0.00000            & 0.00000            & 0.00000             & 0.00000              & 0.00000             & 0.00000           & 0.00000           & 0.00000            & 0.00000           & 0.00000            & 0.00000             & 0.00000             & 0.00000         \\
                        & std.   & 0.00000          & 0.00000            & 0.00000            & 0.00000             & 0.00000              & 0.00000             & 0.00000           & 0.00000           & 0.00000            & 0.00000           & 0.00000            & 0.00000             & 0.00000             & 0.00000         \\
    \midrule
\multirow{5}{*}{mDTLZ4} & min    & 0.14987          & 0.22788            & \textbf{0.62268}   & 0.60772             & 0.39301              & 0.04790             & 0.06076           & 0.04789           & 0.04789            & 0.21700           & \textbf{0.59513}   & 0.00115             & 0.54565             & 0.30253         \\
                        & max    & 0.15064          & 0.66733            & 0.70684            & \textbf{0.75891}    & 0.62987              & 0.44427             & 0.24908           & 0.21974           & 0.40188            & 0.47250           & 0.64282            & 0.54772             & \textbf{0.70683}    & 0.60599         \\
                        & median & 0.14994          & 0.57692            & \textbf{0.67170}   & 0.66779             & 0.54108              & 0.15108             & 0.08335           & 0.06076           & 0.14987            & 0.40027           & 0.61511            & 0.42890             & \textbf{0.65188}    & 0.52318         \\
                        & mean   & 0.15006          & 0.55761            & \textbf{0.67182}   & 0.66795             & 0.53864              & 0.15833             & 0.12413           & 0.10475           & 0.19488            & 0.39087           & 0.61799            & 0.38689             & \textbf{0.63501}    & 0.51598         \\
                        & std.   & \textbf{0.00023} & 0.10212            & 0.02391            & 0.04183             & 0.06392              & 0.11593             & 0.06257           & 0.06677           & 0.12888            & 0.07138           & \textbf{0.01594}   & 0.13626             & 0.05552             & 0.06660         \\
    \midrule
\multirow{5}{*}{WOSGZ1} & min    & 0.09124          & 0.36920            & 0.43538            & 0.35040             & \textbf{0.50121}     & 0.30403             & 0.18840           & 0.20666           & 0.31760            & 0.00000           & \textbf{0.84176}   & 0.21969             & 0.69015             & 0.00000         \\
                        & max    & 0.44456          & 0.51508            & 0.65973            & 0.53922             & \textbf{0.65999}     & 0.53699             & 0.49283           & 0.53554           & 0.62360            & 0.11419           & \textbf{0.86858}   & 0.42256             & 0.74220             & 0.16315         \\
                        & median & 0.28087          & 0.42230            & 0.54496            & 0.41645             & \textbf{0.54675}     & 0.45439             & 0.35697           & 0.38669           & 0.49350            & 0.00413           & \textbf{0.85721}   & 0.33464             & 0.71443             & 0.08500         \\
                        & mean   & 0.29095          & 0.43084            & \textbf{0.55713}   & 0.42684             & 0.55690              & 0.44786             & 0.35117           & 0.38014           & 0.47673            & 0.02307           & \textbf{0.85665}   & 0.32615             & 0.71458             & 0.07358         \\
                        & std.   & 0.10831          & 0.03932            & 0.05818            & 0.05119             & 0.05164              & 0.07715             & 0.09349           & 0.09723           & 0.10625            & 0.03174           & \textbf{0.00891}   & 0.06496             & 0.01689             & 0.06545         \\
    \midrule
\multirow{5}{*}{WOSGZ2} & min    & 0.23095          & 0.17725            & \textbf{0.43861}   & 0.25468             & 0.38003              & 0.42108             & 0.19392           & 0.31603           & 0.19999            & 0.00000           & \textbf{0.81442}   & 0.08887             & 0.52311             & 0.00000         \\
                        & max    & 0.51895          & 0.48484            & \textbf{0.60508}   & 0.60110             & 0.59329              & 0.56796             & 0.54661           & 0.53244           & 0.48873            & 0.13596           & \textbf{0.84831}   & 0.39101             & 0.72006             & 0.18605         \\
                        & median & 0.38811          & 0.40202            & \textbf{0.51982}   & 0.44581             & 0.49753              & 0.49449             & 0.43718           & 0.42908           & 0.35457            & 0.00680           & \textbf{0.83402}   & 0.24676             & 0.68478             & 0.06216         \\
                        & mean   & 0.37360          & 0.37048            & \textbf{0.51870}   & 0.42827             & 0.49641              & 0.48814             & 0.40993           & 0.42844           & 0.37470            & 0.03579           & \textbf{0.83326}   & 0.25877             & 0.65642             & 0.08037         \\
                        & std.   & 0.09342          & 0.09790            & 0.05136            & 0.10008             & 0.06494              & \textbf{0.04312}    & 0.09850           & 0.06177           & 0.08700            & 0.04685           & \textbf{0.01087}   & 0.09488             & 0.07078             & 0.06328         \\
    \midrule
\multirow{5}{*}{WOSGZ3} & min    & 0.00000          & 0.26439            & 0.22301            & 0.02946             & \textbf{0.46227}     & 0.26421             & 0.01186           & 0.15321           & 0.09641            & 0.00000           & \textbf{0.78883}   & 0.03071             & 0.39616             & 0.00000         \\
                        & max    & 0.19215          & 0.51540            & 0.51200            & \textbf{0.59059}    & 0.55601              & 0.52051             & 0.49492           & 0.47166           & 0.47784            & 0.07147           & \textbf{0.83697}   & 0.42316             & 0.66485             & 0.15412         \\
                        & median & 0.04833          & 0.46096            & 0.45600            & 0.42681             & \textbf{0.49695}     & 0.40208             & 0.26267           & 0.37809           & 0.42442            & 0.00000           & \textbf{0.81030}   & 0.26088             & 0.60362             & 0.00957         \\
                        & mean   & 0.06185          & 0.42876            & 0.43707            & 0.40045             & \textbf{0.50529}     & 0.41174             & 0.28079           & 0.35858           & 0.39070            & 0.00786           & \textbf{0.81375}   & 0.25357             & 0.58068             & 0.04271         \\
                        & std.   & 0.06032          & 0.06502            & 0.07623            & 0.16594             & \textbf{0.03281}     & 0.07269             & 0.12227           & 0.07727           & 0.09969            & 0.01957           & \textbf{0.01335}   & 0.12150             & 0.07423             & 0.05636         \\
    \midrule
\multirow{5}{*}{WOSGZ4} & min    & 0.11851          & 0.14708            & 0.24885            & 0.26282             & \textbf{0.37081}     & 0.30107             & 0.26660           & 0.14300           & 0.17616            & 0.00000           & \textbf{0.77697}   & 0.10198             & 0.37824             & 0.00000         \\
                        & max    & 0.43325          & 0.20279            & \textbf{0.59515}   & 0.52653             & 0.49952              & 0.49908             & 0.47525           & 0.44972           & 0.50808            & 0.03116           & \textbf{0.80784}   & 0.47208             & 0.62254             & 0.13715         \\
                        & median & 0.26183          & 0.19191            & 0.40124            & 0.40657             & \textbf{0.40947}     & 0.41984             & 0.39967           & 0.39104           & 0.43017            & 0.00000           & \textbf{0.79285}   & 0.26269             & 0.50351             & 0.05079         \\
                        & mean   & 0.26707          & 0.18503            & \textbf{0.42088}   & 0.40569             & 0.41721              & 0.41106             & 0.39461           & 0.33645           & 0.40426            & 0.00208           & \textbf{0.79315}   & 0.29240             & 0.49249             & 0.05630         \\
                        & std.   & 0.07897          & \textbf{0.02414}   & 0.09036            & 0.07906             & 0.03801              & 0.06226             & 0.06980           & 0.10242           & 0.08855            & 0.00804           & \textbf{0.00983}   & 0.11508             & 0.07814             & 0.05368         \\
\bottomrule
\end{tabular}
\end{adjustbox}
\end{sidewaystable}

\begin{sidewaystable}[]
\centering
\ContinuedFloat
\caption{Empirical Comparisons w.r.t HV (Continued).}
\label{tab:HVComparison3}
\begin{adjustbox}{width=20cm ,center, angle = 0}    

    \begin{tabular}{c|c|ccccccccc|ccccc}
    \toprule
                        Problem &   HV & PoI     & $\kpoi_{\mbox{all}}$  & $\kpoi_{\mbox{one}}$  & $\kpoi_{\mbox{best}}$  & $\kpoi_{\mbox{worst}}$  & $\kpoi_{\mbox{mean}}$  & {KB-PoI}   &{CL-PoI}   & {q-EHVI}     & {TSEMO}    & {DGEMO}     & {MOEA/D-EGO} & {MOEA/D-ASS} & {ParEGO} \\
\midrule
\multirow{5}{*}{WOSGZ4} & min    & 0.11851          & 0.14708            & 0.24885            & 0.26282             & \textbf{0.37081}     & 0.30107             & 0.26660           & 0.14300           & 0.17616            & 0.00000           & \textbf{0.77697}   & 0.10198             & 0.37824             & 0.00000         \\
                        & max    & 0.43325          & 0.20279            & \textbf{0.59515}   & 0.52653             & 0.49952              & 0.49908             & 0.47525           & 0.44972           & 0.50808            & 0.03116           & \textbf{0.80784}   & 0.47208             & 0.62254             & 0.13715         \\
                        & median & 0.26183          & 0.19191            & 0.40124            & 0.40657             & \textbf{0.40947}     & 0.41984             & 0.39967           & 0.39104           & 0.43017            & 0.00000           & \textbf{0.79285}   & 0.26269             & 0.50351             & 0.05079         \\
                        & mean   & 0.26707          & 0.18503            & \textbf{0.42088}   & 0.40569             & 0.41721              & 0.41106             & 0.39461           & 0.33645           & 0.40426            & 0.00208           & \textbf{0.79315}   & 0.29240             & 0.49249             & 0.05630         \\
                        & std.   & 0.07897          & \textbf{0.02414}   & 0.09036            & 0.07906             & 0.03801              & 0.06226             & 0.06980           & 0.10242           & 0.08855            & 0.00804           & \textbf{0.00983}   & 0.11508             & 0.07814             & 0.05368         \\
    \midrule
\multirow{5}{*}{WOSGZ5} & min    & 0.00000          & 0.19230            & \textbf{0.23698}   & 0.01706             & 0.00000              & 0.00000             & 0.00000           & 0.00000           & 0.00000            & 0.00000           & \textbf{0.70921}   & 0.07792             & 0.27686             & 0.00000         \\
                        & max    & 0.16674          & \textbf{0.48142}   & 0.46770            & 0.41157             & 0.45358              & 0.41707             & 0.19333           & 0.30284           & 0.39868            & 0.14773           & \textbf{0.75531}   & 0.35811             & 0.40148             & 0.03008         \\
                        & median & 0.00000          & 0.28674            & \textbf{0.37156}   & 0.24757             & 0.18488              & 0.22081             & 0.00000           & 0.23251           & 0.13808            & 0.00000           & \textbf{0.73486}   & 0.22950             & 0.34272             & 0.00000         \\
                        & mean   & 0.04184          & 0.30041            & \textbf{0.36572}   & 0.21436             & 0.20600              & 0.19533             & 0.05131           & 0.20199           & 0.16600            & 0.01036           & \textbf{0.73498}   & 0.21051             & 0.34780             & 0.00346         \\
                        & std.   & 0.06213          & 0.08786            & \textbf{0.05604}   & 0.10455             & 0.17541              & 0.13544             & 0.07725           & 0.08346           & 0.15124            & 0.03805           & \textbf{0.01694}   & 0.09335             & 0.03590             & 0.00927         \\
    \midrule
\multirow{5}{*}{WOSGZ6} & min    & 0.13158          & 0.06614            & \textbf{0.27422}   & 0.17468             & 0.21386              & 0.01753             & 0.00000           & 0.00000           & 0.00000            & 0.00000           & \textbf{0.69368}   & 0.05556             & 0.27429             & 0.00000         \\
                        & max    & 0.30733          & 0.35878            & \textbf{0.44799}   & 0.44425             & 0.42453              & 0.37226             & 0.26309           & 0.29163           & 0.36966            & 0.00000           & \textbf{0.78148}   & 0.44681             & 0.41242             & 0.00211         \\
                        & median & 0.22215          & 0.27503            & 0.30523            & 0.23442             & \textbf{0.33864}     & 0.24462             & 0.16645           & 0.23003           & 0.19630            & 0.00000           & \textbf{0.72363}   & 0.28269             & 0.32003             & 0.00000         \\
                        & mean   & 0.21768          & 0.25408            & 0.32500            & 0.26932             & \textbf{0.32977}     & 0.21798             & 0.13483           & 0.20758           & 0.19872            & 0.00000           & \textbf{0.72591}   & 0.27614             & 0.32539             & 0.00028         \\
                        & std.   & 0.04521          & 0.07297            & \textbf{0.05122}   & 0.08890             & 0.05317              & 0.10678             & 0.10602           & 0.08595           & 0.11375            & 0.00000           & \textbf{0.02334}   & 0.13764             & 0.03813             & 0.00074         \\
    \midrule
\multirow{5}{*}{WOSGZ7} & min    & 0.04469          & 0.05325            & \textbf{0.10639}   & 0.04497             & 0.09294              & 0.00000             & 0.01362           & 0.05164           & 0.02982            & 0.00000           & 0.00000            & 0.00000             & \textbf{0.15951}    & 0.00000         \\
                        & max    & 0.14575          & 0.22187            & \textbf{0.35448}   & 0.14289             & 0.35399              & 0.16404             & 0.17970           & 0.16555           & 0.18714            & 0.00000           & 0.22443            & 0.00304             & \textbf{0.35101}    & 0.00000         \\
                        & median & 0.08557          & 0.08726            & \textbf{0.17530}   & 0.06914             & 0.16104              & 0.04320             & 0.07002           & 0.10506           & 0.12540            & 0.00000           & 0.17556            & 0.00000             & \textbf{0.27242}    & 0.00000         \\
                        & mean   & 0.08735          & 0.10690            & \textbf{0.18772}   & 0.07791             & 0.17663              & 0.04902             & 0.07721           & 0.10956           & 0.12580            & 0.00000           & 0.14873            & 0.00020             & \textbf{0.26252}    & 0.00000         \\
                        & std.   & 0.03218          & 0.04904            & 0.06204            & \textbf{0.02903}    & 0.06018              & 0.05491             & 0.05461           & 0.03348           & 0.04132            & 0.00000           & 0.07126            & \textbf{0.00078}    & 0.05700             & 0.00000         \\
    \midrule
\multirow{5}{*}{WOSGZ8} & min    & 0.42719          & 0.58854            & 0.76026            & 0.74336             & \textbf{0.76379}     & 0.54815             & 0.58127           & 0.69011           & 0.59456            & 0.00000           & \textbf{1.20256}   & 0.49690             & 0.84191             & 0.07449         \\
                        & max    & 0.77575          & 0.91634            & 0.91561            & 0.96338             & \textbf{0.91701}     & 0.93606             & 0.86019           & 0.88487           & 0.93114            & 0.20486           & \textbf{1.22642}   & 0.96339             & 1.04015             & 0.42273         \\
                        & median & 0.56303          & 0.81753            & 0.81896            & 0.85445             & \textbf{0.84356}     & 0.78780             & 0.75715           & 0.76224           & 0.84950            & 0.01589           & \textbf{1.21432}   & 0.70529             & 0.91844             & 0.24869         \\
                        & mean   & 0.60300          & 0.80795            & 0.82062            & 0.84928             & \textbf{0.84741}     & 0.75319             & 0.75317           & 0.78084           & 0.82227            & 0.05353           & \textbf{1.21488}   & 0.71125             & 0.93911             & 0.25920         \\
                        & std.   & 0.10915          & 0.08498            & 0.05158            & 0.07748             & \textbf{0.04528}     & 0.12052             & 0.08150           & 0.05594           & 0.10718            & 0.06731           & \textbf{0.00705}   & 0.14371             & 0.06766             & 0.11853        \\
\bottomrule
\end{tabular}
\end{adjustbox}
\end{sidewaystable}
\clearpage